\documentclass{fairmeta}
\usepackage{amsmath}
\usepackage{amssymb}
\usepackage{xspace}
\usepackage{arydshln}
\usepackage[symbol]{footmisc}
\usepackage{enumitem}
\usepackage{pgffor}
\usepackage{wrapfig}

\newcommand{\method}{WorldGen\xspace}
\newcommand{\assetgen}{AssetGen2\xspace}
\newcommand{\trellis}{TRELLIS\xspace}
\newcommand{\autopartgen}{AutoPartGen\xspace}
\newcommand{\recast}{Recast\xspace}
\newcommand{\vecset}{VecSet\xspace}
\newcommand{\marble}{Marble\xspace}
\newcommand{\navmesh}{navmesh\xspace}
\newcommand{\sanjose}{Top Image-to-3D Model A\xspace}

\newcommand{\partpacker}{Top PartGen Model A\xspace}
\newcommand{\bigsurpart}{Top PartGen Model B\xspace}

\newcommand{\x}{\boldsymbol{x}}
\newcommand{\z}{\boldsymbol{z}}

\newcommand{\cX}{\mathcal{X}}
\newcommand{\cL}{\mathcal{L}}

\definecolor{tabfirst}{rgb}{1, 0.7, 0.7}
\definecolor{tabsecond}{rgb}{1, 0.85, 0.7}
\definecolor{tabthird}{rgb}{1, 1, 0.7}

\definecolor{metabluelight}{HTML}{CCE0F8}
\definecolor{metabluemedium}{HTML}{6D92BF}

\usepackage{float}

\title{\method: From Text to Traversable and Interactive 3D Worlds}

 \author[\dagger]{Dilin Wang}
 \author[]{Hyunyoung Jung}
 \author[]{Tom Monnier}
 \author[]{Kihyuk Sohn}
 \author[]{Chuhang Zou}
 \author[]{Xiaoyu Xiang}
 \author[]{Yu-Ying Yeh}
 \author[]{Di Liu}
 \author[]{Zixuan Huang}
 \author[]{Thu Nguyen-Phuoc}
 \author[]{Yuchen Fan}
 \author[]{Sergiu Oprea}
 \author[]{Ziyan Wang}
 \author[]{Roman Shapovalov}
 \author[]{Nikolaos Sarafianos}
 \author[]{Thibault Groueix}
 \author[]{Antoine Toisoul}
 \author[]{Prithviraj Dhar}
 \author[]{Xiao Chu}
 \author[]{Minghao Chen}
 \author[]{Geon Yeong Park}
 \author[]{Mahima Gupta}
 \author[]{Yassir Azziz}
\author[\dagger]{Rakesh Ranjan}
\author[\dagger]{Andrea Vedaldi}

 \affiliation[]{Reality Labs, Meta}

\contribution[\dagger]{project lead}

\abstract{
We introduce \method, a system that enables the automatic creation of large-scale, interactive 3D worlds directly from text prompts.
Our approach transforms natural language descriptions into traversable, fully textured environments that can be immediately explored or edited within standard game engines.
By combining LLM-driven scene layout reasoning, procedural generation, diffusion-based 3D generation, and object-aware scene decomposition, \method bridges the gap between creative intent and functional virtual spaces—allowing creators to design coherent, navigable worlds without manual modeling or specialized 3D expertise.
The system is fully modular and supports fine-grained control over layout, scale, and style, producing worlds that are geometrically consistent, visually rich, and efficient to render in real time.
This work represents a step towards accessible, generative world-building at scale, advancing the frontier of 3D generative AI for applications in gaming, simulation, and immersive social environments.
}

\date{\today}
\metadata[Blogpost]{\url{https://www.meta.com/blog/worldgen-3d-world-generation-reality-labs-generative-ai-research/}\\\\}

\makeatletter
\renewcommand\paragraph{\@startsection{paragraph}{4}{\z@}%
  {1.25ex \@plus 1ex \@minus .2ex}
  {-.5em}
  {\normalfont\normalsize\bfseries}} 
\makeatother

\begin{document}

\maketitle

\tableofcontents

\section{Introduction}%
\label{sec:intro}

3D interactive experiences such as video games are a major part of the creative industry.
However, creating 3D content is complex, time-consuming, and requires significant expertise and resources.
There is thus growing interest in leveraging recent advances in generative AI to automate 3D content creation.
This has the potential to dramatically reduce the time required to produce new games.
It can also empower anyone to become a creator and thus support new kinds of experience where content is generated on the fly, customized, and personalized by users.

3D generative AI (3D GenAI for short) has already made substantial progress in the past few years.
Similar to how we can generate high-quality images and videos, it is now possible to generate high-quality 3D objects from simple text prompts.
While there remain difficult challenges such as optimizing and reusing geometry and textures, these 3D generators are \emph{already} useful to artists and creators, as exemplified by our own \assetgen~\citep{MetaAssetGen2}.

Even so, generating 3D objects is but a small task in the process of creating full 3D experiences---the latter also require scenes, animations, interactions, gameplay mechanics, game levels, playable and non-playable characters, storylines, and more.
Interactive video generators~\citep{parker-holder24genie} may one day address in one swoop all such challenges by generating pixels directly from high-level prompts and user interactions; however, these are likely many years away from becoming a mature technology that can displace traditional world creation paradigms.
Meanwhile, 3D GenAIs will still need to output traditional representations of worlds that are compatible with existing game engines, hardware, and content creation models and pipelines.

In this technical report, we address some of these challenges by introducing \method, a system for generating 3D worlds from a single text prompt, end-to-end.
Compared to object generation, the key challenge of world generation is to create a composition of 3D objects that, as a whole, form a coherent and functional scene corresponding well to the user's intent.
Objects must fit together thematically, stylistically, and contextually.
For instance, a medieval scene should not contain a modern Oxford chair.
More subtly, all objects should conform to the same artistic style.
Structurally, objects must make sense in context; for example, a dining table should be surrounded by chairs.
The scene should also meet certain functional requirements to support a particular interactive experience.
For example, a common requirement is that the scene can be traversed by a character without getting `stuck' due to obstructions.

A major difficulty in building a system like \method is that there is no sufficiently large training set of 3D scenes that would allow learning a direct mapping from text prompt to 3D scene.
Hence, as commonly done in 3D object generation, we reduce the problem to first generating an image of the 3D scene, followed by image-to-3D reconstruction, both cast as conditional generation tasks.
In this way, we can leverage the impressive capabilities of existing text-to-image models, trained on billions of images, to help interpret the textual prompt and produce a `scene plan' that establishes which objects should be contained in the scene, how they should look, and how they should relate to one another.

While very helpful, we still found that even the best image generators struggle to imagine scenes that are functional, including being traversable.
To address this issue, we propose to guide image generation with a procedurally-generated layout of the scene.
\emph{Procedural generation} (PG) is a well-established technique in computer graphics that creates 3D environments algorithmically.
Because PGs are rule based, they can satisfy given constraints, which is useful to guarantee that the resulting scenes are viable.
However, they are controlled by means of custom parameters instead of natural language, and can only produce certain types of scenes, with limited stylistic and thematic diversity.

To address the first issue, we make the PG controllable via natural language by mapping the user-provided text prompt to the parameters required to configure the PG using a Large Language Model (LLM).
To address the second issue, we limit the PG to generating only the \emph{basic layout} of the 3D scene, similar to how 3D artists sketch worlds using ``blockouts''.
A blockout defines only the main volumes of the scene, its rough geometry, and its connectivity, represented by a so-called \emph{navigation mesh} (\navmesh).
A key aspect of \method is that the details of the scene are still determined by the image generator.
In particular, the PG does not specify the meaning or semantic class of the objects it places, leaving it to the image generator to interpret a given box as a tree, rock, or building.
Furthermore, the image generator is free to \emph{hallucinate additional small objects} that are not even hinted at in the blockout.
This allows \method to produce a large variety of detailed scenes while ensuring that they remain functional.

Given the blockout, \navmesh, and the generated image of the scene, we then apply an image-to-3D model to obtain a first \emph{holistic} reconstruction of the world.
For this, we use \assetgen, a state-of-the-art image-to-3D method that we recently developed.
We further finetune the model to also account for the \navmesh, preserving as much as possible the navigability of the world, even in areas that are not clearly visible in the image due to occlusions.
This also allows the model to maintain navigability, even if the image generator has hallucinated objects that are not fully compatible with the blockout intent and may obstruct navigation.

Reconstructing the entire scene holistically is a simple and yet highly effective way of ensuring global coherence and consistency.
However, this is insufficient to obtain a high-quality 3D world.
In particular, the resolution of the holistic reconstruction, as well as of the initial image, is insufficient to capture all details of the scene to a satisfactory degree.
Furthermore, if the output is a single mesh, the scene is difficult to handle by game engines, and also difficult to edit and make interactive.

We thus build on progress in \emph{compositional} 3D generation to decompose the scene into its constituent objects.
It is not sufficient to use the components in the initial blockout because these are \emph{not} in 1-to-1 correspondence with the objects in the scene, particularly because the image generator is free to hallucinate additional geometry.
Instead, we decompose the scene using a variant of AutoPartGen~\citep{chen25autopartgen}, a method we recently proposed for automated 3D part discovery and generation,
with optimisations to deal with the large number of parts typically present in scenes.
Notably, this model only requires as input the holistic mesh of the scene, and automatically extracts meaningful constituent objects such as buildings, trees, and so on.

Once the objects are separated, we further improve their quality by re-generating them individually.
This uses two ideas.
First, an image generator is used to create a new high-resolution view of each object, thus hallucinating many new details.
Second, a specialized version of \assetgen is used to re-generate the geometry and texture of each object from the new image.
Crucially, this step is conditioned on the low-resolution geometry of the part, which controls deviation from the initial reconstruction and ensures that the refreshed parts still fit together correctly.

Complementary to the architecture described above is significant work on data curation and engineering to train the various components effectively.
To this end, we have collected a substantial number of high-quality artist-created 3D scenes.
Additionally, we have developed a method to generate a large number of synthetic 3D scenes by composing 3D objects from our internal database.
This method can be considered a basic world generator in its own right, albeit without the control, diversity, and quality of \method.
Nevertheless, the data obtained in this manner is highly valuable for training various conditional generative models, including \assetgen and AutoPartGen, fine-tuning them for the construction of worlds rather than individual objects.

In the rest of this technical report, we first describe each component of \method in detail (\cref{sec:method}), along with its different stages:
scene planning (\cref{sec:scene-planning}),
scene reconstruction (\cref{sec:scene-reconstruction}),
scene decomposition (\cref{sec:scene-decomposition}),
and scene enhancement (\cref{sec:scene-enhancement}).
We then present results of the approach (\cref{sec:experiments}).
We discuss the related work in \cref{sec:related-work} and conclude in \cref{sec:conclusions}.

\begin{figure}[htbp]
\centering
\includegraphics[width=0.99\linewidth]{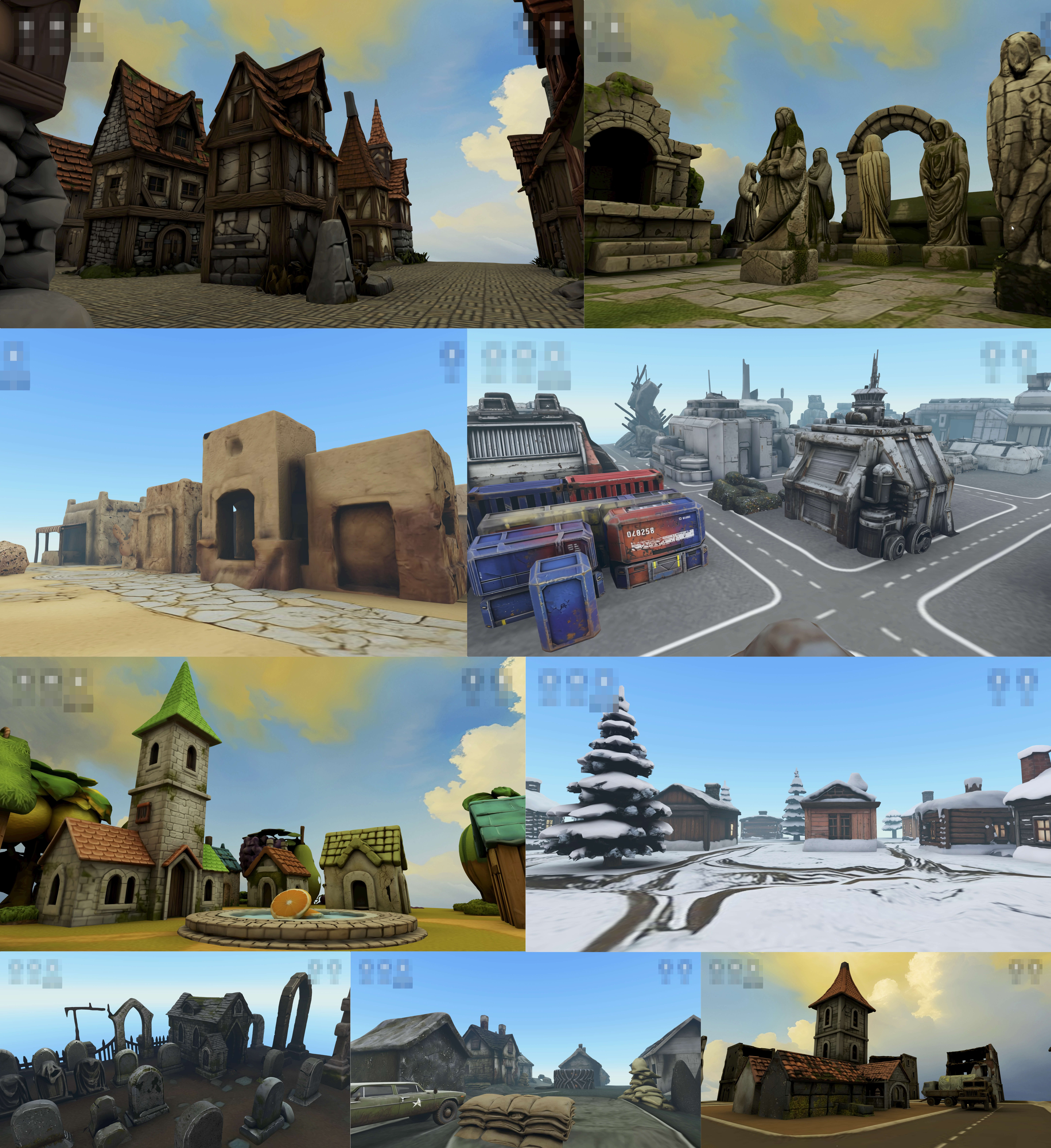}
\captionof{figure}{\textbf{World snapshots generated by \method.}
Each scene consists of individually editable objects represented as fully textured 3D meshes. As explicit geometry, these worlds naturally support collision, and navigation—allowing characters to climb, jump, and interact. 
The resulting assets are immediately deployable in game engines.}
\label{fig:teaser_0}
\end{figure}

\begin{figure}[htbp]
\centering
\includegraphics[width=0.99\linewidth]{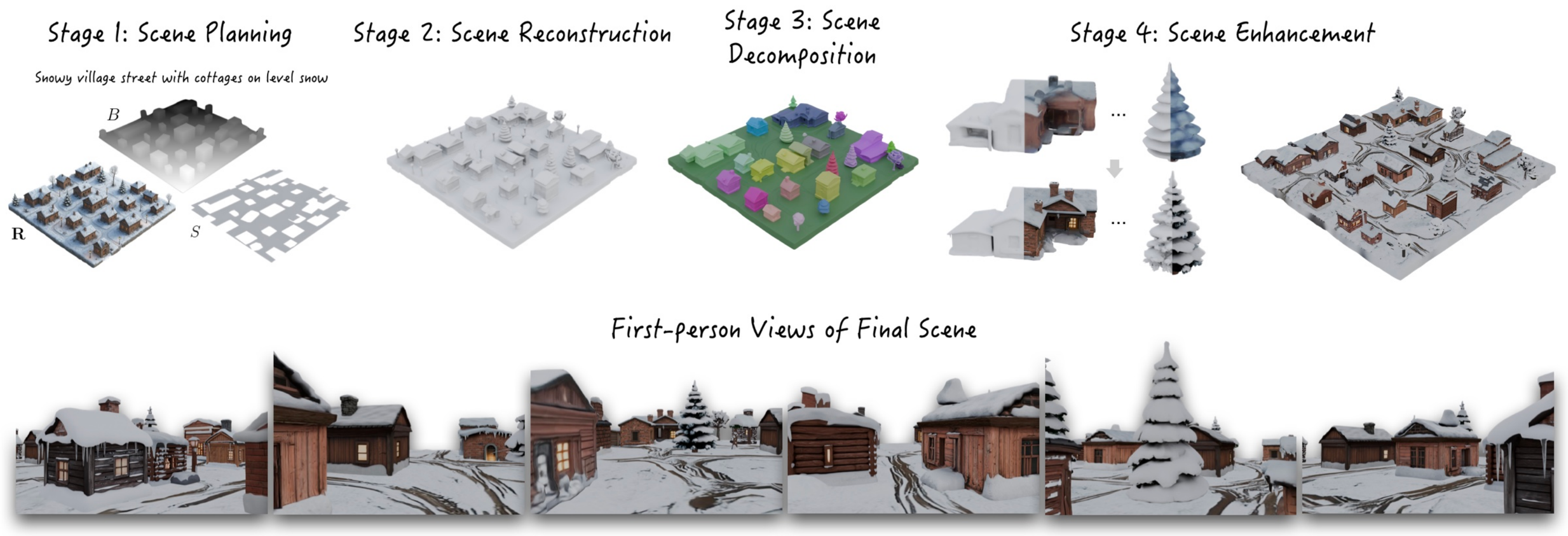}
\caption{\textbf{\method overview}. Our pipeline begins by planning the scene layout, producing a blockout ($B$), reference image ($\mathbf{R}$), and navigation mesh (S) (Stage 1). Next, we generate a single 3D mesh that aligns with this plan, preserving navigable areas and overall composition (Stage 2). The scene is then decomposed into individual entities (Stage 3), which are refined at higher resolution (Stage 4), resulting in a high-quality, traversable, and visually cohesive final scene.}
\label{fig:overview}
\end{figure}

\section{\method Overview}%
\label{sec:method}

\newcommand{\textPrompt}{y}
\newcommand{\xfine}{\x}
\newcommand{\xcoarse}{\hat{\x}}
\newcommand{\referenceImage}{\mathbf{R}}
\newcommand{\image}{\mathbf{I}}
\newcommand{\objectImageCoarse}{\hat{\mathbf{I}}}
\newcommand{\objectImageFine}{\mathbf{I}}
\newcommand{\blockout}{B}
\newcommand{\nav}{S}
\newcommand{\camera}{g}
\newcommand{\plan}{\cL}
\newcommand{\sceneMesh}{M}
\newcommand{\sceneFine}{\cX}
\newcommand{\sceneCoarse}{\hat{\cX}}

We now describe \method, our system for generating functional and compositional 3D scenes end-to-end, starting from a single text prompt.
Let $\textPrompt$ be a user prompt (e.g., ``medieval village'').
The output is a scene
$
\sceneFine = (\{ (\xfine_i, \camera_i) \}_{i=1}^N, S),
$
where each object $\xfine_i$ is a 3D shape with a UV texture, $\camera_i \in SE(3)$ specifies its rigid pose, and $S$ is its \navmesh, i.e., the walkable surface.
The generation process is stochastic, and $\sceneFine$ can be thought of as a sample from a conditional distribution $p(\sceneFine | \textPrompt)$ captured by the model.

\method consists of four stages which start from establishing the high-level structure of the scene based on the user intent and finish by specifying the scene components and low-level details.
We summarize the four stages below, and provide further details in
\cref{sec:scene-planning,sec:scene-reconstruction,sec:scene-decomposition,sec:scene-enhancement}. See Figure~\ref{fig:overview} for an illustration of the full pipeline. 

\paragraph{Stage I\@: Scene Planning.}

Given the text prompt $\textPrompt$, the first stage generates a \emph{scene plan} $\plan$ that specifies the overall spatial layout, style, theme, and composition of the scene.
The plan begins by generating a 3D \emph{blockout} $\blockout$ of the scene—a 3D sketch that describes the main scene structures, including open spaces, loops, or chokepoints, which affect the scene \emph{functionality}.
Based on the blockout $B$, planning produces two further outputs:
(i) a \emph{reference image} $\referenceImage$, obtained by rendering a depth map of $B$ and using it to condition a diffusion-based image generator, which establishes the theme, style, and details of the scene; and
(ii) a \emph{\navmesh} $\nav$, baked from $\blockout$, capturing the traversable regions of the scene.
Together, the scene plan
$
\plan = (\blockout, \referenceImage, \nav)
$
provides sufficient geometric and stylistic guidance for the subsequent stages to reconstruct a functional and high-quality 3D scene.

\paragraph{Stage II\@: Scene Reconstruction.}

Given the scene plan $\plan = (\blockout, \referenceImage, \nav)$, the second stage performs a \emph{holistic 3D reconstruction} of the scene, producing a \emph{single} textured mesh $\sceneMesh$ representing the scene \emph{as a whole}.
The reconstruction utilizes an image-to-3D generative model conditioned on both the reference image $\referenceImage$ and the \navmesh $\nav$.
This is an extension of our \assetgen model, which utilizes latent 3D diffusion.
The reconstruction is relatively low resolution (for a scene), but, crucially, it is performed holistically, which allows the model to resolve global spatial relationships between objects, ensure overall scene coherence, and complete the scene behind occlusions.
The reconstruction is further conditioned on the \navmesh $\nav$, which provides explicit spatial constraints to ensure the generated geometry conforms to traversable regions and maintains global connectivity, reducing artifacts such as non-reachable areas.
A rough colorization of the entire mesh $M$ is also obtained utilizing a volumetric texture generator.

\paragraph{Stage III\@: Scene Decomposition.}

Given the scene mesh $\sceneMesh$, the third stage decomposes it into a set
$
\sceneCoarse = \{ (\xcoarse_i, \camera_i) \}_{i=1}^N
$
of individual (low-resolution) 3D assets.
For example, in a ``medieval village'' scene, the decomposition could break the scene mesh into terrain, buildings, trees, and props.
This facilitates the subsequent enhancement of the scene, which can be done piece-by-piece.
It also facilitates editing the generated scene locally---for example to add a layer of moss to specific buildings---without having to re-generate the entire world from scratch.
To this end, we utilize AutoPartGen~\citep{chen25autopartgen}, a model that extract parts from a 3D mesh in an autoregressive manner.
This model only requires the mesh $\sceneMesh$ as input, and automatically determines the 3D components and their number.
We further upgrade AutoPartGen to work better with scene-like 3D objects and to deal with a large number of parts efficiently, combining it with ideas from PartPacker~\citep{tang25efficient}.
In this way, we obtain a complete set of parts that, once assembled, well reconstruct the original mesh $\sceneMesh$.

\paragraph{Stage IV\@: Scene Enhancement.}

Given the decomposed low-resolution scene objects
$
\sceneCoarse = \{(\xcoarse_i, \camera_i)\}_{i=1}^N
$
and the reference image $\referenceImage$ from Stage~I, the final stage enhances individual objects to achieve high visual fidelity and detail, outputting the final high-resolution scene
$
\sceneFine = (\{(\xfine_i, \camera_i)\}_{i=1}^N, \nav)
$.
First, we render image $\objectImageCoarse_i$ from 
a viewpoint selected to provide a clear and representative view of the low-resolution object’s 3D geometry.
These images are then used to condition an image generator, which produces a high-quality image $\objectImageFine_i$ of the object.
This approach preserves the object's overall structure while hallucinating fine-grained shape and appearance details.
Next, we reconstruct the shape of each object based on the new image $\objectImageFine_i$ as well as the coarse shape $\xcoarse_i$, using a mesh refinement model, which outputs a refined mesh $\xfine_i$ which is geometrically well aligned to $\xcoarse_i$.
Finally, a texturing model is applied to the refined mesh to generate high-resolution textures that align with the enhanced details in $\objectImageFine_i$.

\begin{figure}[t]
\centering
\includegraphics[width=\textwidth]{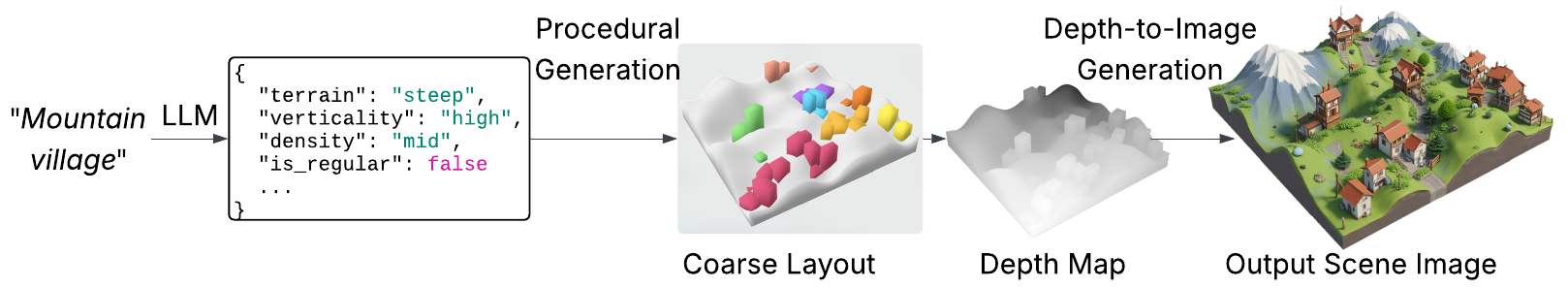}
\caption{\textbf{3D Layout Generation}. An LLM parses the input prompt into structured parameters (JSON) to drive a procedural generator, producing a coarse 3D blockout. This blockout is then rendered to depth, which conditions the generation of the final scene reference image.
}
\label{fig:procedural-layout-generation-pipeline}
\end{figure}

\section{Stage I: Scene Planning}
\label{sec:scene-planning}

\setlength\intextsep{0pt}

The goal of the scene planning stage is to transform a user text prompt $y$ into a rough but functionally-correct scene plan $\plan = (\blockout, \referenceImage, \nav)$, consisting of a \emph{block-out layout} $\blockout$, a \emph{reference image} $\referenceImage$,
and a \emph{navigational mesh (\navmesh)} $\nav$~\citep{snook2000simplified}.

The blockout is obtained via a text-conditioned procedural generation process that ensures global spatial coherence and navigability (\cref{sec:blockout-generation}).
Then the \navmesh and reference image are derived from the blockout using standard mesh processing tools and a depth-conditioned image generator (\cref{sec:navmesh-reference-image-generation}).
An overview of this process is illustrated in \cref{fig:procedural-layout-generation-pipeline} and detailed below.
This is a critical step to ensure the scene layout is structural and traversable.

\subsection{Procedural Blockout Generation}%
\label{sec:blockout-generation}

Traditional procedural generation (PG) produces coherent and functional environments based on hand-crafted rules and procedures, but can only handle a certain type of environment, and cannot be controlled using text expressed in natural language.
Inspired by recent text-conditioned procedural systems~\citep{raistrick23infinite,maleki24procedural}, we extend a PG system with a language interface.
An LLM parses the user prompt $\textPrompt$ into a structured JSON specification of parameters such as terrain type, object density, verticality, and placement regularity.
These parameters configure a modular PG pipeline that procedurally constructs a blockout $\blockout$ aligned with the user's intent.

In more detail, our procedural generation (PG) pipeline constructs the blockout in three steps: \emph{terrain generation}, \emph{spatial partitioning}, and \emph{hierarchical asset placement}.
First, \emph{terrain generation} constructs a base landscape that defines the large-scale geometry of the scene—such as elevation, slopes, and flat regions—providing a base for where structures and traversal paths can exist.
Next, \emph{spatial partitioning} divides the terrain into distinct regions that serve different scene purposes (e.g., open areas, clusters of structures, or transition zones).
This step provides a high-level organizational layout, ensuring that the scene has variation in density and structure while maintaining overall navigability.
Finally, \emph{hierarchical asset placement} populates each region with 3D assets in multiple passes: large landmark assets are placed first to establish structure and focal points, followed by smaller objects and decorative elements that add realism and detail.
This multi-level placement strategy produces consistent yet varied scenes, maintaining both functional organization and visual diversity.

\emph{(1) Terrain Generation.}
We synthesize the terrain using either a Perlin-noise generator~\citep{perlin1985image} or a rule-based height map configured by the parsed JSON specification.
The JSON parameters further define terrain attributes such as type (e.g., ``flat'',  ``steep''), surface roughness, and elevation range, which together control the overall topography and structural variation of the scene.

\emph{(2) Spatial Partitioning.} 
Spatial partitioning divides the terrain into distinct regions that provide structural organization for the scene.  
This step determines where dense clusters, open areas, and transitional zones appear.
For structured environments (e.g., ``urban,'' ``grid village''), we employ binary space partitioning~\citep{fuchs1980visible}, uniform grids, or $k$-d trees~\citep{bentley1975multidimensional} to produce regular, orthogonal layouts.  
For natural or irregular landscapes (e.g., ``archipelago,'' ``jungle''), we use Voronoi diagrams, noise-based partitions, or Drunkard’s Walk~\citep{pearson1905problem} to create organic, non-uniform boundaries.  
This process defines the macro-level organization of the environment, balancing structured regions with open space to ensure both navigability and visual diversity.

\emph{(3) Hierarchical Asset Placement.}
Finally, we populate the layout with blocks, which serve as placeholders for different categories of elements, in three passes to reflect structural hierarchy and spatial semantics.
i)~\emph{Hero assets} (major landmarks or buildings) are placed first.
ii)~\emph{Medium-scale elements} such as trees, walls, or bridges are distributed relative to hero assets.
iii)~\emph{Small decorative assets} fill residual spaces according to density and clustering parameters.
A final terrain-smoothing step prevents asset collisions, improving realism and playability of the blockout geometry.
While we use categories (i-iii) to generate a reasonable distribution of volumes, we do not make hard decisions on what these represent at this point; instead, we let the image generator in \cref{sec:navmesh-reference-image-generation} decide what they are.

The resulting blockout $\blockout$ above is a 3D mesh composed of simple primitives (ground plane and boxes) that encode the essential geometry of the scene.
It serves as an editable structural scaffold from which we subsequently derive the \navmesh $\nav$ and reference image $\referenceImage$.

\subsection{Navmesh Extraction and Reference Image Generation}%
\label{sec:navmesh-reference-image-generation}

The \navmesh $\nav$ (see ~\Cref{fig:procedural-layout-generation-pipeline}) is extracted directly from the solid blockout geometry $\blockout$
using a standard algorithm such as \recast~\citep{recast},
which robustly identifies exterior traversable surfaces while excluding indoor areas.
This ensures that $\nav$ accurately captures the walkable areas of the scene and serves as a structural constraint for downstream synthesis.

To generate the reference image $\referenceImage$, we render the block-out geometry $\blockout$ into an isometrically projected depth map,
which is then used as a condition to our depth-conditioned image generator.
We adopt a camera elevation of approximately $45^{\circ}$ to maximize visible coverage of the scene within the frame.
Additionally, to reduce the appearance of overly rectilinear shapes, we apply a small Gaussian perturbation to the non-terrain depth values,
scaled proportionally to depth, which encourages more natural, visually varied structural outlines in the resulting image.

\subsection{Planning Stage Results}

\begin{figure}[t]
\centering
\includegraphics[width=1\textwidth]{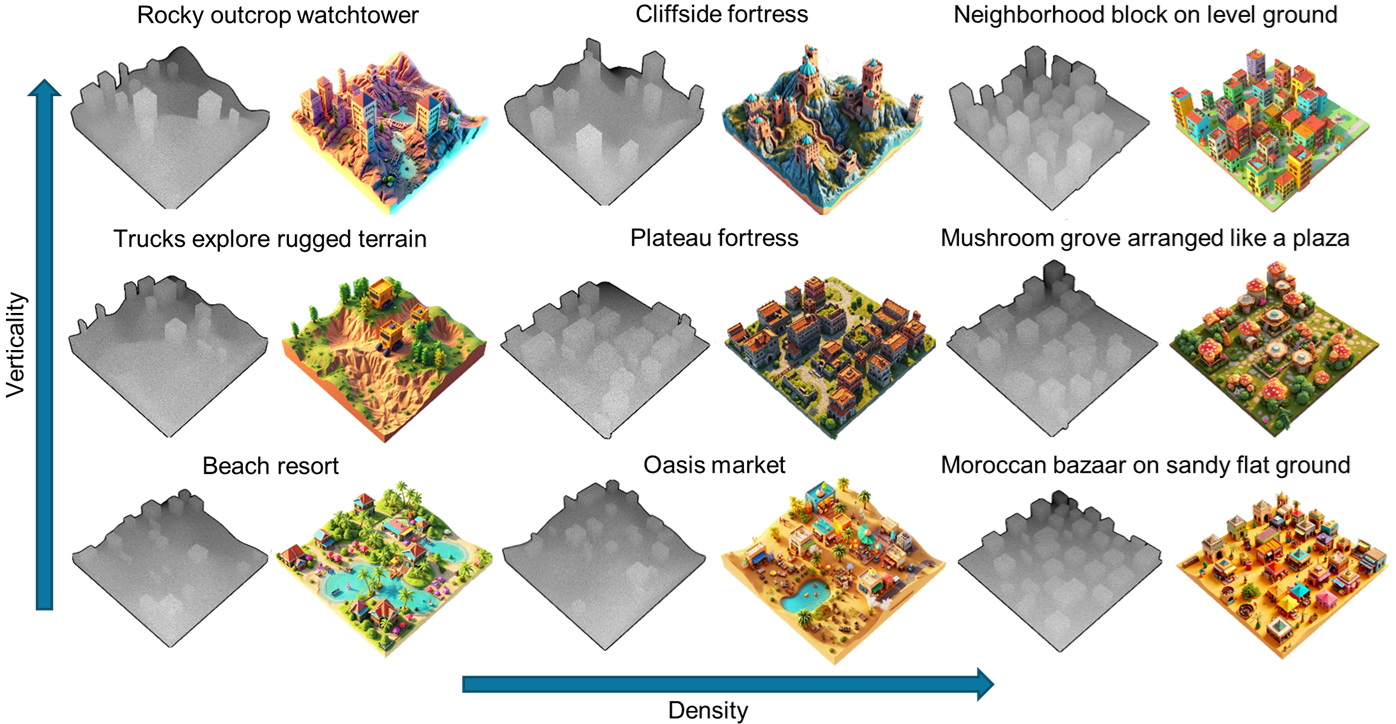}
\caption{
\textbf{Depth-conditioned generation across density (columns) and verticality range (rows).} In each grid cell, we show the input depth map (left) and the corresponding generated image conditioned on that depth (right). Columns are ordered by increasing density from left to right; rows are ordered by increasing verticality range from low to high.
}%
\label{fig:layout_results}
\end{figure}

In \Cref{fig:layout_results} we show several representative examples generated by our text-conditioned layout module.
Each example illustrates different combinations of terrain types, verticality and object density—the primary factors governing scene structure and downstream difficulty. The explicit procedural generation process guarantees that the entire area is navigable.

The first column depicts low density layouts with various terrain types, producing an open and easily traversable layout.
The second column introduces a medium density object placement with diverse verticality changes, leading to a richer spatial composition that includes both open grounds and dense activity areas.
The third column features a dense asset distribution, which results in a complex environment.

\section{Stage II: Scene Reconstruction}
\label{sec:scene-reconstruction}

Given the scene plan $\plan=(\blockout, \referenceImage, \nav)$ obtained in Stage~I, our goal in this stage is to generate a 3D scene mesh $\sceneMesh$ that faithfully aligns with the plan.
In particular, the mesh should respect the navigable regions encoded in the \navmesh $\nav$, while also matching the overall composition and appearance suggested by the reference image $\referenceImage$.

Our pipeline learns this mapping through triplets $(\sceneMesh, \referenceImage, \nav)$: the ground-truth scene mesh and the corresponding reference image and \navmesh.
Unfortunately, large-scale datasets that contain such detailed scene-level supervision are scarce.
To address this limitation, we adopt a simple yet effective two-stage training strategy.
First, we pre-train an image-to-3D model on a broad set of generic object categories,
allowing the model to learn a robust prior for projecting 2D visual cues into 3D geometry and texture space.
We then fine-tune the model on our curated dataset of scene triplets, where the generation is conditioned on both the reference image $\referenceImage$ and the \navmesh $\nav$.

\begin{figure}[ht]
\centering
\begin{tabular}[b]{cc}
\raisebox{0.0\height}{\includegraphics[width=0.49\linewidth]{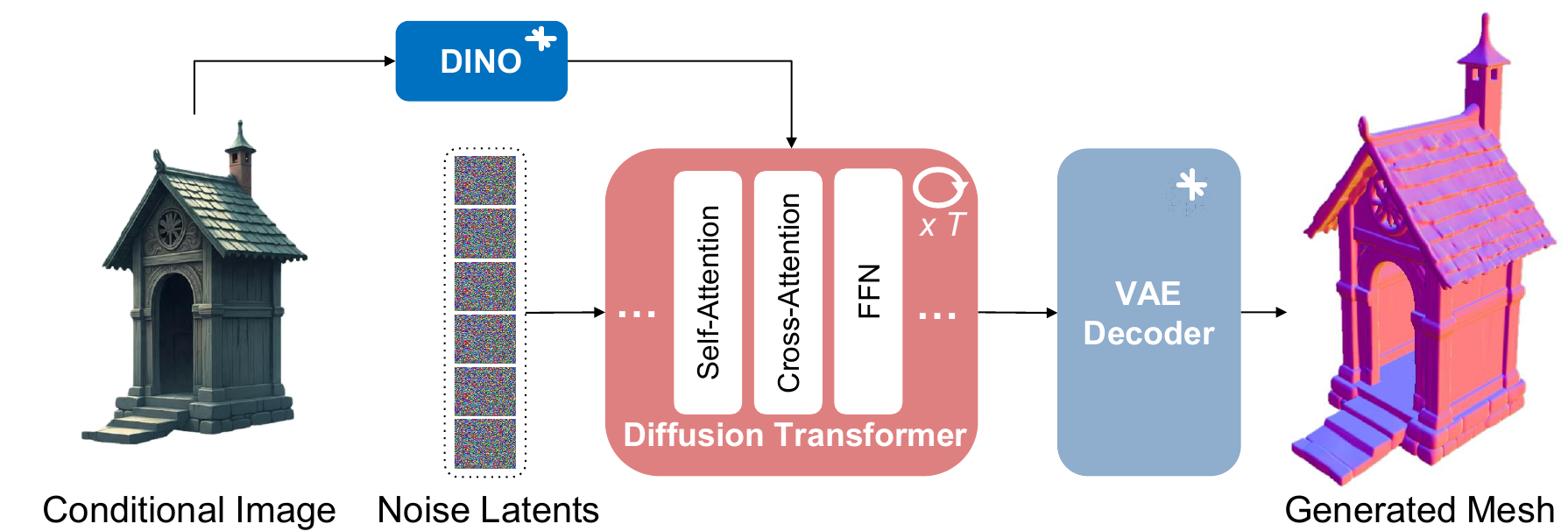}} &
\includegraphics[width=0.49\linewidth]{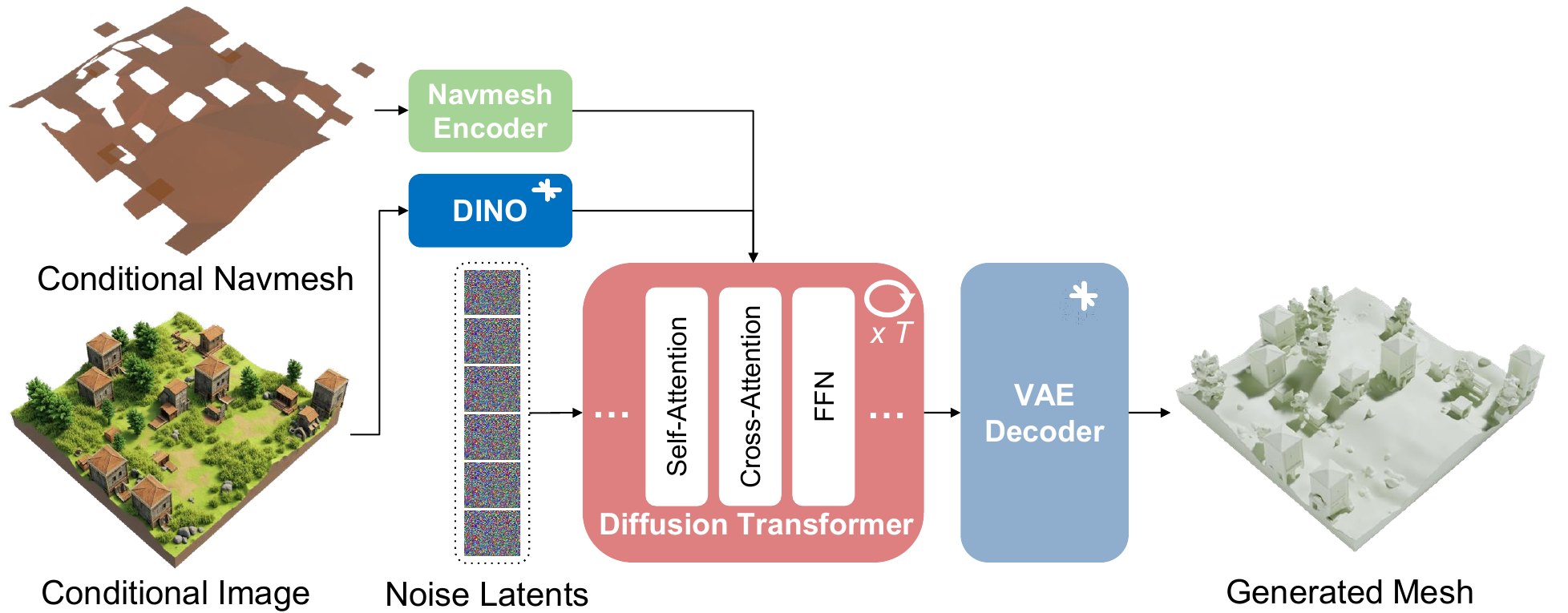}  \\
(a) \assetgen  architecture & (b) Navmesh conditioning  \\
\end{tabular}
\caption{\textbf{AssetGen2 and Navmesh architectures.} Left: Overview of the base AssetGen mesh generation architecture. Right: Our Navmesh conditioned scene mesh generation (Stage II) based on cross-attention}
\label{fig:meshgen_navmesh_arch}
\end{figure}

\subsection{Image-to-3D Base Model}%
\label{sec:assetgen}

We use \assetgen as our base model for image-to-3D generation.
The shape generator in \assetgen adopts the popular \emph{VecSet}~\citep{zhang233dshape2vecset:} representation for 3D diffusion modeling,
where a scene or object is represented as an unordered set of latent vectors.
The diffusion model learns to denoise these vector sets to reconstruct signed distance fields (SDFs) conditioned on the input image.

\newcommand{\cE}{\mathcal{E}}
\newcommand{\cD}{\mathcal{D}}
\newcommand{\cP}{\mathcal{P}}
\newcommand{\pointCloudDense}{\cP}
\newcommand{\pointCloudSparse}{\hat{\cP}}
\newcommand{\encoder}{\cE}
\newcommand{\encoderNavmesh}{\cE'}
\newcommand{\decoder}{\cD}

\paragraph{VecSet Latent Representation.}

VecSet learns a 3D latent space for compact object representation using an autoencoder.
Given a 3D object $\xfine$ represented by a point cloud
$\pointCloudDense = \{(p_i, n_i)\}_{i=1}^M$, with points $p_i \in \mathbb{R}^3$ and normals $n_i \in \mathbb{S}^2$,
the encoder $\encoder$ maps $\pointCloudDense$ to a latent code
$\z \in \mathbb{R}^{K \times D}$ consisting of $K$ $D$-dimensional tokens:
$
\z = \encoder(\pointCloudDense).
$
The decoder $\cD$ reconstructs the signed distance function (SDF) of the object,
assigning each query location $q \in \mathbb{R}^3$ with an SDF value
$
\decoder(q \mid \z) \in \mathbb{R}
$.
The encoder starts by randomly downsampling the input point cloud to $K$ points, one per token, extracting a subset
$
\pointCloudSparse = \operatorname{FPS}(\pointCloudDense \mid K) = \{\hat p_1,\dots, \hat p_K \}
$
using farthest point sampling (FPS).
The encoder then projects the full point clouds (with normals) $\pointCloudDense$ to the selected points $\pointCloudSparse$ using sinusoidal spatial encoding followed by cross-attention and several standard transformer layers until the final code $\z$ is obtained.
The decoder operates in `reverse', taking as input a query point $q$ in order to compute the corresponding SDF value.
The resulting latent tokens form a compact, permutation-invariant 3D representation suitable for diffusion-based generation.

\paragraph{Image-to-3D Latent Diffusion Model.}

\assetgen learns a diffusion model that generates 3D object latents conditioned on an input image $\objectImageFine$.
Specifically, it models the conditional distribution of latent codes as
$
p(\z \mid \objectImageFine; \Phi)
$
and learns it via a denoising diffusion process parameterized by a transformer $\Phi$.
The resulting latent $\z$ defines a signed distance field (SDF), from which a watertight triangular mesh is extracted using Marching Cubes~\citep{lorensen87marching}.
An overview of the model architecture is shown in \Cref{fig:meshgen_navmesh_arch}(a). 

\paragraph{Training.}

We train the \vecset autoencoder and the image-to-3D diffusion model in \assetgen using am internal dataset of artist-authored 3D assets.

\subsection{Navmesh-Based Scene Generation}%
\label{sec:navmesh-scene-generation}

While \assetgen samples $p(\z \mid \referenceImage;\Phi)$ to obtain a (latent) 3D shape $\z$ from the input image $\referenceImage$, we wish the scene reconstruction to be also compatible with the \navmesh $\nav$.
Although the image $\referenceImage$ already reflects the structure of the \navmesh to a large degree, it may not do so perfectly; furthermore, the image does not show the scene in full due to self-occlusions, so the \navmesh is not fully represented by it.
Hence, we modify \assetgen to sample from the distribution
$
p(\z \mid \referenceImage, \nav; \Phi)
$
to explicitly account for both reference image and \navmesh.
This provides structural guidance that preserves traversability and spatial coherence during scene generation.

We fine-tune \assetgen for \navmesh-conditioned scene generation using the architecture illustrated in \Cref{fig:meshgen_navmesh_arch}(b). 
The design follows the VecSet paradigm given in \Cref{sec:assetgen}, but modifies the transformer to attend both the input image $\referenceImage$ and the \navmesh $\nav$ via cross attention.
The latter requires to encode the \navmesh as a set of tokens, which we detail next.

\paragraph{Navmesh Encoder and Conditioning.}

To tokenize the navmesh $\nav$, we use an encoder $\encoderNavmesh$ similar to the VecSet one $\encoder$ descried in \cref{sec:assetgen}.
First, we sample uniformly at random points from the surface of the \navmesh $S$ to form a point cloud $\pointCloudDense \in \mathbb{R}^{M \times 3}$ and downsample it to a smaller size $\pointCloudSparse = \operatorname{FPS}(\pointCloudDense \mid K) \in \mathbb{R}^{K \times 3}$ using farthest point sampling.
Both point sets are independently embedded using a coordinate positional encoder that maps 3D coordinates into $D$-dimensional feature vectors.
Then, the encoded sparse points attend to the dense points through cross-attention to capture fine-grained geometric details of the \navmesh.
The \navmesh encoder thus differs from the \vecset one in that it does not use point normals and there are no additional transformer layers, which saves considerable memory.
Also the positional encoding of the sparse points are added to the output of the cross-attention layer to reinforce the location of those points in the representation.
The resulting sparse \navmesh embeddings $\encoderNavmesh(S)$ are integrated into the \assetgen denoising diffusion transformer backbone through additional cross-attention layers.

\paragraph{Training Strategy.}

To train the diffusion model $p(\z \mid \referenceImage, \nav;\Phi)$, we start from the pretrained \assetgen weights described in \Cref{sec:assetgen} and fine-tune them using our data. We compare updating only the weights of the newly-added cross-attention layers, or the entire transformer model, including the pre-trained weights, end-to-end.
Empirically, the latter yields consistently lower validation loss compared to conditioning-only training.
This suggests that producing a mesh that aligns faithfully with the \navmesh requires joint adaptation across the entire generator network,
as the task involves non-trivial geometric alignment and scene-level completion beyond the capacity of the newly introduced conditioning layers alone.

\paragraph{Data Normalization.}
\assetgen\ operates in a normalized space, where all meshes are rescaled within a $[-1,1]^3$ cube.  
During training, we rescale each \navmesh using the scale factor computed from its corresponding scene mesh to ensure spatial alignment.
We also jointly translate the \navmesh and the scene mesh so that the \navmesh ground plane is centered at $(0,0,0)$. 
This trick provides a stable spatial reference, leading to better alignment between the conditioned \navmesh and the generated scene meshes.

At inference time, where ground-truth scene meshes are unavailable, we normalize the \navmesh using the scale derived from the procedurally generated blockout $B$  and apply the translation to the \navmesh.

\paragraph{Re-extraction of the NavMesh.}

Once the scene mesh $\sceneMesh$ is generated, we re-extract the \navmesh $\nav'$ from it using the same algorithm as in \Cref{sec:navmesh-reference-image-generation} to ensure compatibility with the generated geometry.

\subsection{Scene Texture Generation}%
\label{sec:scene-texture-generation}

At this point, we have generated a scene mesh $\sceneMesh$ that aligns well with both the reference image $\referenceImage$ and the \navmesh $\nav$, but the scene mesh is texture-less.
While we do have a powerful texture generator, further described in \cref{sec:object-mesh-enhancement}, it is based on multi-view image generation and texture reprojection.
Because scene meshes contain plenty of self-occlusions due to their complex and packed shapes, the latter may leave significant chunks of the geometry `unpainted'.

We thus leverage the \trellis~\citep{xiang24structured} texture generator to output a texture for the whole mesh $M$.
Although \trellis typically produces lower-resolution textures compared to multi-view generation based methods, it produces textures in 3D directly, so it is not affected by self-occlusions as much.
The texture does not need to be very high-quality; instead, it only needs to provide sufficient guidance for the scene enhancement stage, where a high-quality texture is obtained on an object-by-object basis (\Cref{sec:object-mesh-enhancement}).

Since \trellis is trained primarily on object-level 3D data and generalizes less effectively to full scenes, we re-implement and retrain this model on our in-house dataset containing both object- and scene-level assets.

\begin{figure}[tbp]
\centering
\includegraphics[width=1.0\linewidth]{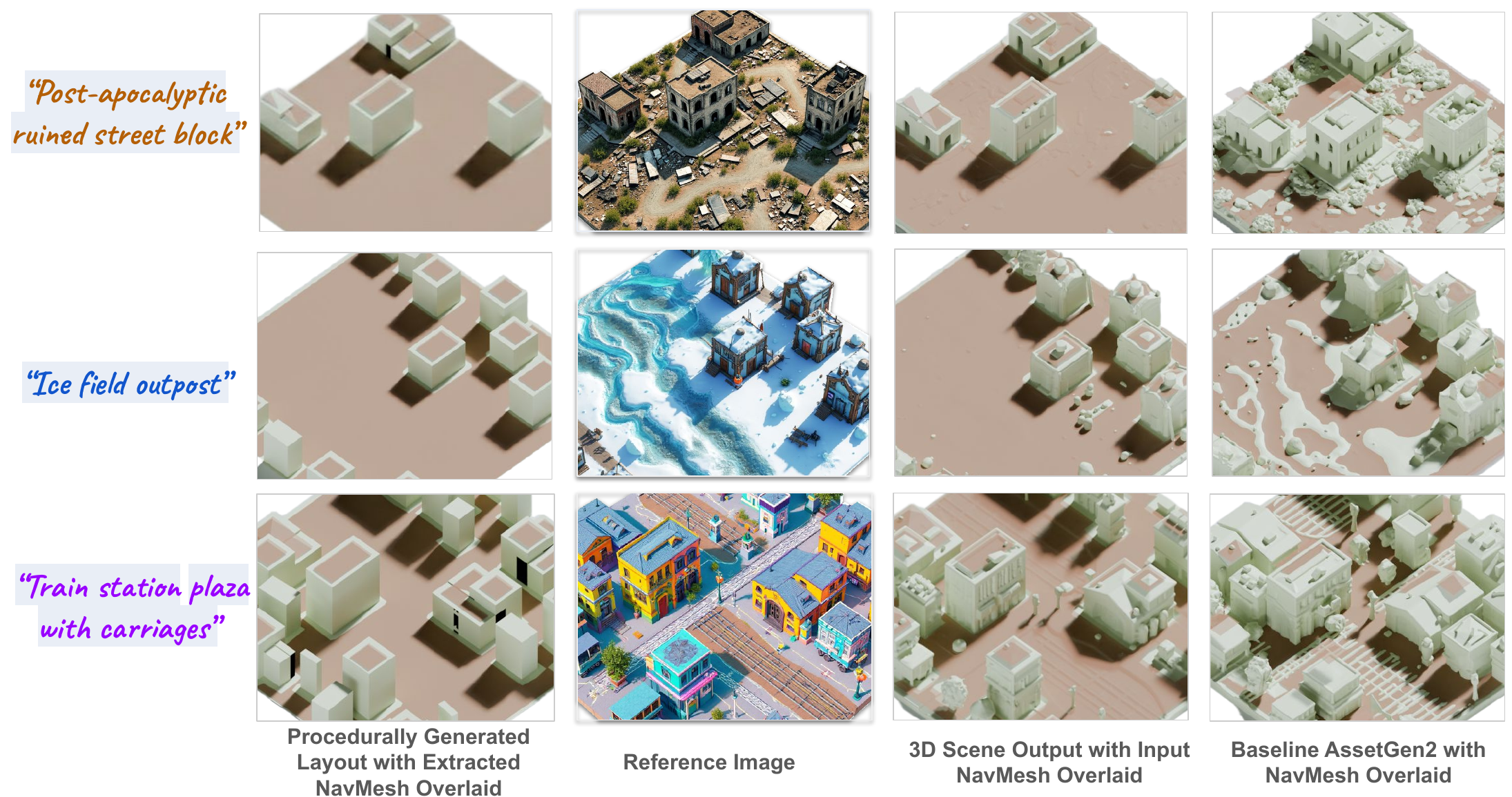}
\caption{ 
\textbf{Navmesh-conditioned scene generation}. Left to right: the procedurally generated 3D layout and the extracted \navmesh (used as input condition for 3D scene generation), the generated reference image conditioned on the 3D layout, the input \navmesh overlaid on our final generated scene, produced by the navmesh-conditioned model and the baseline \assetgen, respectively.
This confirms our generated scene successfully adheres to the specified navigable path.
}%
\label{fig:navmesh_mesh_gen}
\end{figure}

\begin{figure}[htbp]
\centering
\includegraphics[width=1.0\linewidth]{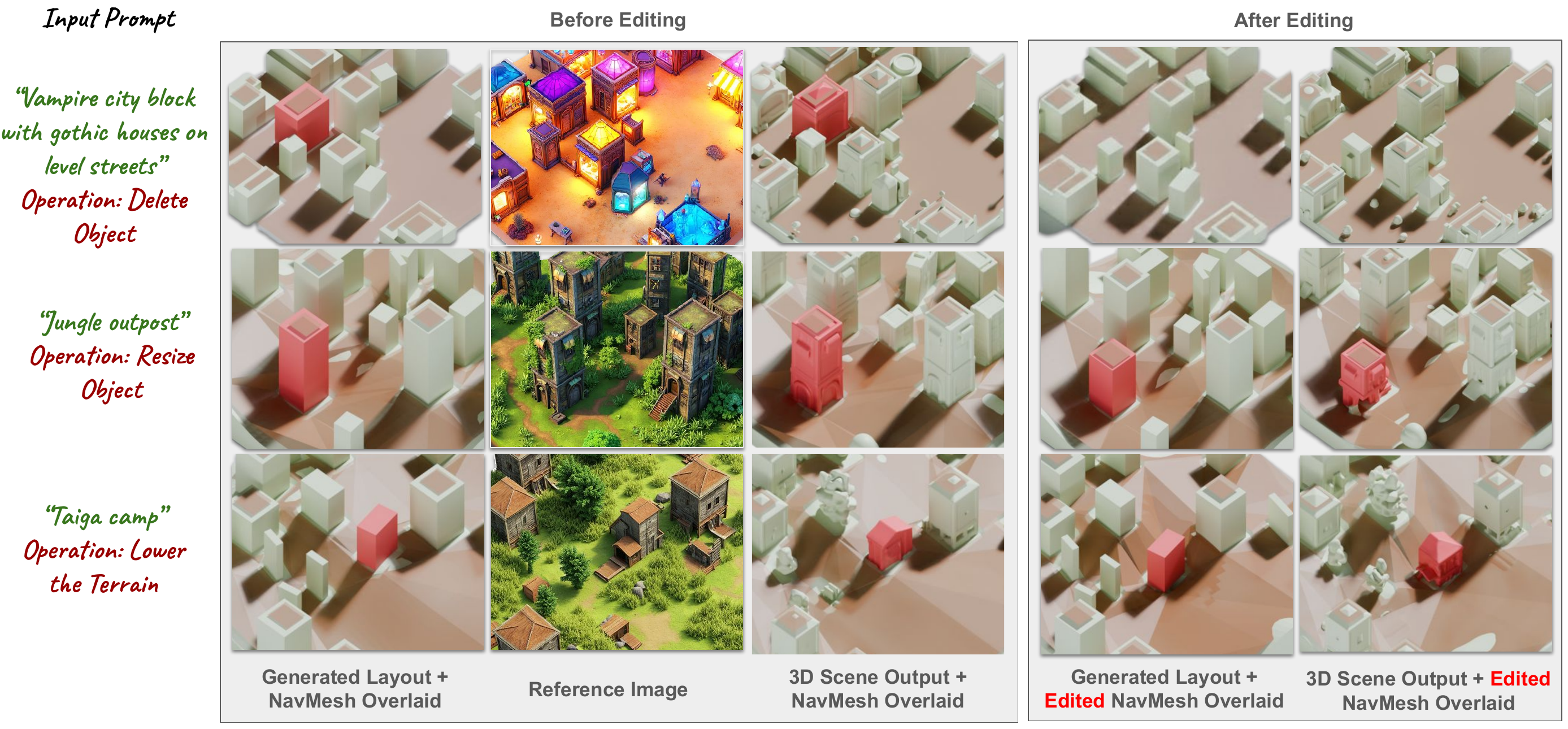}
\caption{\textbf{Layout editing.}
Our \navmesh-conditioned scene generation allows input editing to the procedurally generated layout and the corresponding navigable path.
For each row we show the generation results with manual editing on the initial procedurally generated layout.
For each column from left to right: procedurally generated scene layout with extracted \navmesh overlaid, reference image, 3D scene generation output with input \navmesh condition overlaid, procedurally generated scene layout with manual edits, 3D scene generation output with edited \navmesh condition overlaid.}%
\label{fig:navmesh_generalize}
\end{figure}

\subsection{Scene Reconstruction Results}%
\label{sec:scene-reconstruction-results}

In \Cref{fig:navmesh_mesh_gen}, we show the procedurally generated layout (with overlaid \navmesh), the reference image, our \navmesh-conditioned scene generation, and the baseline \assetgen results conditioned only on the reference image.
our \navmesh-conditioned model produces smoother and cleaner terrain surfaces that adhere closely to the input navigation constraints---key for ensuring smooth traversability in the final scene.
Note that some fine terrain details may be visually approximated through texturing, accurate geometric grounding is critical for gameplay functionality.
Moreover, the \navmesh-conditioned model achieves better alignment with the reference image in structural regions such as buildings, particularly in more complex scenes.

To quantify this improvement, we compare our finetuned model with baseline \assetgen, baseline \assetgen trained on our curated dataset of scene triplets (denoted as Baseline*),  as well as a \sanjose on a curated scene benchmark, detailed in \Cref{tab:navmesh-ablation}. 
This benchmark comprises 50 procedurally generated scenes, each featuring terrain with moderate verticality and 10-30 densely sampled objects. After normalizing all geometries to a $[-1,1]^3$ cube, we extract the \navmesh and align it to the ground-truth \navmesh using ICP\@. We then compute the Chamfer Distance between these aligned meshes. Our \navmesh-conditioned model achieves Chamfer distances that are 40--50\% lower compared to the baselines.
The quantitative comparisons clearly demonstrate stronger spatial alignment with the input navigation conditions.

\begin{table*}[t]
  \centering
  \caption{\textbf{Quantitative evaluation of \navmesh alignment}, measured via Chamfer distance (CD) between the input \navmesh and the \navmesh extracted from the generated scene (lower is better). Baseline denotes the baseline \assetgen, and Baseline* denotes the baseline \assetgen trained on our curated dataset of scene triplets.}
  \label{tab:navmesh-ablation}
  \renewcommand\tabcolsep{10pt}

  {\small
  \begin{tabular}{lcccc}
    \toprule
    Model &  {\sanjose} & {Baseline} & {Baseline*} & {\textbf{Ours}} \\
    \midrule
    {NavMesh CD} & 0.038 & 0.042 & 0.038 & \textbf{0.022}\\
    \bottomrule
  \end{tabular}
  }
\end{table*}

In \Cref{fig:navmesh_generalize}, we further evaluate the model's generalization under slight misalignments between the \navmesh and the reference image.
This scenario is particularly important, as editing a 2D reference image to accurately reflect spatial intent is often cumbersome---or even infeasible---due to inherent ambiguities in projecting 3D structure onto a single view.
In contrast, \navmesh edits can be easily and precisely specified in 3D by modifying the layout directly.
For example, in the ``Jungle outpost'' scene, we remove a structure to create additional walkable space; in ``Vampire city block'', we reduce the height of a building; and in ``Taiga camp'', we slightly lower the terrain to form a shallow concave dip, adding subtle variation to the landscape.
These examples illustrate that our \navmesh-conditioned model does not merely replicate the reference image but instead learns to reason about spatial organization and navigability, maintaining structural and stylistic coherence even when the layout and image conditions diverge.

\section{Stage III: Scene Decomposition}
\label{sec:scene-decomposition}

The coarse, monolithic textured mesh $\sceneMesh$ produced by Stage~II represents the entire scene and fuses all objects into a single geometry.
This makes it difficult to edit or refine individual assets.
To address this limitation, we first decompose $\sceneMesh$ into semantically meaningful objects and parts, and then enhance them individually in the subsequent stage.

Our approach builds upon our recent \autopartgen~\citep{chen25autopartgen} model,
which decomposes a mesh into parts sequentially in an autoregressive manner,
where each part is generated conditioned on the holistic scene mesh and all previously generated parts.
However, two key limitations make the original \autopartgen unsuitable for large-scale scene decomposition.
First, its autoregressive nature leads to slow inference, making it computationally expensive for complex scenes with many objects and parts.
Second, the model was trained primarily on generic object-level datasets and therefore fails to generalize effectively to full scenes containing diverse, spatially interacting assets.

\subsection{Accelerating AutoPartGen for Scenes}%
\label{sec:scene-autopartgen}

To accelerate part extraction, we draw inspiration from PartPacker~\citep{tang25efficient}.
While \autopartgen generates parts in a fixed lexicographical order (z-x-y), we instead generate parts according to their \emph{connectivity degree}, defined as the number of other parts each collides with.
By generating parts in decreasing order of connectivity, we prioritize \emph{pivot parts} that serve as structural anchors connecting many other components.
Once pivots are generated, the remaining parts can be efficiently recovered through spatial connectivity analysis of the residual geometry.
For example, in outdoor scenes with many buildings and trees, the ground often exhibits the highest connectivity degree. Once the ground is extracted,
the decomposition of other objects becomes easier through connected-component analysis.

To support this strategy, we extend \autopartgen to explicitly generate the \emph{remainder geometry} as a special part.
We introduce a binary \emph{flag token} that, when activated, instructs the model to produce all remaining geometry in a single forward pass.
In practice, we use a five-step schedule: the model first generates four pivot parts, followed by the remainder part, which is further decomposed via connected-component analysis.
This design accelerates decomposition significantly, even for complex scenes, and reduces overall generation time from ten to about one minute.

\subsection{Scene Decomposition Data}%
\label{sec:scene-decomposition-data}

We find it important to finetune \autopartgen on scene-level data.
However, there is no readily available dataset of 3D scenes with part annotations.
We address this gap by creating a dataset of compositional 3D scene.

First, we mine 3D scenes from a large internal 3D asset repository.
To do this, we use a vision-language model (VLM) and identify assets that represent whole scenes.
Specifically, we prompt the VLM to assess rendered images and determine whether they exhibit the characteristics of multi-object environments
(e.g., sufficient object diversity, plausible spatial layout, and visible ground context).

Once initial scene-like assets are selected, we apply heuristics to convert the raw geometry into assets with meaningful object and part decompositions.
Our processing pipeline combines connectivity-based part splitting with ground-aware reasoning. The pipeline includes four major steps:
(1) we detect topologically connected components after vertex welding as minimal parts,
(2) we detect the ground and merge thin overlays (e.g., traffic lines) onto the ground as a standalone part,
(3) we de-duplicate and iteratively merge small parts into their nearest spatial neighbors while ensuring the ground remains separate, and
(4) we filter decomposed assets with several quality constraints, including part count, part imbalance, and ground confidence.

\begin{table*}[t]
  \centering
  \caption{\textbf{Quantitative evaluation of scene decomposition methods.} Our model achieves significantly better results than previous state-of-the-art methods across all metrics.}
  \label{tab:partgen-ablation}
  \begin{tabular}{lcccccr}
    \toprule
    \textbf{Model} & \textbf{Chamfer} & \textbf{F-score@0.01} & \textbf{F-score@0.02} & \textbf{F-score@0.03} & \textbf{F-score@0.05} & \textbf{Time} \\
    \midrule
    \partpacker    & 0.171 & 0.090 & 0.215 & 0.307 & 0.443 & 1 min \\
    \bigsurpart     & 0.136 & 0.155 & 0.357 & 0.481 & 0.633 & 3 min \\
    AutoPartGen & 0.144 & 0.281 & 0.526 & 0.613 & 0.683 & 10 min \\
    \textbf{Ours}  & \textbf{0.061} & \textbf{0.322} & \textbf{0.644} & \textbf{0.761} & \textbf{0.853} & \textbf{1 min} \\
    \bottomrule
  \end{tabular}
\end{table*}
\begin{figure}[t]
\centering
\includegraphics[width=1.0\linewidth]{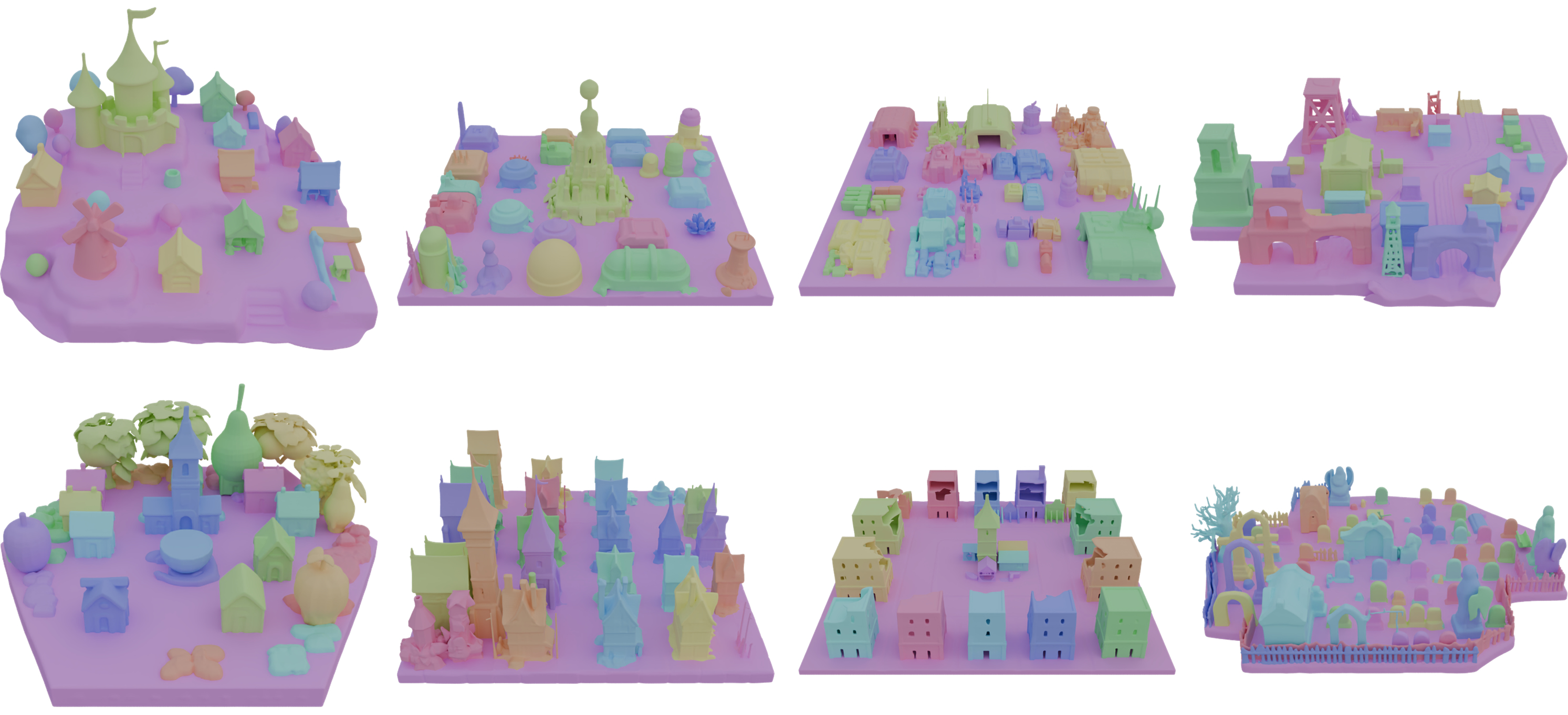}
\caption{\textbf{Decomposition results.} Our model is finetuned on scene data with part annotations such that given an input 3D scene, the model successfully decomposes it into its constituents, starting from the ground.}
\label{fig:partgen_resutls}
\end{figure}

\subsection{Decomposition Results}

To evaluate the decomposition capability of our model, we perform a quantitative comparison between our method and other state-of-the-art methods. We curate a diverse synthetic evaluation dataset by placing objects on various terrains. The holistic mesh is obtained by watertighting all objects and terrains jointly. This provides us pairs of holistic meshes and groundtruth part annotations which we can benchmark on. Our evaluation protocol follows prior work~\citep{chen25autopartgen,liu2025partfield}: for each groundtruth part, we find its nearest neighbor in predictions and compute their Chamfer Distance and F-score at various thresholds. We show results in Table~\ref{tab:partgen-ablation}, where our model clearly outperforms other state-of-the-art methods, while maintaining the top inference speed.

We also qualitatively demonstrate the effectiveness of our model for scene decomposition across a variety of environments,
including both flat terrains and scenes with mild elevation variations (see Figure~\ref{fig:partgen_resutls}).
Our method performs robustly across these settings:
the ground terrain is cleanly separated from the overall scene mesh,
and objects are segmented into semantically meaningful components that facilitate subsequent enhancement and texturing.

\begin{figure}[t]
\centering
\includegraphics[width=0.99\linewidth]{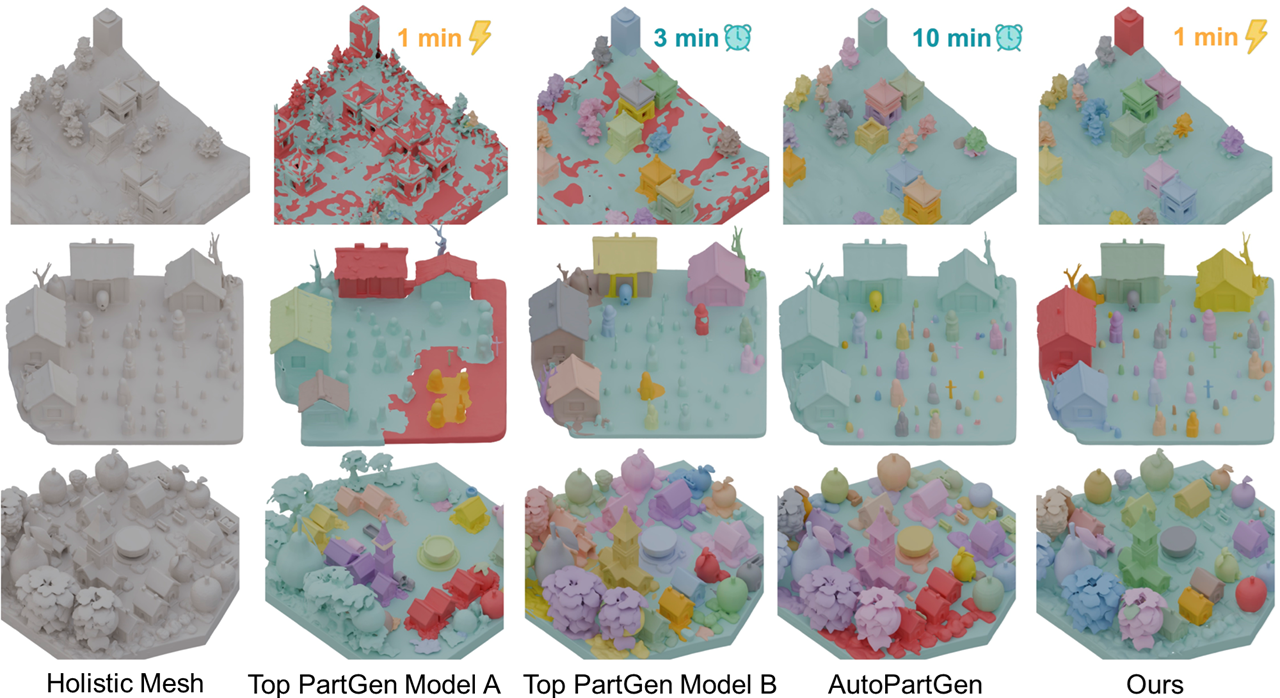}
\caption{ 
\textbf{Scene decomposition comparison}. Our model demonstrates decomposition results of the highest quality and the least noise compared to other state-of-the-art methods, yet maintaining a fast inference speed.}%
\label{fig:partgen_ablate}
\end{figure}

This capability is rarely achieved by existing part segmentation models, which often fail to generalize effectively to complex scene-level inputs (see Figure~\ref{fig:partgen_ablate}).
For instance, \partpacker struggles to generalize and produces unstable decompositions for large outdoor scenes.
\bigsurpart often erroneously segments terrain regions, sometimes fragmenting buildings into excessively small components or merging ground and object geometry into a single part.
The original \textsc{AutoPartGen}~\citep{chen25autopartgen} sometimes fails to decompose major objects, and its high latency makes it impractical for handling large scenes.

\newcommand{\zfine}{\z}
\newcommand{\zcoarse}{\hat{\z}}

\begin{figure}[t]
\centering
\includegraphics[width=0.99\linewidth]{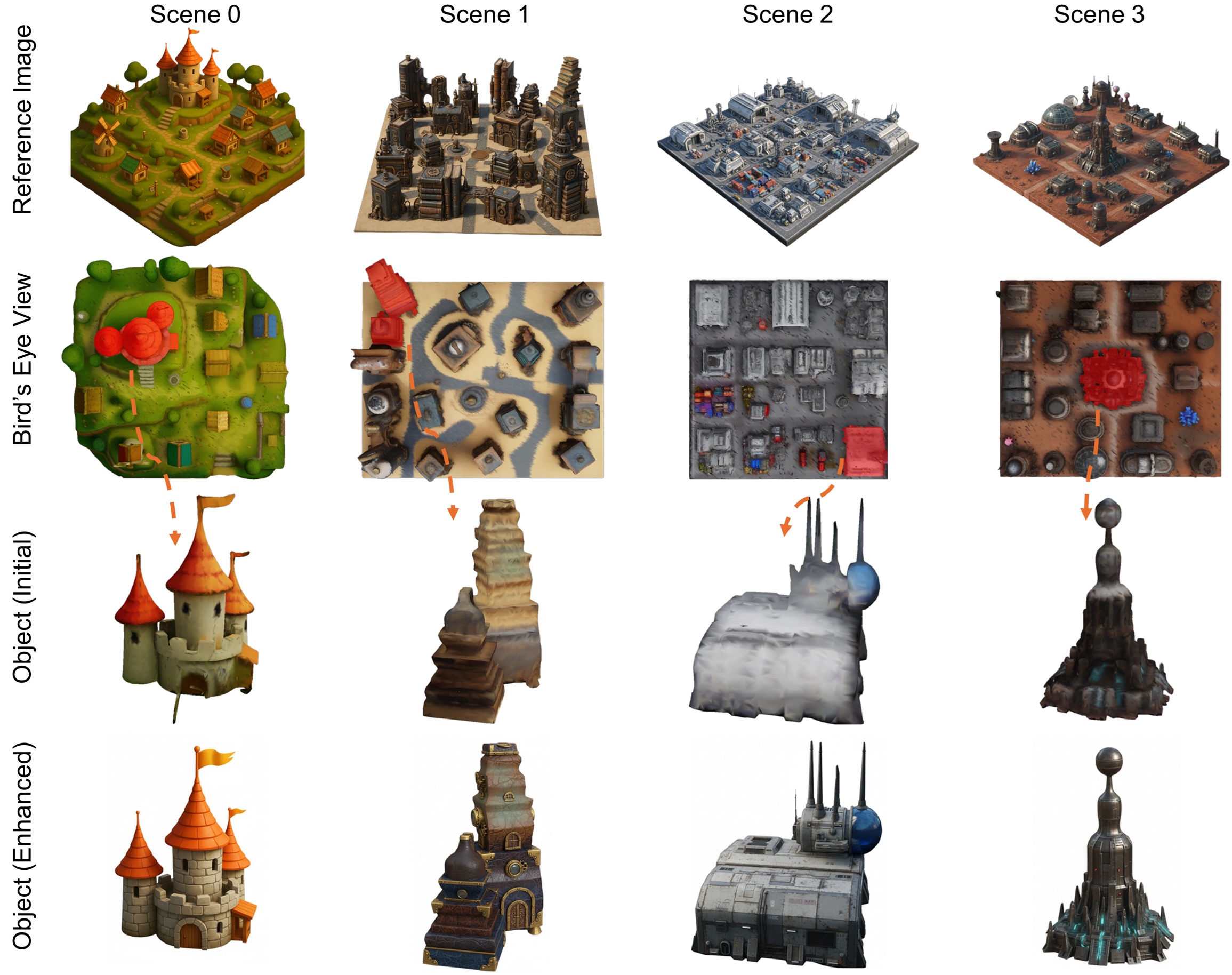}
\caption{
\textbf{Per-object image enhancement.} 
The reference image, the top-down view with target object highlighted with red and the coarse textured rendering are sent to an LLM-VLM that outputs the final per-object enhanced image.
}%
\vspace{-0.15cm}
\label{fig:object-image-enhancement}
\end{figure}

\begin{figure}[t]
\centering
\includegraphics[width=0.95\linewidth]{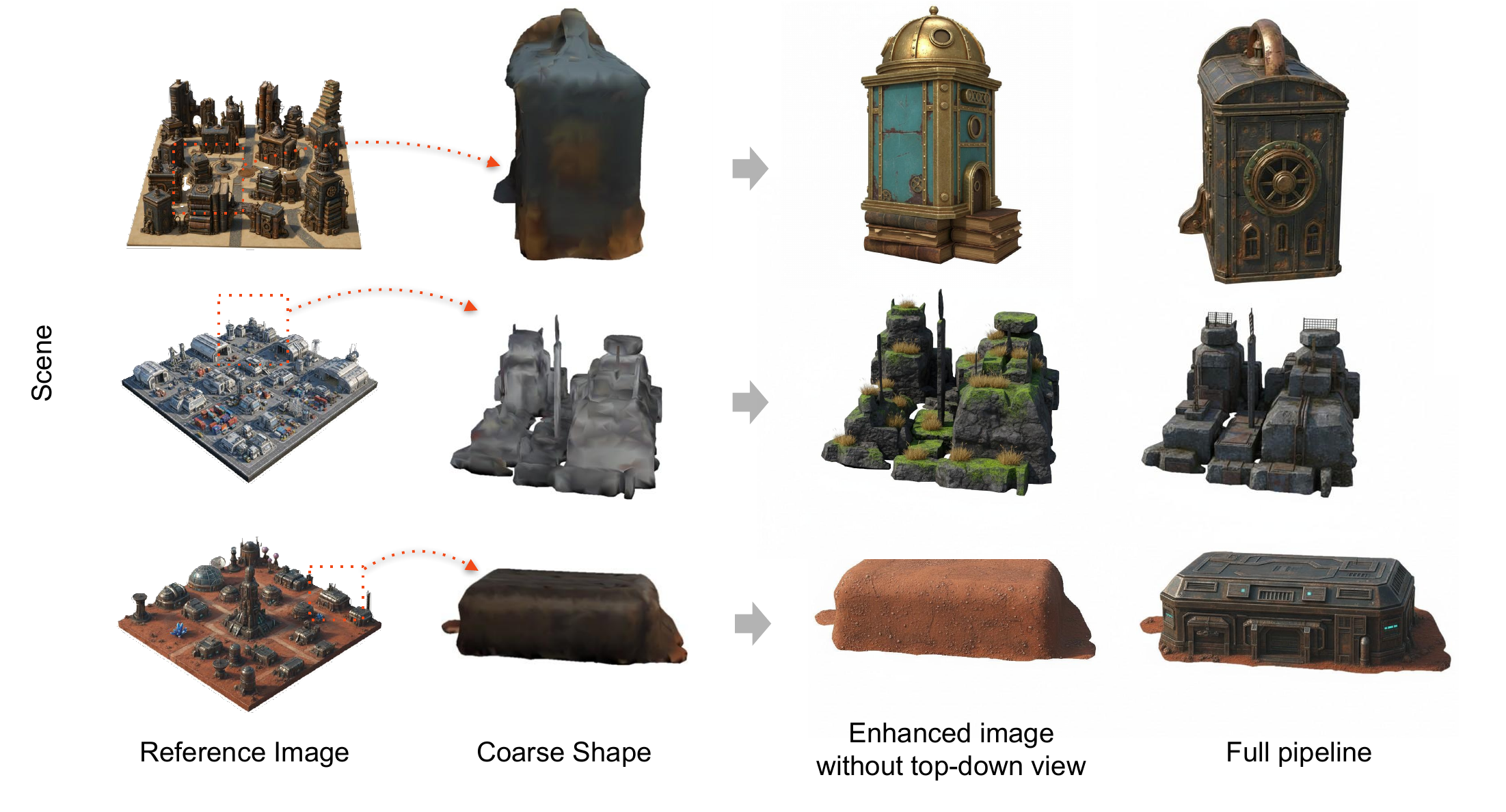}
\caption{
\textbf{Per-object image enhancement without using the top-down view.} 
Without access to a top-down view of the entire scene—which gives the LLM-VLM important information about object location and surrounding context—the model has difficulty generating object images that are visually consistent with the scene’s style or faithful to the reference image. As a result, the generated images may not match the overall style of the scene or may differ from the appearance of the object in the reference image.
}%
\vspace{-0.15cm}
\label{fig:object-image-enhancement-ablation}
\end{figure}

\begin{figure}[t]
\centering
\includegraphics[width=0.95\linewidth]{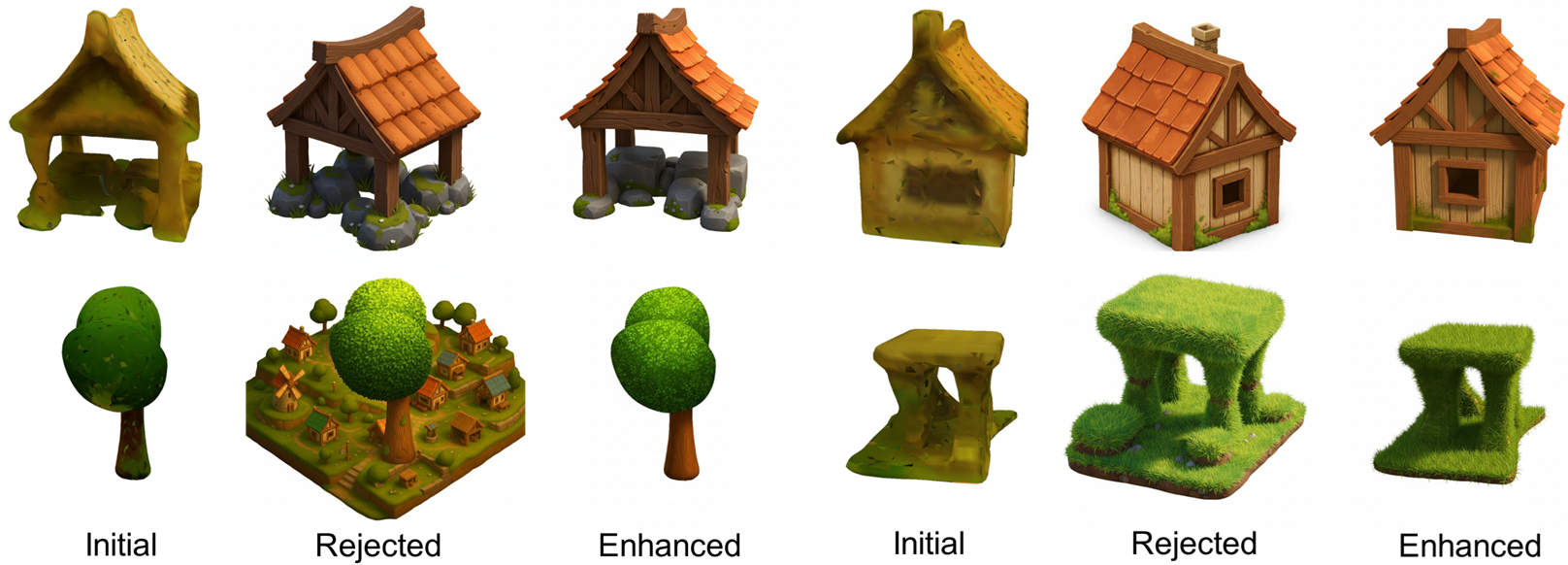}
\caption{\textbf{Object image verification.}
This stage may require multiple iterations to achieve the desired visual quality, followed by an automatic verification step.
Here, “\textit{initial}” denotes the low-resolution input render, “\textit{rejected}” refers to enhanced images rejected by the verification step, and “\textit{enhanced}” represents the final accepted results.
Common failure cases include changes in object orientation, omission or hallucination of geometric details, and incorrect overlaying of objects onto the background scene.}%
\label{fig:image_verification}
\end{figure}

\section{Stage IV: Scene Enhancement}
\label{sec:scene-enhancement}

The goal of this stage is to enhance the visual and geometric quality of individual objects and parts generated in the previous stage, while ensuring their consistency and seamless composability within the overall scene.
This process comprises three main components.
First, since the resolution and coverage of the reference scene image $\referenceImage$ is limited, 
we \emph{generate a new high-quality image} for each object.
We design a generator to take as input a render $\objectImageCoarse$ of the object $\xcoarse_i$ and output a new image $\objectImageFine_i$ 
that is more correct and detailed.
Second, we \emph{regenerate the shape of each object} $\xcoarse_i$ based on the new image $\objectImageFine_i$ and the coarse geometry of $\xcoarse_i$.
We do so via a new mesh enhancer which, guided by $\objectImageFine_i$, improves the shape without departing too much from the original.
Finally, we \emph{generate a high-quality texture} for each object, guided by the refined geometry and visual cues from the enhanced object image $\objectImageFine_i$ (\Cref{sec:object-mesh-enhancement}).
Thus, the output of Stage~IV is a new, improved version $\xfine_i$ of each object $\xcoarse_i$ from Stage~III\@.

\begin{figure}[t]
\centering
\includegraphics[width=0.99\linewidth]{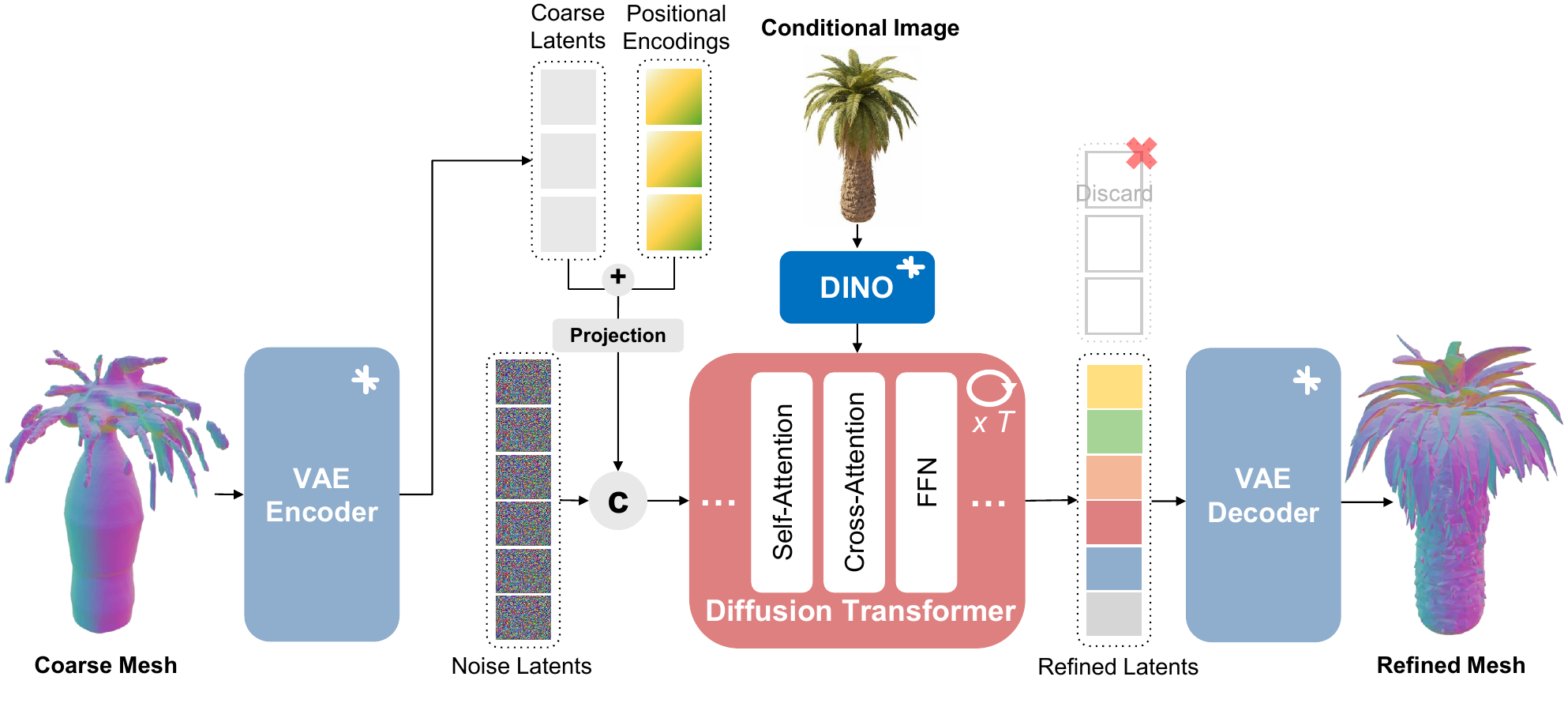}
\caption{\textbf{Per-object mesh refinement}. Given a coarse object mesh and a high-resolution image, we feed them to our mesh refinement model which outputs a refined high-quality mesh that adheres to the orientation and shape of the coarse input yet incorporates fine details from the image input.}%
\label{fig:partrefiner_arch}
\end{figure}
 
\begin{figure}[t]
    \centering
    \includegraphics[width=0.95\linewidth]{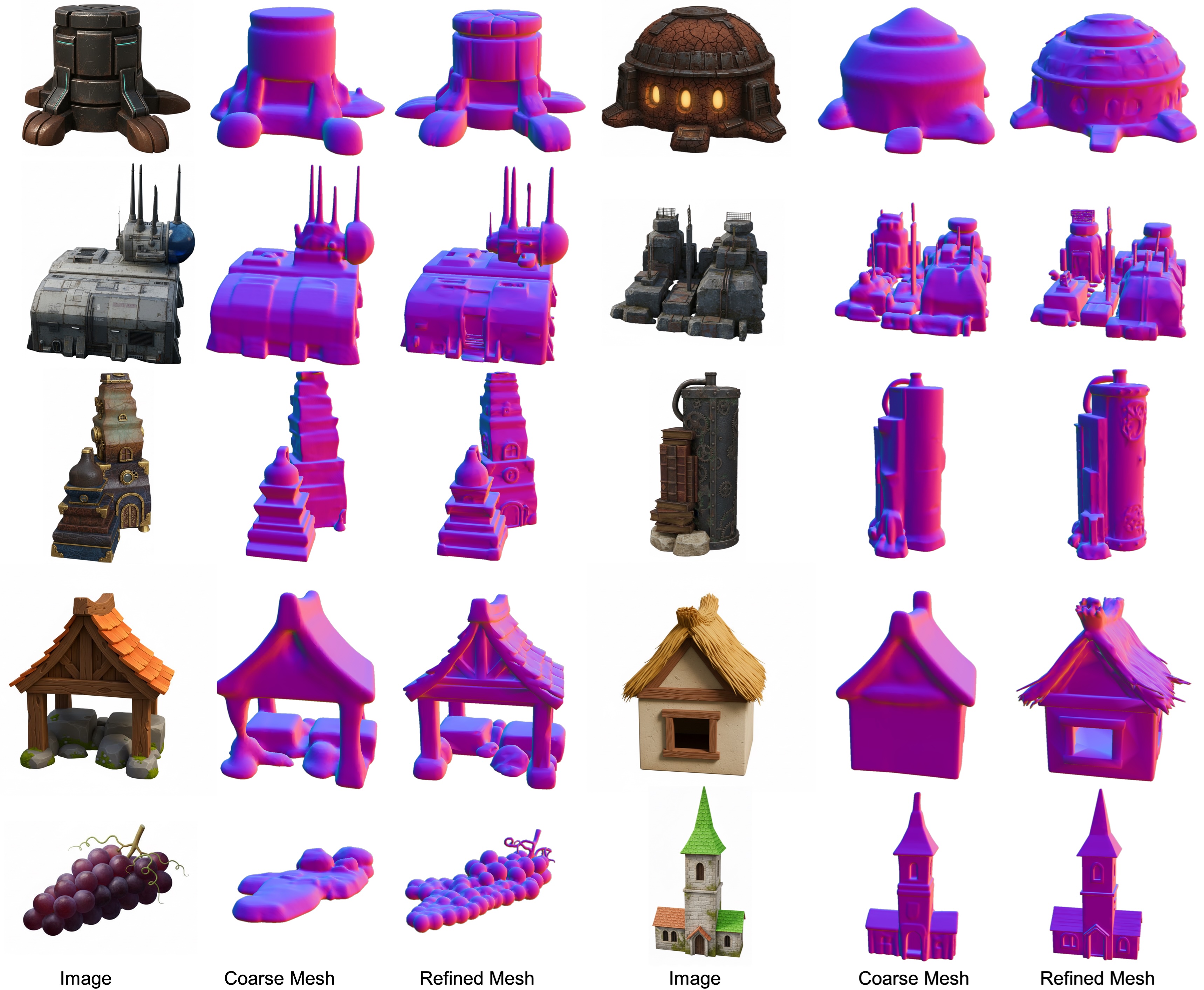}
    \caption{\textbf{Per-object mesh enhancement.} Each row shows two objects from the same scene, with the columns corresponding to image, coarse mesh, and refined mesh.}%
    \label{fig:part_refiner_results}
\end{figure}

\begin{figure}[t]
\centering
\includegraphics[width=0.99\linewidth]{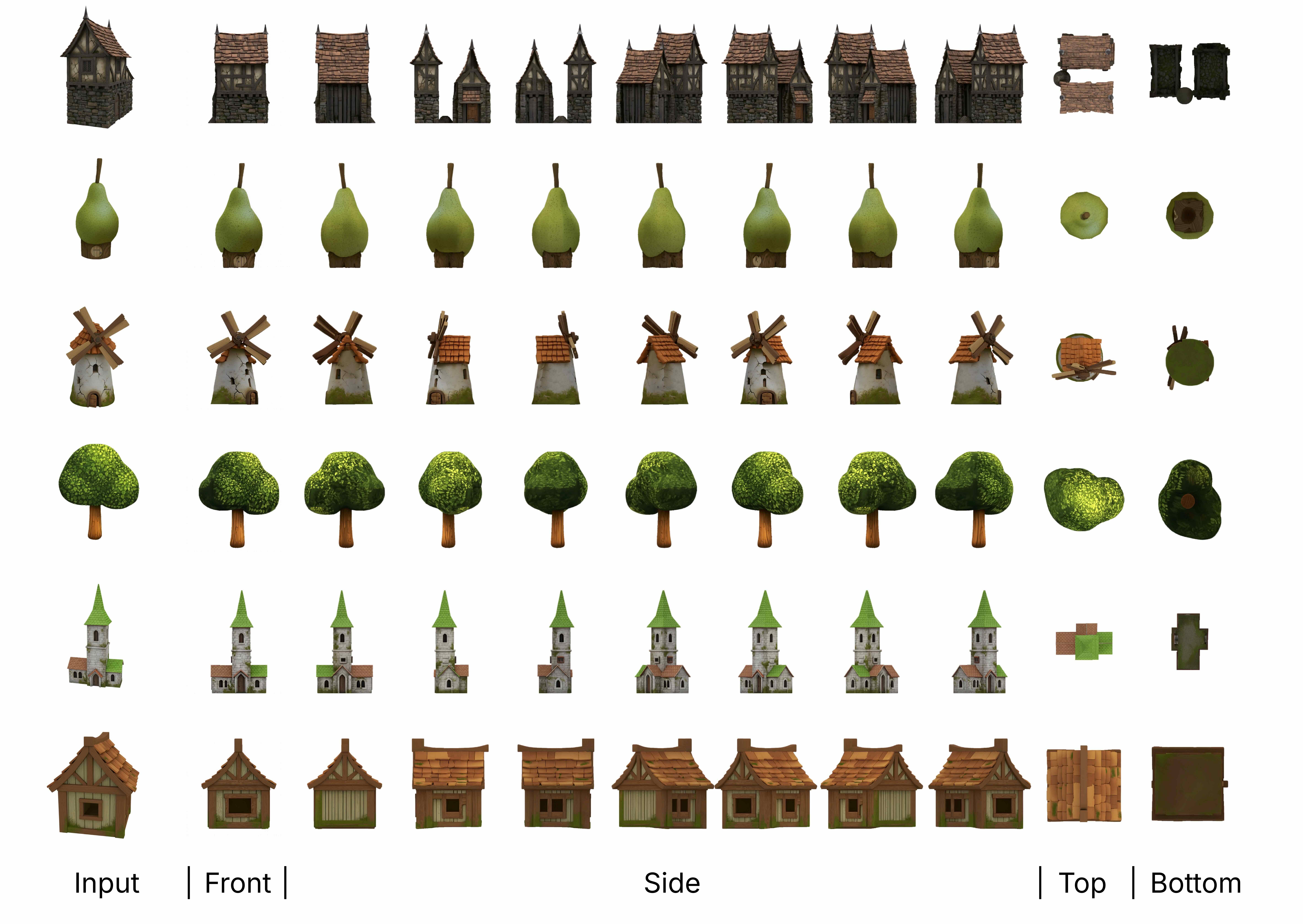}
\caption{\textbf{Multi-view texture generation.} Given a reference image as input, we sequentially generate: (1) frontal views, (2) side views conditioned on the frontal view, and (3) top and (4) bottom views conditioned on all previously generated views.}%
\label{fig:gridview}
\end{figure}

\subsection{Per-Object Image Enhancement}%
\label{sec:object-image-enhancement}


Given the scene reference image $\referenceImage$ from Stage~I, the initial textured mesh $\sceneMesh$ from Stage~II, and the decomposed per-object meshes $\sceneCoarse = \{(\xcoarse_i, \camera_i)\}_{i=1}^N$ from Stage~III, our goal for this step is to enhance the visual fidelity of each object's rendering $\objectImageCoarse$ by generating a high-resolution, detailed image $\objectImageFine$ that maintains stylistic consistency with the scene as a whole.

To make the image generator aware of the global scene context, we render a top-down view of the entire scene from $\sceneMesh$, where the target object is highlighted in red (\Cref{fig:object-image-enhancement}, second row).
This rendering, together with the global reference image $\referenceImage$, is provided to a large language-vision model (LLM-VLM), which identifies the corresponding object region in the global scene and analyzes its visual characteristics, including material attributes and color palette. 

Additionally, for each object $\xcoarse_i$, we render a view $\objectImageCoarse_i$ that captures its coarse geometry and low-resolution texture, which serves as the input reference for image enhancement (\Cref{fig:object-image-enhancement}, third row).
The top three rows of \Cref{fig:object-image-enhancement} show the input to the LLM-VLM model.
The LLM-VLM synthesizes a high-quality image $\objectImageFine_i$ that remains spatially aligned with the coarse input while enhancing fine-scale and decorative details.
Throughout this process, geometric alignment and global style consistency are maintained to ensure fidelity to the overall appearance of the scene (\Cref{fig:object-image-enhancement}, bottom row).

To evaluate the impact of our image enhancement strategy, we present ablation results without the top-down view in \Cref{fig:object-image-enhancement-ablation}. The LLM-VLM struggles to generate style-consistent or reference-faithful object images when conditioned only on the global reference image and without access to a top-down view. This highlights the importance of including a top-down view with the target object highlighted, as it provides essential context about the object’s location, semantics, and surroundings within the scene.

Since generative image enhancement can sometimes introduce geometric or stylistic drift, such as shape distortion or camera view, we apply a verification module that compares the enhanced and coarse renderings.
In particular, we compute the Intersection over Union (IoU) between the foreground object in the original coarse mesh render $\objectImageCoarse_i$ and $\objectImageFine_i$, and only accept results that have a high IoU within a certain threshold.
Feedback from this verification is used to iteratively refine the enhancement process with the LLM-VLM, ensuring alignment to the underlying object geometry.
See \Cref{fig:image_verification} for examples of the initial $\objectImageCoarse$, failure cases by the LLM-VLM model, and final results after verification and iterative refinement.

Although objects are refined independently, style coherence across the entire scene is mostly preserved because each object image enhancement is conditioned on the same global reference image $\referenceImage$ and the initial textured global mesh $\sceneMesh$.
The resulting set of enhanced images $\{\objectImageFine_i\}_{i=1}^N$ provides rich, high-resolution visual cues for subsequent geometry and texture refinement stages.

\subsection{Per-Object Mesh Enhancement}%
\label{sec:object-mesh-enhancement}

Given a coarse object mesh $\xcoarse_i$ and a high-resolution image $\objectImageFine_i$ from the \textit{Image Enhancement} step as input conditions, our \emph{Mesh Refinement Model} is trained to generate a high-resolution object mesh $\xfine_i$ that preserves the orientation of the coarse mesh while adding fine geometric details.

\paragraph{Architecture.}

The Mesh Refinement Model closely follows the \assetgen architecture (\Cref{sec:assetgen}), with an extended input dimension to accommodate coarse shape conditioning.
To leverage pre-trained 3D priors, we fine-tune this model from the base \assetgen.
Specifically, we first encode the coarse mesh using \assetgen's VAE to obtain its latent representation $\zcoarse_i$.
We enhance $\zcoarse_i$ by incorporating positional embeddings and applying a zero-initialized linear projection, which helps preserve the pre-trained prior at the start of training. The resulting codes are concatenated with the noise latent along the sequence dimension and subsequently used as input to the diffusion model, as illustrated in Figure~\ref{fig:partrefiner_arch}.
After denoising, $\zcoarse_i$ is discarded.
This design lets the model account for the rough geometry of $\xcoarse_i$, outputting a new object $\xfine_i$ that incorporates cues from the highly-detailed image $\objectImageFine$ while preserving the orientation and overall shape of the original coarse mesh. 

\paragraph{Training Data Curation.}

To train the \emph{Mesh Refinement Model}, we need triplets $\{\xcoarse_i, \xfine_i, \image_i\}$ comprising the input coarse mesh $\xcoarse_i$, the target high-resolution mesh $\xfine_i$, and its corresponding image $\objectImageFine_i$.

The low quality of $\xcoarse_i$ obtained from Stage~II (\Cref{sec:scene-reconstruction}) stems from the limited capacity of the latent 3D representation.
Because the latent size is fixed, the model must encode the entire scene within a constrained representation.
As scene complexity increases—i.e., as more objects are introduced—the representational capacity allocated to each object decreases.
Consequently, geometric fidelity deteriorates, resulting in lower-quality reconstructions.

We use this observation to construct an approximation of $\xcoarse$ from high-quality 3D objects $\xfine$.
We do so by creating synthetic ``scenes'' by arranging several (ground-truth) objects $\xfine_i$ in $2 \! \times \! 2$ and $3 \! \times \! 3$ grids (by varying the grid size, we control the complexity of the scene and thus the degree of geometry degradation).
We then render images of these scenes and feed them to \assetgen to reconstruct back the scenes, thus simulating the degradation observed in Stage~II\@.
From this reconstruction, we extract the degraded objects $\xcoarse_i$ using their known grid locations.
Finally, the images $\objectImageFine_i$ are obtained by rendering the ground-truth objects $\xfine_i$ from different camera viewpoints.

During training, we further augment the coarse objects $\xcoarse_i$ by simulating additional artifacts, including floaters, randomly masked-out regions, and broken surfaces, to improve robustness.
Additionally, the conditioning images for each object mesh are augmented with color jitter, randomized backgrounds, and random blur.

\paragraph{Mesh Enhancement Results.}

\Cref{fig:part_refiner_results} demonstrates the effectiveness of our Mesh Refinement Model.
Compared to the coarse inputs, the refined meshes are significantly sharper and more detailed, are more plausible, and contain fewer artifacts like floaters and surface discontinuities.
Overall, the resulting meshes are clean and ready for high-quality texture synthesis.

Since the Mesh Refinement Model generates objects in normalized scales, we rescale them to reconstruct the original scene layout.
Importantly, the Mesh Refinement Model preserves the orientation of each input coarse mesh, so only the axis-wise scaling factors and centroid positions of the coarse inputs $\xcoarse_i$ are required.
Using these parameters, each refined object is restored to its original position and orientation within the scene while recovering its proper scale.
In this way, the refined objects maintain the spatial relationships, relative scales, and orientations defined by the initial layout, so the scene remains consistent.

\subsection{Per-Object Texture Enhancement}%
\label{sec:object-texture-enhancement}

We finally generate high-resolution textures for each object $\xfine_i$ based on its enhanced image $\objectImageFine_i$ and the super-resolved geometry.
We use the texturing model part of \assetgen. Following an established paradigm for texture generation~\citep{bensadoun24metatexture}, we fine-tune a pretrained text-to-image latent diffusion model to produce 3D-consistent multi-view renderings of the object conditioned on normal and position maps for the target views, as well as the delighted version of an enhanced image $\objectImageFine_i$. We backproject the generated multi-view images to UV map to get the final texture image.

\paragraph{Delighting the Conditioning Image.}
\label{sec:texturing-delight}

Since the enhanced object image $\objectImageFine_{i}$ contains baked-in lighting and shading effects, it often exhibits complex illumination patterns, shadows, and specular highlights.
To mitigate this, we train a delighting model by fine-tuning a text-to-image latent diffusion model, where the latent representation of the shaded input image is provided as an in-context conditioning signal for generation.

\paragraph{Generating Multi-view Images.}
\label{sec:texturing-multiview}

We build upon Meta 3D TextureGen~\citep{bensadoun24metatexture} with the following design choices. 
First, we make the generator conditioned on the image. The latent of the condition image is supplied as an in-context input to guide the generation process. Second, we generate ten orthographic multi-view images, including eight side views evenly spaced at $45^\circ$ around the object at $0^\circ$ elevation, along with top and bottom views (see \Cref{fig:gridview}). Third, we employ sequential generation strategy, where we first generate the frontal view, then side views, and finally the top and bottom views. We empirically find that this improves the cross-view coherence and reduces geometric distortion.

\paragraph{Disentangled Multi-View Attention.}

We employ disentangled attention, where self-attention block is decomposed into 3 attention blocks -- in-plane self-attention, reference attention, and multi-view attention.
The first is \emph{in-plane self-attention}, where each view independently attends to its own spatial features, preserving local coherence and detail within individual renderings.
The second is \emph{reference attention}, where the generated views (i.e., views $1$ to $N-1$) attend to the reference view (i.e., view $0$) via cross-attention, ensuring all synthesized views remain consistent with the input enhanced image.
The third is \emph{multi-view attention}, where the generated views attend to each other, promoting global 3D consistency across different viewpoints while maintaining strong adherence to the reference image.
This factorization enables more structured feature interactions across views.

\paragraph{Texture Post-processing.}
\label{sec:texturing-postproc}

Once the ten views are generated, we initialize the UV texture by back-projecting the multi-view images onto the object's surface. This step yields sharp and well-aligned textures for all regions visible in at least one generated view. 
Finally, we apply an inpainting algorithm in UV space to fill small gaps and unobserved areas, producing complete, high-quality textures ready for scene assembly.

\vspace{1cm}

\section{Results}%
\label{sec:experiments}

We first present qualitative examples generated by our \method in \Cref{sec:generated-scenes}.
Then, we compare these qualitatively to relevant prior work in \Cref{sec:comparison-prior-work}.

\foreach \name/\captiontext in {
    scene_dilin1/Medieval town square,
    scene_1023_015/Snowy village
}{
    \clearpage
    \begin{figure}[ht]
        \centering
        \includegraphics[width=0.99\linewidth]{assets/scene_render_picked/\name.jpg}
        \caption{\captiontext}%
        \label{fig:\name}
    \end{figure}
    \clearpage
}

\subsection{Examples of Generated Scenes}%
\label{sec:generated-scenes}

In Figure~\ref{fig:scene_dilin1} and Figure~\ref{fig:scene_1023_015} and Appendix \Cref{sec:appendix}, we provide a gallery of several scenes generated by our \method system, end-to-end from single user prompts.
Overall, our method is capable of generating interesting, diverse, and good-looking scenes that can be navigated freely by characters, and are thus suitable for use in game engines.
Each scene contains multiple semantically consistent objects, coherent textures, and a valid navmesh that enables real-time exploration.
Despite the multi-stage nature of our approach, the entire pipeline—from a single text prompt to a fully textured, navigable 3D scene—completes in approximately five minutes because many submodules (e.g., object enhancement and texture generation) can run in parallel (assuming that sufficient GPUs are available for this purpose).
This enables rapid prototyping of interactive worlds with minimal user intervention.

\subsection{Qualitative Comparison with Prior Work}%
\label{sec:comparison-prior-work}

There are few prior systems capable of generating game-like or immersive environments, and they all differ substantially in assumptions and nature of the generations.
Hence, these are difficult to compare directly.

\begin{figure}[t]
    \centering
    \includegraphics[width=0.98\linewidth]{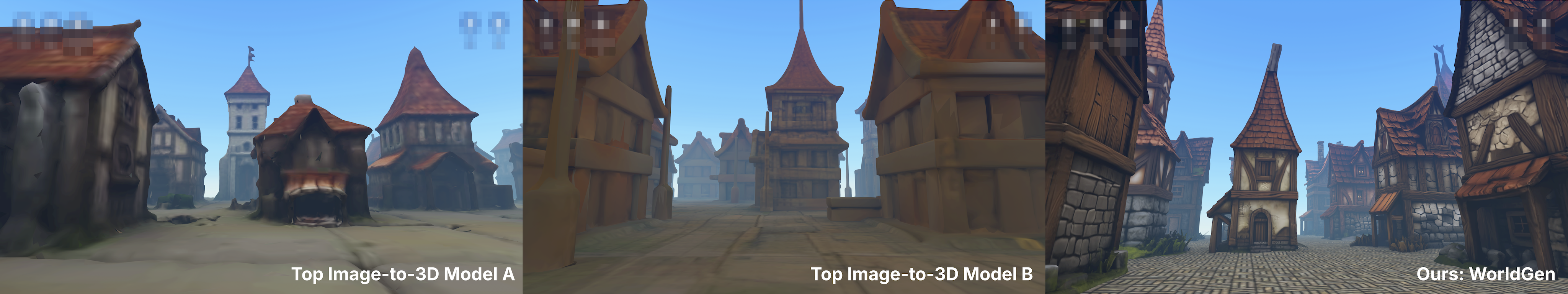}
    \caption{\textbf{Comparison with state-of-the-art image-to-3D methods}. \method generates scenes that are significantly more detailed than single-shot reconstructions.}
    \label{fig:comparison-prior-work}
\end{figure}
\paragraph{Comparison with Image-to-3D}
The most relevant prior work to ours lies in image-to-3D generation. We compare with state of the art image-to-3D solutions in this domain.
While these methods achieve impressive results on single objects and small-scale scenes, they are inherently non-compositional and not designed for navigable or large-scale environments.
Their generated geometry and texture resolution remain insufficient for direct use in game engines.
As shown in~\Cref{fig:comparison-prior-work}, these single-shot models produce outputs that lack the geometry and texture details required to support immersive, explorable 3D experiences.

\begin{figure}[t]
\includegraphics[width=0.95\linewidth]{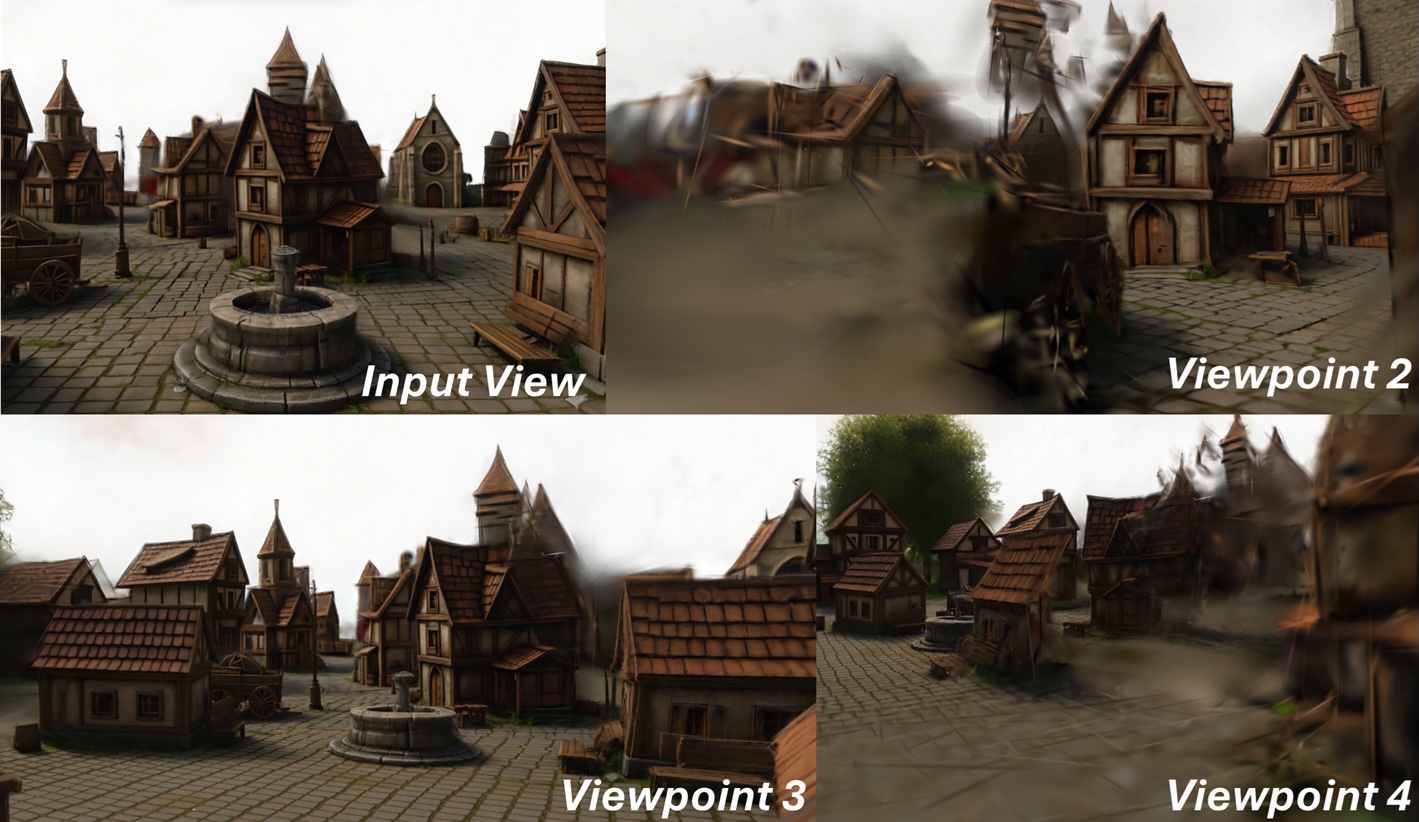}
\caption{\textbf{World Labs examples.} Example results of the marble scene generation under different configurations.}%
\label{fig:worldlabs_comparison}
\end{figure}

\paragraph{Comparison with Marble}
Another significant class of scene generation methods are the view-based ones (\Cref{sec:view-based-3d-scene-generation}).
While the details are not confirmed, a reasonable guess makes us to believe that recent systems such as \marble~\footnote{\url{https://marble.worldlabs.ai/worlds}} from World Labs are likely the best that this class of generators has to offer.

A direct comparison with \marble is yet again not straightforward, as the \method and \marble systems differ fundamentally in input and scope.
\marble grows a scene outward from a single specified viewpoint rather than conditioning on a global reference image or full layout.
They represent environments using millions of
Gaussian splats to achieve high visual fidelity.
Their most significant advantage compared to \method is that Gaussian splats easily bake a radiance field, giving scenes a more photorealistic appearance.
However, there are several important limitations that make \marble less suitable for generating large-scale, interactive 3D worlds compatible with standard game engines.

To enable a qualitative comparison, we rendered a central view of our ``medieval village'' scene and provided it as input to \marble.
A first limitation is \emph{extente}.
While \marble produces high-quality geometry and textures near the conditioned view, its fidelity still degrades as the camera moves just a few meters away (e.g., 3--5 m), as shown in \Cref{fig:worldlabs_comparison}.
In contrast, our generated scenes span approximately $50 \times 50$ m, are fully textured, and maintain geometric and stylistic consistency throughout the environment,
allowing users to freely navigate and interact within the world.

As another key advantage over \marble is that our approach produces scenes that are directly compatible with standard game engines such as Unreal and Unity.
Gaussians splats, while visually impressive, are not natively supported by these engines and require specialized rendering pipelines.
Likewise, they are not supported by standard toolsets that 3D artists use to create games.
In contrast, \method outputs worlds which are compositions of textured meshes, enabling seamless integration with any game engine and toolset.
Compositionality also makes it easier to edit, move or replace objects in the scene.
Furthermore, while Gaussian splats can be rendered on-device in real time, they remain orders of magnitude slower than reasonably well optimised meshes, which make them difficult to support on mobile and low-end handrware.
Our method's compact textured-mesh representation is significantly more efficient and portable,
supporting large, persistent, and navigable 3D worlds that can be readily deployed across a wide range of hardware.

\section{Related Work}%
\label{sec:related-work}

We review work of relevance to 3D scene generation, distinguishing
scene reconstruction (\Cref{sec:related-scene-reconstruction}) and
monolithic (\Cref{sec:related-monolithic-scene-generation}),
compositional (\Cref{sec:related-compositional-scene-generation}), and
procedural (\Cref{sec:related-procedural-scene-generation}) scene generation.

We focus almost exclusively on static 3D scenes, touching only occasionally on dynamic scene reconstruction and generation when relevant here.
Recent surveys on 3D scene generation~\citep{wen253d-scene,tang25recent} provide further pointers and comparisons.

\subsection{Image-based Scene Reconstruction}%
\label{sec:related-scene-reconstruction}

\method is based on reconstructing the geometry and appearance of a 3D scene from an image of it, so methods for single and few-view scene reconstruction are relevant.

Some of the most successful approaches for reconstructing complex 3D scenes from images are based on Learnable Radiance Fields (LRF).
Pioneered by Neural RFs, or NeRFs~\citep{mildenhall20nerf:}, they have become even more popular with 3D Gaussian Splatting (3DGS)~\citep{kerbl233d-gaussian}.
3DGS has introduced a more explicit, straightforward, and efficient representation of RF than NeRF, based on a mixture of colored 3D Gaussians.
While RFs can represent complex scenes with high fidelity, they typically require hundreds of views for reconstruction.
These are seldom available, particularly when the scene is generated from a text prompt.
They also use relatively slow energy minimisation, which introduces latency in applications.

More relevant here are methods that can output RF representations of 3D scenes efficiently, from a single or few images.
This requires learning reconstructors, generally based on deep neural network architectures.
An early example is Layer-Structured~\citep{tulsiani18layer-structured}, which predict a multi-layer reconstruction of a scene from a single input image.
Others are
Behind the Couch~\citep{kulkarni21whats} and
Behind the Scenes~\citep{wimbauer23behind}, which estimates, respectively, a per-pixel ray distance function and density field from a single image.
SinNeRF~\citep{xu22sinnerf:} attempt to reconstruct NeRF representation from a single image using ad-hoc regularizers.

A popular family of such reconstructors uses \emph{pixel-aligned 3DGS}, where each pixel in each view is mapped to a corresponding 3D Gaussian by a network.
Two pioneers include
the Splatter Image~\citep{szymanowicz24splatter}, which reconstructs objects from a single image, and
PixelSplat~\citep{charatan24pixelsplat:}, which interpolates between two views of a scene.
MVSplat~\citep{chen24mvsplat:} considers more than two input views and
MVSplat360~\citep{chen24mvsplat360} a system of views that cover a scene in a 360\textdegree{} manner.
Flash3D~\citep{szymanowicz25flash3d} extends Splatter Image to scenes, and allows monocular 3DGS reconstruction.
LVT~\citep{imtiaz25lvt:} introduces a more efficient geometry-aware attention module.
GaussianRoom~\citep{xiang25gaussianroom:} introduce a sign-distance function (SDF) regularizer in the 3DGS framework to improve its statistical efficiency.
Some approaches specialise to driving data.
For example,
Neural Urban Scene Reconstruction~\citep{shen24neural} and 
AutoSplat~\citep{khan24autosplat:} make use of the LIDAR sensor that is popular in automotive applications to help with scene reconstruction and
Omni-Scene~\citep{wei25omni-scene:} combine several camera-mounted vehicle for 360\textdegree{} scene reconstruction.

With the exception of the Splatter Image and Flash3D, which are monocular, a disadvantage of these methods is that they assume known camera poses.
Authors originally assumed that camera poses are either available externally, or can be estimated via optimization-based approaches like bundle adjustment, which significantly undermines their applicability.
Hence, some recent works have focused on reconstructing 3D geometry and camera parameters in a feed forward manner.
Learning to Recover 3D Scene Shape~\citep{yin21learning} recognised early the importance of point cloud prediction to recover the camera intrinsics.
Point maps were then used by
DUSt3R~\citep{wang24dust3r:} and
MASt3R-SfM~\citep{duisterhof24mast3r-sfm:}
and, more recently, by
VGGT~\citep{wang25vggt},
MV-DUSt3R+~\citep{tang2025mv},
$\pi^3$~\citep{wang25p3},
Fast3R~\cite{yang25fast3r:},
Pow3R~\cite{jang25pow3r:}, and
Map Anything~\citep{keetha25mapanything:}
with excellent reconstruction results.
These tools can estimate camera poses automatically and, in VGGT and follow ups, accurately without post-optimization from several views simultaneously.

While these methods focus on reconstructing geometry, several authors have already built on them to obtain reconstructions of both geometry and appearance.
An example is
Splatt3R~\citep{smart24splatt3r:}, which builds on MASt3R to perform feed-forward 3DGS reconstruction without the need to specify cameras.
NoPoSplat~\citep{ye24no-pose} learns a vision transformer~\citep{dosovitskiy21an-image} from scratch to reconstruct 3DGS in a shared canonical space from multiple views, which is conceptually analogous to DUSt3R and follow ups.
AnySplat~\citep{jiang25anysplat:} is instead based on VGGT\@.

One limitation of RFs like 3DGS is that they \emph{lack structure}: all objects comprising a scene are represented indistinctly as a single whole.
A few authors have considered the problem of also extracting components (objects) from them.
For instance
ObjCompNeRF~\citep{yang21learning},
Nerflets~\citep{zhang23nerflets:},
CompoNeRF~\citep{lin23componerf:} and
Generalizable 3D Scene Reconstruction~\citep{dogaru24generalizable}
uses an implicit RF-based representation and
InstaScene~\citep{yang25instascene:} and
DecoupledGaussian~\citep{wang25decoupledgaussian:}
use 3DGS to do so.

For our purposes, all these methods have another major limitation: they reconstruct only the visible part of a scene-hence, a complete scene reconstruction requires a complete set of views, covering all aspects of it.
In contrast, in \method we develop a method to extract a complete and well structured scene representation from a single input image.

\subsection{Monolithic 3D Scene Generation}%
\label{sec:related-monolithic-scene-generation}

There is also significant work in generating 3D scenes.
Here, we consider works that generates scenes as a whole, entirely or mostly disregarding their compositional structure.
We consider
view-based monolithic scene generation in \Cref{sec:view-based-3d-scene-generation} and
latent-space monolithic scene generation in \Cref{sec:latent-monolithic-scene-generation}.

\paragraph{View-based Monolithic 3D Scene Generation}%
\label{sec:view-based-3d-scene-generation}

Several authors reduce the problem of generating scene to those of generating novel \emph{views} of the scenes, often while simultaneously building a 3D representation of it.
These approaches are often incremental, adding one view at a time, and use priors such as depth predictors and image/depth inpainters to build out the scene.

An early example is SynSin~\citep{wiles20synsin:}, which generates new view of a scene starting from a single image: using  depth prediction, they infer a latent representation of the 3D scene which can be rendered from novel viewpoints.

Follow up works are often based on using depth to warp the pixels directly as a seed to generate a novel view.
PixelSynth~\citep{rockwell21pixelsynth:} is one of the first to do so, and also uses an autoregressive 2D inpainter to fill the holes left by the warping.
CompNVS~\citep{li22compnvs:} uses inpainters for both appearance and geometry.
Text2NeRF~\citep{zhang23text2nerf:} uses a more powerful image diffusion model for inpainting, and builds a Neural Radiance Field (NeRF)~\citep{mildenhall20neural} representation of the 3D scene.
Text2Room~\citep{hollein23text2room:} generates instead a texture 3D mesh, and has a mechanism to select informative novel views to complete the scene.
Text2Immersion~\citep{ouyang23text2immersion:} uses a 3D Gaussian Splatting (3DGS)~\citep{kerbl233d-gaussian} representation and
3D-SceneDreamer~\citep{zhang243d-scenedreamer:} one based on triplanes~\citep{chan22efficient}.
Infinite Nature~\citep{liu21infinite},
DiffDreamer~\citep{cai23diffdreamer:},
RGBD2~\citep{lei23rgbd2:},
Text2Immersion~\citep{ouyang23text2immersion:},
SceneScape~\citep{fridman23scenescape:},
WonderJourney~\citep{yu23wonderjourney:}, and
RealmDreamer~\citep{shriram25realmdreamer:}
also alternate between estimating depth, moving the camera to generate a new view, and inpainting the latter using image generators, but with the difference of generating perpetual walks out of the initial image, inspired by the earlier image stitching work of~\citep{kaneva10infinite}.
Ctrl-Room~\citep{fang25ctrl-room:} extends Text2Room separating layout and appearance generation.

LucidDreamer~\citep{chung23luciddreamer:} incrementally constructs a point cloud representation of the 3D scene along with the new views, and uses it to improve alignment and consistency, ultimately reconstructing a 3DGS representation of the scene.
3D-Aware Indoor Scene Synthesis~\citep{shi223d-aware} and
LDM3D~\citep{stan23ldm3d:}
learn a diffusion model that generates novel views of the scene along with a depth map, i.e., a RGBD image, which helps stitching them.
Similarly, the works of~\citep{xiang233d-aware,lei23rgbd2:} trains an RGBD diffusion model to inpaint RGBD images, and Invisible Stitch~\citep{engstler25invisible} does so for depth images separately.
Other recent examples of view-based scene generators include
BloomScene~\citep{hou25bloomscene:}.

WonderWorld~\citep{yu25wonderworld:} and
WonderTurbo~\citep{ni25wonderturbo:}
focuses on speed, generating new views in less than a second, close or at interactive rates.
Learning Object Context~\citep{qiao22learning}
consider the problems of generating 2D segmentation maps instead of RGB images, thus capturing the compositional structure of the scene, but still primarily in a view-based manner.

Specialised methods for driving scenes include MagicDrive3D~\citep{gao24magicdrive3d:}, DriveDreamer4D~\citep{zhao25drivedreamer4d:}, DreamDrive~\citep{mao25dreamdrive:}, UniScene~\citep{li25uniscene:}, ReconDreamer~\citep{ni25recondreamer:,zhao25recondreamer:}, HERMES~\citep{zhou25hermes:}, and DiST-4D~\citep{guo25dist-4d:}. Many of these address dynamic scene generation (i.e., moving traffic). Other view-based dynamic scene generators include Free4D~\citep{liu25free4d:}, which extracts a 3DGS representation from a generated 2D video using MonST3R~\citep{zhang24monst3r:} as an initial scaffold. We do not consider dynamics here.

Some view-based scene generators operate in parallel on several images of the scene rather than sequentially in order to improve consistency.
A first example is MVDiffusion~\citep{tang23mvdiffusion:}, which however assume known geometry (depth), or that geometry is not relevant (no parallax).
Other examples are
SceneDreamer360~\citep{li24scenedreamer360:} and
DreamScene360~\citep{zhou24dreamscene360:}
which starts by generating a 360\textdegree{} panoramic image of the scene as a reference.
LayerPano3D~\citep{yang25layerpano3d:}, which generates a panoramic 360\textdegree{} image of the scene, which is then extended to layer-wise 3D representation of the content primarily based on depth prediction and inpainting behind occlusions.
Other approaches, such as IM-3D~\citep{melas-kyriazi24im-3d}, V3D~\citep{chen24v3d:}, CAT3D~\citep{gao24cat3d:}, and ReconX~\citep{liu24reconx:}, first generate multiple highly consistent views and then reconstruct a 3D representation from them in a subsequent phase.

Some works like
Director3D~\citep{li24director3d:},
StarGen~\citep{zhai25stargen:},
Generative Gaussian Splatting~\citep{schwarz25generative}
build on video generators, which help with the generation of longer sequences of views, and hence larger scenes.
DimensionX~\citep{sun24dimensionx:} and
4Real-Video-V2~\citep{wang254real-video-v2:}
extend a video generator to output a time-viewpoint grid of images, which are then reconstructed into a 3DGS representation of the scene based on VGGT\@.

Some methods straddles the boundary between view-based and latent-space 3D scene generation (\Cref{sec:latent-monolithic-scene-generation})
GAUDI~\citep{bautista22gaudi:} uses a auto-decoder to learn a latent space of RF scenes, and then learns to sample latents from it.
NeuralField-LDM~\citep{kim23neuralfield-ldm:} learn a `translational' VAE that maps several RGBD views of a scene to a latent space, which can then be rendered into novel views.
This latent space is then used to generate complete scene using denoising diffusion.
Prometheus~\citep{yang25prometheus:} learns an analogous VAE, but decoding from the latent representation pixel-aligned 3DGS maps.
It then learns a denoising diffusion model to sample this latent space given a text prompt.
Bolt3D~\citep{szymanowicz25bolt3d:} also learns a latent space to represent a scene based on point maps and 3DGS from several simultaneous views.
They learn the latent space by obtaining first 3DGS `ground-truth' reconstructions from dense multi-view images of a large number of scenes, which limits the diversity of the training data.
Wonderland~\citep{liang24wonderland:} learns to extract a 3DGS representation of a scene from the latents generated by a camera-controlled video diffusion model.
Lyra~\citep{bahmani25lyra:} also learns an 3DGS decoder on top of the camera-controlled video generator Gen3C~\citep{ren25gen3c:}; their main contribution is that they self-distill the 3DGS decoder from the images generated by the video generator, without requiring multi-view images of real scenes, a significant limitation of prior works.

View-based methods are often limited to the size of the walkable scene space they can generate.
Often they generate a `bubble' of a few meters across: while the observer may look at infinity (e.g., at the sky), it cannot move by more than a few steps before defects and incompleteness of the scene geometry become apparent.
Some like Text2Room do generate full enclosed spaces, but the incremental approach tends to drift as more pieces are stitched together, which results in inconsistencies and accumulates distortions.
Many of these approaches, furthermore, prioritise the generation of views of the scene over the recovery of an underlying 3D model.

\paragraph{Latent Monolithic 3D Scene Generation}%
\label{sec:latent-monolithic-scene-generation}

View-based scene generation can easily tap powerful 2D image and video generators, and often results in high-quality, photorealistic views of the generate scene.
However, these methods struggle to recover a robust 3D scene geometry, particularly when moving away from small bubbles.
An alternative approach is to directly generate the 3D geometry of the scene, usually encoded in a suitable \emph{latent space}.
This approach has been very successful for single object generation
(%
VecSet~\citep{zhang233dshape2vecset:},%
Clay/Rodin~\citep{zhang24clay:,deemos24rodin},%
Tripo~\citep{tripo3d24text-to-3d},%
Trellis~\citep{xiang24structured},%
Sparc3D~\citep{li25sparc3d:},%
and several others).
SynCity~\citep{engstler25syncity} repurposes 3D object latent spaces to generating whole scenes in a tile-by-tile manner.
LT3SD~\citep{meng24lt3sd:},
SceneFactor~\citep{bokhovkin24scenefactor:},
BlockFusion~\citep{wu24blockfusion:}, and
NuiScene~\cite{lee25nuiscene:}
learn 3D latent spaces specifically for scene generation.
While these latent space models are robust and effective, they lacks diversity, partially due to the challenge of collecting 3D training data for varied scenes.
Controllable 3D Outdoor Scene Generation~\citep{liu25controllable} generates first a graph representing the scene elements and then uses denoising diffusion to generate a voxel representation of the scene conditioned on the graph.

Generative GS for Cities~\citep{xie25generative} and
CityGen~\citep{deng25citygen:}
proposed latent representation for generating urban scenes.

\paragraph{Other Models}%
\label{sec:other-models}

Several scenes and object generators are based on Generative Adversarial Networks (GANs)~\citep{goodfellow14generative}.
The idea is to learn to output 3D representations from which realistic images, as judged by a discriminator network, can be rendered.
An early example is HoloGAN~\citep{nguyen-phuoc19hologan:}, which produces an implicitly 3D latent representation and renderer.
Models like 
Semantic UV Mapping~\citep{vermandere24semantic},
SceneTex~\citep{chen24scenetex:},
LumiNet~\citep{xing25luminet:} and
RoomPainter~\citep{huang25roompainter:}
focus on generating only the appearance of scenes instead of their geometry.

\paragraph{Evaluating 3D scene generation}

WorldScore~\citep{duan25worldscore:} provides a fairly general framework that can be used to asses such models by measuring the quality of the views generated from the scenes.

\subsection{Compositional 3D Scene Generation}%
\label{sec:related-compositional-scene-generation}

Compositional 3D scene generators focus on building scenes out of objects, either
retrieved from a database (\Cref{sec:related-existing-objects}) or
generated ad-hoc (\Cref{sec:related-create-and-arrange}).

\paragraph{Selecting and Arranging Existing Objects}%
\label{sec:related-existing-objects}

Some methods formulate scene generation as arranging existing 3D assets, retrieving them from a database, and generating plausible layouts out of them.
Interactive Learning of Spatial Knowledge~\citep{chang14interactive} is an example of an early method that employs lexical analysis to interpret user instructions, and derives from them constraints to place objects in a scene.
Deep Convolutional Indoor Scene Synthesis~\citep{wang18deep-conv} utilizes a convolutional network to decide where to place objects in a scene.
Fast Indoor Synthesis~\citep{zhang21fast} represents scenes as graphs, learning how furniture distributes within rooms.
CommonScenes~\citep{zhai23commonscenes:} and
RoomDesigner~\citep{zhao23roomdesigner:}
encodes individual objects as vectors, which allows the layout generator to account for their fine-grained 3D shape.
Hierarchically-Structured~\citep{sun25hierarchically-structured}
DiffuScene~\citep{tang24diffuscene:} utilizes a diffusion approach to generate scene layouts.
InstructScene~\citep{lin24instructscene:} proposes to use a Large 
Language Model (LLM) to generate a scene graph based on user instructions.
Open-Universe~\citep{aguina-kang24open-universe} develops an LLM that can combine objects extracted automatically from a large, unannotated dataset.
PhyScene~\citep{yang24physcene:} learns to retrieve and arrange objects in a physically-plausible manner.
SceneTeller~\citep{ocal24sceneteller:} use an LLM to first generate a 3D layout of the scene as a collection of named 3D bounding boxes, which are then filled in with 3D assets retrieved from a database.
Auto-regressive
CasaGPT~\citep{feng25casagpt:} generates scenes auto-regressively as composition of cuboids; the resulting scene approximation can then be used to recall and substitute-in 3D assets from a database.

\paragraph{Image-to-scene.}

Particularly relevant to our work are 3D scene generators prompted from an input image, which we call `image-to-scene'.
Some of these are based on reproducing the input scene based on retrieved 3D assets.
Patch2CAD~\citep{kuo21patch2cad:},
ROCA~\citep{gumeli22roca:},
DiffCAD~\citep{gao24diffcad:}, and
Digital Cousins~\citep{dai24automated}
retrieve CAD models from a database to reconstruct objects in a scene based on partial 3D scans of it.

Sketch2Scene~\citep{xu24sketch2scene:} generates first a view of the scene, and use the latter to reconstruct it in 3D.
Differently from us, they do not perform a full 3D reconstruction, but first segment the background and objects in 2D, then reconstruct the background in 3D, and finally substitute the objects by retrieving similar ones from a database and add details using a PG\@.
Diorama~\citep{wu24diorama:} can also generate a scene by parsing and reconstructing a single image of it based on existing 3D assets.

Scene-Conditional 3D Object Stylization~\citep{zhou23scene-conditional} consider the problem of adjusting objects to fit in a scene, e.g., changing their style to match the scene's lighting and materials.

\paragraph{Generating and Arranging Objects}%
\label{sec:related-create-and-arrange}

More relevant to this work are methods that generate both the objects and their arrangement in a scene.

SceneDreamer~\citep{chen23scenedreamer:} and
BerfScene~\citep{zhang24berfscene:}
generate a scene starting from a bird-eye view (BEV) of its layout, and then adding details to the elevation utilizing adversarial learning.
DisCoScene~\citep{xu23discoscene:} generates local RFs for individual objects as well as one for the background using adversarial losses.
Disentangled 3D Scene Generation~\citep{epstein24disentangled},
Compositional Scene Generation~\citep{po24compositional}, and
Set-the-Scene~\citep{cohen-bar23set-the-scene:}
express layouts using 3D bounding boxes to control scene generation via scored distillation sampling~\citep{poole23dreamfusion:}.
GenUSD~\citep{lin24genusd:} and
GALA3D~\citep{zhou24gala3d:} follow an analogous approach, but use a text-to-3D generator to initialize the 3D objects, which is closer to our approach.
SceneWiz3D~\citep{zhang24scenewiz3d:} uses an LLM to propose objects to fill a scene, generates them using text-to-3D, and finally generate a RF of the scene background and the arrangement of the objects by optimizing the score assigned to renders of the scene by an image generator based on diffusion.
Direct Numerical Layout Generation~\citep{ran25direct} uses LLMs to reason about plausible spatial layouts before generating the objects.
Methods like
Scenethesis~\citep{ling25scenethesis:}, and
PhiP-G~\citep{li25phip-g:}
try to generate object configurations that satisfy physical constraints, such as stability and support relationships.
MMGDreamer~\citep{yang25mmgdreamer:} follows instructions expressed using a combination of language and text, generating a scene graph, layout and 3D object shapes.
GraphDreamer~\citep{gao24graphdreamer:} and
EchoScene~\citep{zhai24echoscene:} also guides diffusion-based object generator via a graph.

\paragraph{Image-to-scene.}

Particularly relevant to our work are 3D scene generators prompted from an input image, which we call `image-to-scene'.
MIDI~\citep{huang25midi:} and
HiScene~\citep{dong25hiscene:} start by generating an image of an indoor space, followed by its 3D reconstruction and decomposition into objects.
Physically-based compositional approaches such as 
CAST~\citep{yao25cast:},
SceneGen~\citep{meng25scenegen:} introduces a feed-forward model that, given ain image of the scene, simultaneously reconstructs the objects layout, shapes and textures (the latter building on the TRELLIS latent space~\citep{xiang24structured}).
LAYOUTDREAMER~\citep{zhou25layoutdreamer:}
3D-Scene-Former~\citep{chatterjee243d-scene-former:}
attempt to construct scenes by composing objects in a way which makes `physical sense', which generally means that objects support each other on top of the ground plane.
Coherent 3D Scene Diffusion~\citep{dahnert24coherent} reconstructs multiple objects in a single view as joint conditional 3D generation.

\subsection{Procedural 3D Scene Generation}%
\label{sec:related-procedural-scene-generation}

Procedural Content Generation use sad-hoc procedures and rule sets to generate 3D scenes.
While this technology was popularised primarily in computer graphics and gaming~\citep{hendrikx13procedural}, it has found several applications in machine learning, including the generation of popular 3D datasets like
Infinigen~\citep{liu21infinite} and
Infinigen Indoors~\citep{raistrick24infinigen}.

Some authors have started to combine LLMs and agents with procedural generation so as to incorporate more high-level reasoning and planning in the generation process, as well as to make the generator controllable by non-experts, using natural language.
Examples include
SceneX~\citep{zhou24scenex:} and
Text-Guided City Generation~\citep{feng25text-guided}.
SceneMotifCoder~\citep{tam24scenemotifcoder:} use LLMs not only to control procedural generation, but also to automatically write the procedures themselves.
SceneCraft~\citep{hu24scenecraft:} uses powerful LLMs to first map a user prompt into a graph representing a scene, and then the latter into Python code that uses Blender to generate the 3D scene.
In our work, we use procedural generation to generate and initial layout of the scene and then add details automatically using an image generator.

\section{Conclusions and Limitations}%
\label{sec:conclusions}

We presented \method, a system for generating traversable, interactive 3D worlds directly from text prompts.
Our approach unifies text-conditioned procedural layout generation, navmesh-guided scene synthesis, object-level decomposition, and fine-grained geometry and texture enhancement into an end-to-end pipeline that transforms high-level user prompt into game-engine-ready environments that are coherent, detailed, and  explorable.

While our results demonstrate the potential of text-driven world generation, several limitations remain.
Currently, \method relies on generating a single reference view of the scene, which restricts the scale of scenes that can be produced.
Large open worlds spanning kilometers are not supported natively and would require generating and stitching multiple local regions, which risks introducing non-smooth transitions or visual artifacts at region boundaries~\citep{engstler25syncity}.
The single-view conditioning also limits the ability to model multi-layered environments, such as multi-floor dungeons or seamless interior-exterior transitions.
Finally, since each object is represented independently without geometry or texture reuse, rendering efficiency may become a concern in very large scenes.
Future work should explore strategies such as texture tiling, reuse, and shared materials to improve scalability and runtime performance.

Overall, \method illustrates how language-guided procedural reasoning and 3D diffusion models can pave the way toward accessible and scalable interactive world generation for next-generation games and social experiences.

\section{Acknowledgement}%

We thank Ocean Quigley, 
Zack Dawson, Alexander Dawson, Vladimir Mironov, Kam Zambel, Vu Ha, Yoav Goldstein, Dhwaj Agrawal, Scott Nagy, Stephen Madsen, 
John Niehuss, Chin Fong, Christopher Ocampo, Milton Cadogan, Sandy Kao, Ryan Cameron and Barrett Meeker for their advice and support throughout this project. 

\clearpage
\newpage
\bibliographystyle{assets/plainnat}
\bibliography{paper,vedaldi_general_pruned,vedaldi_specific}

@string{ai = {Artificial Intelligence}}

@string{arxiv = {arXiv}}

@string{corr = {arXiv.cs}}

@string{cvpr = {Proc. {CVPR}}}

@string{eccv = {Proc. {ECCV}}}

@string{icassp = {Proc. ICASSP}}

@string{iccv = {Proc. {ICCV}}}

@string{iccvw = {Proc. {ICCV} Workshops}}

@string{iclr = {Proc. {ICLR}}}

@string{icml = {Proc. {ICML}}}

@string{nips = {Proc. {NeurIPS}}}

@string{siggraph = {Proc. {SIGGRAPH}}}

@string{texture = {Proc. Intl. Workshop on {Texture Analysis and Synthesis}}}

@string{threedv = {Proc. {3DV}}}

@misc{recast,
  title        = {Recast Navigation},
  author       = {Mikko Mononen and contributors},
  year         = {2016--2026},
  howpublished = {\url{https://github.com/recastnavigation/recastnavigation}},
  note         = {State-of-the-art navmesh generation and navigation for games.}
}

@article{perlin1985image,
  title={An image synthesizer},
  author={Perlin, Ken},
  journal={ACM Siggraph Computer Graphics},
  volume={19},
  number={3},
  pages={287--296},
  year={1985},
  publisher={ACM New York, NY, USA}
}

@article{pearson1905problem,
  title={The problem of the random walk},
  author={Pearson, Karl},
  journal={Nature},
  volume={72},
  number={1867},
  pages={342--342},
  year={1905},
  publisher={Nature Publishing Group UK London}
}

@inproceedings{fuchs1980visible,
  title={On visible surface generation by a priori tree structures},
  author={Fuchs, Henry and Kedem, Zvi M and Naylor, Bruce F},
  booktitle={Proceedings of the 7th annual conference on Computer graphics and interactive techniques},
  pages={124--133},
  year={1980}
}

@article{bentley1975multidimensional,
  title={Multidimensional binary search trees used for associative searching},
  author={Bentley, Jon Louis},
  journal={Communications of the ACM},
  volume={18},
  number={9},
  pages={509--517},
  year={1975},
  publisher={ACM New York, NY, USA}
}

@article{snook2000simplified,
  title={Simplified 3D movement and pathfinding using navigation meshes},
  author={Snook, Greg},
  journal={Game programming gems},
  volume={1},
  number={1},
  pages={288--304},
  year={2000},
  publisher={Charles River Media Newton Centre, MA, USA}
}

@inproceedings{liu2025partfield,
  title={Partfield: Learning 3d feature fields for part segmentation and beyond},
  author={Liu, Minghua and Uy, Mikaela Angelina and Xiang, Donglai and Su, Hao and Fidler, Sanja and Sharp, Nicholas and Gao, Jun},
  booktitle=iccv,
  pages={9704--9715},
  year={2025}
}

@inproceedings{tang2025mv,
  title={Mv-dust3r+: Single-stage scene reconstruction from sparse views in 2 seconds},
  author={Tang, Zhenggang and Fan, Yuchen and Wang, Dilin and Xu, Hongyu and Ranjan, Rakesh and Schwing, Alexander and Yan, Zhicheng},
  booktitle=cvpr,
  pages={5283--5293},
  year={2025}
}

@misc{MetaAssetGen2,
  author       = {Rakesh Ranjan and Andrea Vedaldi and Mahima Gupta and Christopher Ocampo and Ocean Quigley},
  title        = {Introducing Meta 3D AssetGen 2.0: A new foundation model for 3D content creation},
  howpublished = {\url{https://developers.meta.com/horizon/blog/worlds/AssetGen2/}},
  year         = {2025},
  note         = {Accessed: \today}
}

@STRING{ai      = {Artificial Intelligence} }

@STRING{arxiv   = {arXiv} }

@STRING{corr    = {arXiv.cs} }

@STRING{cvpr    = {Proc. {CVPR}} }

@STRING{eccv    = {Proc. {ECCV}} }

@STRING{icassp  = {Proc. ICASSP} }

@STRING{iccv    = {Proc. {ICCV}} }

@STRING{iccvw   = {Proc. {ICCV} Workshops} }

@STRING{iclr    = {Proc. {ICLR}} }

@STRING{icml    = {Proc. {ICML}} }

@STRING{nips    = {Proc. {NeurIPS}} }

@STRING{siggraph= {Proc. {SIGGRAPH}} }

@STRING{texture = {Proc. Intl. Workshop on {Texture Analysis and Synthesis}} }

@STRING{threedv = {Proc. {3DV}} }

@InProceedings{lee25nuiscene:,
  author     = {Han-Hung Lee and Qinghong Han and Angel X. Chang},
  booktitle  = iccv,
  title      = {{NuiScene:} Exploring Efficient Generation of Unbounded
               Outdoor Scenes},
  year       = {2025}
}

@InProceedings{gumeli22roca:,
  author     = {G{\"u}meli, Can and Dai, Angela and Nie{\ss}ner, Matthias},
  booktitle  = cvpr,
  title      = {{ROCA}: Robust {CAD} model retrieval and alignment from a
               single image},
  year       = {2022}
}

@InProceedings{huang25midi:,
  author     = {Huang, Zehuan and Guo, Yuan-Chen and An, Xingqiao and Yang,
               Yunhan and Li, Yangguang and Zou, Zi-Xin and Liang, Ding and
               Liu, Xihui and Cao, Yan-Pei and Sheng, Lu},
  booktitle  = cvpr,
  title      = {{MIDI}: Multi-instance diffusion for single image to {3D}
               scene generation},
  year       = {2025}
}

@InProceedings{kuo21patch2cad:,
  author     = {Kuo, Weicheng and Angelova, Anelia and Lin, Tsung-Yi and Dai,
               Angela},
  booktitle  = iccv,
  title      = {{Patch2CAD}: Patchwise embedding learning for in-the-wild
               shape retrieval from a single image},
  year       = {2021}
}

@Article{dai24automated,
  author     = {Dai, Tianyuan and Wong, Josiah and Jiang, Yunfan and Wang,
               Chen and Gokmen, Cem and Zhang, Ruohan and Wu, Jiajun and
               Fei-Fei, Li},
  journal    = arxiv,
  title      = {Automated creation of digital cousins for robust policy
               learning},
  volume     = {2410.07408},
  year       = {2024}
}

@Article{jiang25anysplat:,
  author     = {Lihan Jiang and Yucheng Mao and Linning Xu and Tao Lu and
               Kerui Ren and Yichen Jin and Xudong Xu and Mulin Yu and
               Jiangmiao Pang and Feng Zhao and Dahua Lin and Bo Dai},
  journal    = arxiv,
  title      = {{AnySplat:} Feed-forward {3D Gaussian Splatting} from
               Unconstrained Views},
  volume     = {2505.23716},
  year       = {2025}
}

@Article{yang25fast3r:,
  author     = {Yang, Jianing and Sax, Alexander and Liang, Kevin J and
               Henaff, Mikael and Tang, Hao and Cao, Ang and Chai, Joyce and
               Meier, Franziska and Feiszli, Matt},
  journal    = cvpr,
  title      = {{Fast3R:} Towards {3D} Reconstruction of 1000+ Images in One
               Forward Pass},
  year       = {2025}
}

@InProceedings{maleki24procedural,
  author     = {Maleki, Mahdi Farrokhi and Zhao, Richard},
  booktitle  = {Proc. Conf. on Artificial Intelligence and Interactive
               Digital Entertainment},
  title      = {Procedural content generation in games: a survey with
               insights on emerging LLM integration},
  year       = {2024}
}

@Article{hendrikx13procedural,
  author     = {Hendrikx, Mark and Meijer, Sebastiaan and Van Der Velden,
               Joeri and Iosup, Alexandru},
  journal    = {ACM Trans. Multimedia Comput. Commun. Appl.},
  number     = {1},
  title      = {Procedural content generation for games: A survey},
  volume     = {9},
  year       = {2013}
}

@Article{wang25p3,
  author     = {Yifan Wang and Jianjun Zhou and Haoyi Zhu and Wenzheng Chang
               and Yang Zhou and Zizun Li and Junyi Chen and Jiangmiao Pang
               and Chunhua Shen and Tong He},
  journal    = arxiv,
  title      = {$\pi^3$: Permutation-Equivariant Visual Geometry Learning},
  volume     = {2507.13347},
  year       = {2025}
}

@Article{bahmani25lyra:,
  author     = {Sherwin Bahmani and Tianchang Shen and Jiawei Ren and Jiahui
               Huang and Yifeng Jiang and Haithem Turki and Andrea
               Tagliasacchi and David B. Lindell and Zan Gojcic and Sanja
               Fidler and Huan Ling and Jun Gao and Xuanchi Ren},
  journal    = arxiv,
  title      = {Lyra: Generative {3D} Scene Reconstruction via Video
               Diffusion Model Self-Distillation},
  volume     = {2509.19296},
  year       = {2025}
}

@Article{smart24splatt3r:,
  author     = {Brandon Smart and Chuanxia Zheng and Iro Laina and Victor
               Adrian Prisacariu},
  journal    = arxiv,
  title      = {{Splatt3R}: Zero-shot Gaussian Splatting from Uncalibrated
               Image Pairs},
  volume     = {2408.13912},
  year       = {2024}
}

@Article{kaneva10infinite,
  author     = {Kaneva, Biliana and Sivic, Josef and Torralba, Antonio and
               Avidan, Shai and Freeman, William T.},
  journal    = {Proceedings of the IEEE},
  number     = {8},
  title      = {Infinite Images: Creating and Exploring a Large
               Photorealistic Virtual Space},
  volume     = {98},
  year       = {2010}
}

@InProceedings{wei25omni-scene:,
  author     = {Dongxu Wei and Zhiqi Li and Peidong Liu},
  booktitle  = cvpr,
  title      = {Omni-Scene: Omni-Gaussian Representation for Ego-Centric
               Sparse-View Scene Reconstruction},
  year       = {2025}
}

@Article{szymanowicz25bolt3d:,
  author     = {Stanislaw Szymanowicz and Jason Y. Zhang and Pratul
               Srinivasan and Ruiqi Gao and Arthur Brussee and Aleksander
               Holynski and Ricardo Martin-Brualla and Jonathan T. Barron and
               Philipp Henzler},
  journal    = arxiv,
  title      = {{Bolt3D}: Generating {3D} Scenes in Seconds},
  volume     = {2503.14445},
  year       = {2025}
}

@Article{khan24autosplat:,
  author     = {Mustafa Khan and Hamidreza Fazlali and Dhruv Sharma and
               Tongtong Cao and Dongfeng Bai and Yuan Ren and Bingbing Liu},
  journal    = arxiv,
  title      = {{AutoSplat:} Constrained Gaussian Splatting for Autonomous
               Driving Scene Reconstruction},
  volume     = {2407.02598},
  year       = {2024}
}

@Article{zhou25hermes:,
  author     = {Xin Zhou and Dingkang Liang and Sifan Tu and Xiwu Chen and
               Yikang Ding and Dingyuan Zhang and Feiyang Tan and Hengshuang
               Zhao and Xiang Bai},
  journal    = arxiv,
  title      = {{HERMES:} A Unified Self-Driving World Model for Simultaneous
               {3D} Scene Understanding and Generation},
  volume     = {2501.14729},
  year       = {2025}
}

@Article{keetha25mapanything:,
  author     = {Nikhil Keetha and Norman M{\"u}ller and Johannes
               Sch{\"o}nberger and Lorenzo Porzi and Yuchen Zhang and Tobias
               Fischer and Arno Knapitsch and Duncan Zauss and Ethan Weber
               and Nelson Antunes and Jonathon Luiten and Manuel
               Lopez-Antequera and Samuel Rota Bul{\`o} and Christian
               Richardt and Deva Ramanan and Sebastian Scherer and Peter
               Kontschieder},
  journal    = arxiv,
  title      = {{MapAnything:} Universal Feed-Forward Metric {3D}
               Reconstruction},
  volume     = {2509.13414},
  year       = {2025}
}

@InProceedings{jang25pow3r:,
  author     = {Wonbong Jang and Philippe Weinzaepfel and Vincent Leroy and
               Lourdes Agapito and Jerome Revaud},
  booktitle  = cvpr,
  title      = {{Pow3R}: Empowering Unconstrained {3D} Reconstruction with
               Camera and Scene Priors},
  year       = {2025}
}

@InProceedings{ni25recondreamer:,
  author     = {Chaojun Ni and Guosheng Zhao and Xiaofeng Wang and Zheng Zhu
               and Wenkang Qin and Guan Huang and Chen Liu and Yuyin Chen and
               Yida Wang and Xueyang Zhang and Yifei Zhan and Kun Zhan and
               Peng Jia and Xianpeng Lang and Xingang Wang and Wenjun Mei},
  booktitle  = cvpr,
  title      = {{ReconDreamer:} Crafting World Models for Driving Scene
               Reconstruction via Online Restoration},
  year       = {2025}
}

@Article{zhao25recondreamer:,
  author     = {Guosheng Zhao and Xiaofeng Wang and Chaojun Ni and Zheng Zhu
               and Wenkang Qin and Guan Huang and Xingang Wang},
  journal    = {arXiv},
  title      = {{ReconDreamer++:} Harmonizing Generative and Reconstructive
               Models for Driving Scene Representation},
  volume     = {2503.18438},
  year       = {2025}
}

@Article{chatterjee243d-scene-former:,
  author     = {Chatterjee, Jit and Torres Vega, Maria},
  journal    = {The Visual Computer},
  title      = {{3D-Scene-Former}: {3D} scene generation from a single RGB
               image using Transformers},
  volume     = {41},
  year       = {2024}
}

@InProceedings{xing25luminet:,
  author     = {Xing, Xiaoyan and Groh, Konrad and Karaoglu, Sezer and
               Gevers, Theo and Bhattad, Anand},
  booktitle  = cvpr,
  title      = {LumiNet: Latent Intrinsics Meets Diffusion Models for Indoor
               Scene Relighting},
  year       = {2025}
}

@Article{xiang25gaussianroom:,
  author     = {Haodong Xiang and Xinghui Li and Kai Cheng and Xiansong Lai
               and Wanting Zhang and Zhichao Liao and Long Zeng and Xueping
               Liu},
  journal    = arxiv,
  title      = {{GaussianRoom:} Improving {3D Gaussian} Splatting with {SDF}
               Guidance and Monocular Cues for Indoor Scene Reconstruction},
  volume     = {2405.19671},
  year       = {2025}
}

@Article{liu25free4d:,
  author     = {Tianqi Liu and Zihao Huang and Zhaoxi Chen and Guangcong Wang
               and Shoukang Hu and Liao Shen and Huiqiang Sun and Zhiguo Cao
               and Wei Li and Ziwei Liu},
  journal    = arxiv,
  title      = {Free4D: Tuning-free 4D Scene Generation with Spatial-Temporal
               Consistency},
  volume     = {2503.20785},
  year       = {2025}
}

@InProceedings{li25uniscene:,
  author     = {Bohan Li and Jiazhe Guo and Hongsi Liu and Yingshuang Zou and
               Yikang Ding and Xiwu Chen and Hu Zhu and Feiyang Tan and Chi
               Zhang and Tiancai Wang and Shuchang Zhou and Li Zhang and
               Xiaojuan Qi and Hao Zhao and Mu Yang and Wenjun Zeng and Xin
               Jin},
  booktitle  = cvpr,
  title      = {{UniScene:} Unified Occupancy-centric Driving Scene
               Generation},
  year       = {2025}
}

@Article{yang25layerpano3d:,
  author     = {Shuai Yang and Jing Tan and Mengchen Zhang and Tong Wu and
               Yixuan Li and Gordon Wetzstein and Ziwei Liu and Dahua Lin},
  journal    = arxiv,
  title      = {{LayerPano3D:} Layered {3D} Panorama for Hyper-Immersive
               Scene Generation},
  volume     = {2408.13252},
  year       = {2025}
}

@Article{zhou25layoutdreamer:,
  author     = {Yang Zhou and Zongjin He and Qixuan Li and Chao Wang},
  journal    = arxiv,
  title      = {{LAYOUTDREAMER:} Physics-guided Layout for {Text-to-3D}
               Compositional Scene Generation},
  volume     = {2502.01949},
  year       = {2025}
}

@Article{guo25dist-4d:,
  author     = {Jiazhe Guo and Yikang Ding and Xiwu Chen and Shuo Chen and
               Bohan Li and Yingshuang Zou and Xiaoyang Lyu and Feiyang Tan
               and Xiaojuan Qi and Zhiheng Li and Hao Zhao},
  journal    = arxiv,
  title      = {{DiST-4D:} Disentangled Spatiotemporal Diffusion with Metric
               Depth for {4D} Driving Scene Generation},
  volume     = {2503.15208},
  year       = {2025}
}

@Article{hou25bloomscene:,
  author     = {Xiaolu Hou and Mingcheng Li and Dingkang Yang and Jiawei Chen
               and Ziyun Qian and Xiao Zhao and Yue Jiang and Jinjie Wei and
               Qingyao Xu and Lihua Zhang},
  journal    = arxiv,
  title      = {{BloomScene:} Lightweight Structured {3D Gaussian} Splatting
               for Crossmodal Scene Generation},
  volume     = {2501.10462},
  year       = {2025}
}

@Article{li25phip-g:,
  author     = {Qixuan Li and Chao Wang and Zongjin He and Yan Peng},
  journal    = arxiv,
  title      = {{PhiP-G:} Physics-Guided {Text-to-3D} Compositional Scene
               Generation},
  volume     = {2502.00708},
  year       = {2025}
}

@Article{duan25worldscore:,
  author     = {Haoyi Duan and Hong-Xing Yu and Sirui Chen and Li Fei-Fei and
               Jiajun Wu},
  journal    = arxiv,
  title      = {{WorldScore:} A Unified Evaluation Benchmark for World
               Generation},
  volume     = {2504.00983},
  year       = {2025}
}

@Article{mao25dreamdrive:,
  author     = {Jiageng Mao and Boyi Li and Boris Ivanovic and Yuxiao Chen
               and Yan Wang and Yurong You and Chaowei Xiao and Danfei Xu and
               Marco Pavone and Yue Wang},
  journal    = arxiv,
  title      = {{DreamDrive:} Generative {4D} Scene Modeling from Street View
               Images},
  volume     = {2501.00601},
  year       = {2025}
}

@InProceedings{feng25casagpt:,
  author     = {Feng, Weitao and Zhou, Hang and Liao, Jing and Cheng, Li and
               Zhou, Wenbo},
  booktitle  = cvpr,
  title      = {{CasaGPT}: Cuboid Arrangement and Scene Assembly for Interior
               Design},
  year       = {2025}
}

@Article{ni25wonderturbo:,
  author     = {Chaojun Ni and Xiaofeng Wang and Zheng Zhu and Weijie Wang
               and Haoyun Li and Guosheng Zhao and Jie Li and Wenkang Qin and
               Guan Huang and Wenjun Mei},
  journal    = arxiv,
  title      = {{WonderTurbo:} Generating Interactive {3D} World in 0.72
               Seconds},
  volume     = {2504.02261},
  year       = {2025}
}

@InProceedings{deng25citygen:,
  author     = {Jie Deng and Wenhao Chai and Jianshu Guo and Qixuan Huang and
               Junsheng Huang and Wenhao Hu and Shengyu Hao and Jenq-Neng
               Hwang and Gaoang Wang},
  booktitle  = cvpr,
  title      = {{CityGen:} Infinite and Controllable City Layout Generation},
  year       = {2025}
}

@Proceedings{fang25ctrl-room:,
  author     = {Chuan Fang and Xiaotao Hu and Kunming Luo and Ping Tan},
  journal    = threedv,
  title      = {{Ctrl-Room}: Controllable {Text-to-3D} Room Meshes Generation
               with Layout Constraints},
  year       = {2025}
}

@Article{imtiaz25lvt:,
  author     = {Tooba Imtiaz and Lucy Chai and Kathryn Heal and Xuan Luo and
               Jungyeon Park and Jennifer Dy and John Flynn},
  journal    = arxiv,
  title      = {{LVT:} Large-Scale Scene Reconstruction via Local View
               Transformers},
  volume     = {2509.25001},
  year       = {2025}
}

@InProceedings{zhao25drivedreamer4d:,
  author     = {Guosheng Zhao and Chaojun Ni and Xiaofeng Wang and Zheng Zhu
               and Xueyang Zhang and Yida Wang and Guan Huang and Xinze Chen
               and Boyuan Wang and Youyi Zhang and Wenjun Mei and Xingang
               Wang},
  booktitle  = cvpr,
  title      = {{DriveDreamer4D:} World Models Are Effective Data Machines
               for 4D Driving Scene Representation},
  year       = {2025}
}

@Article{meng25scenegen:,
  author     = {Yanxu Meng and Haoning Wu and Ya Zhang and Weidi Xie},
  journal    = arxiv,
  title      = {{SceneGen:} Single-Image {3D} Scene Generation in One
               Feedforward Pass},
  volume     = {2508.15769},
  year       = {2025}
}

@Article{tang25efficient,
  author     = {Jiaxiang Tang and Ruijie Lu and Zhaoshuo Li and Zekun Hao and
               Xuan Li and Fangyin Wei and Shuran Song and Gang Zeng and
               Ming-Yu Liu and Tsung-Yi Lin},
  journal    = arxiv,
  title      = {Efficient Part-level {3D} Object Generation via Dual Volume
               Packing},
  volume     = {2506.09980},
  year       = {2025}
}

@Article{ran25direct,
  author     = {Xingjian Ran and Yixuan Li and Linning Xu and Mulin Yu and Bo
               Dai},
  journal    = arxiv,
  title      = {Direct Numerical Layout Generation for 3D Indoor Scene
               Synthesis via Spatial Reasoning},
  volume     = {2506.05341},
  year       = {2025}
}

@InProceedings{wang25decoupledgaussian:,
  author     = {Miaowei Wang and Yibo Zhang and Weiwei Xu and Rui Ma and
               Changqing Zou and Daniel Morris},
  booktitle  = cvpr,
  title      = {{DecoupledGaussian:} Object-Scene Decoupling for
               Physics-Based Interaction},
  year       = {2025}
}

@Article{yang25instascene:,
  author     = {Zesong Yang and Bangbang Yang and Wenqi Dong and Chenxuan Cao
               and Liyuan Cui and Yuewen Ma and Zhaopeng Cui and Hujun Bao},
  journal    = arxiv,
  title      = {{InstaScene:} Towards Complete {3D} Instance Decomposition
               and Reconstruction from Cluttered Scenes},
  volume     = {2507.08416},
  year       = {2025}
}

@Article{wang254real-video-v2:,
  author     = {Chaoyang Wang and Ashkan Mirzaei and Vidit Goel and Willi
               Menapace and Aliaksandr Siarohin and Avalon Vinella and
               Michael Vasilkovsky and Ivan Skorokhodov and Vladislav
               Shakhrai and Sergey Korolev and Sergey Tulyakov and Peter
               Wonka},
  journal    = arxiv,
  title      = {4Real-Video-V2: Fused View-Time Attention and Feedforward
               Reconstruction for 4D Scene Generation},
  volume     = {2506.18839},
  year       = {2025}
}

@Article{li25sparc3d:,
  author     = {Zhihao Li and Yufei Wang and Heliang Zheng and Yihao Luo and
               Bihan Wen},
  journal    = arxiv,
  title      = {{Sparc3D}: Sparse Representation and Construction for
               High-Resolution 3D Shapes Modeling},
  volume     = {2505.14521},
  year       = {2025}
}

@Article{zhai25stargen:,
  author     = {Shangjin Zhai and Zhichao Ye and Jialin Liu and Weijian Xie
               and Jiaqi Hu and Zhen Peng and Hua Xue and Danpeng Chen and
               Xiaomeng Wang and Lei Yang and Nan Wang and Haomin Liu and
               Guofeng Zhang},
  journal    = arxiv,
  title      = {{StarGen:} A Spatiotemporal Autoregression Framework with
               Video Diffusion Model for Scalable and Controllable Scene
               Generation},
  volume     = {2501.05763},
  year       = {2025}
}

@Article{wen253d-scene,
  author     = {Beichen Wen and Haozhe Xie and Zhaoxi Chen and Fangzhou Hong
               and Ziwei Liu},
  journal    = axiv,
  title      = {{3D} Scene Generation: A Survey},
  volume     = {2505.05474},
  year       = {2025}
}

@Article{chen23scenedreamer:,
  author     = {Z. Chen and G. Wang and Z. Liu},
  journal    = pami,
  number     = {12},
  title      = {{SceneDreamer}: Unbounded {3D} Scene Generation From {2D}
               Image Collections},
  volume     = {45},
  year       = {2023}
}

@Article{ling25scenethesis:,
  author     = {Lu Ling and Chen-Hsuan Lin and Tsung-Yi Lin and Yifan Ding
               and Yu Zeng and Yichen Sheng and Yunhao Ge and Ming-Yu Liu and
               Aniket Bera and Zhaoshuo Li},
  journal    = arxiv,
  title      = {Scenethesis: A Language and Vision Agentic Framework for {3D}
               Scene Generation},
  volume     = {2505.02836},
  year       = {2025}
}

@InProceedings{dahnert24coherent,
  author     = {Manuel Dahnert and Angela Dai and Norman M{\"u}ller and
               Matthias Niessner},
  booktitle  = nips,
  title      = {Coherent {3D} Scene Diffusion From a Single {RGB} Image},
  year       = {2024}
}

@InProceedings{lin24genusd:,
  author     = {Lin, Tsung-Yi and Lin, Chen-Hsuan and Cui, Yin and Ge, Yunhao
               and Nah, Seungjun and Mallya, Arun and Hao, Zekun and Ding,
               Yifan and Mao, Hanzi and Li, Zhaoshuo and Lin, Yen-Chen and
               Zeng, Xiaohui and Zhang, Qinsheng and Xiang, Donglai and Ma,
               Qianli and Lewis, J.P. and Jin, Jingyi and Jannaty, Pooya and
               Liu, Ming-Yu},
  booktitle  = siggraph,
  title      = {{GenUSD}: {3D} scene generation made easy},
  year       = {2024}
}

@Article{sun25hierarchically-structured,
  author     = {Weilin Sun and Xinran Li and Manyi Li and Kai Xu and Xiangxu
               Meng and Lei Meng},
  journal    = arxiv,
  title      = {Hierarchically-Structured Open-Vocabulary Indoor Scene
               Synthesis with Pre-trained Large Language Model},
  volume     = {2502.10675},
  year       = {2025}
}

@Article{aguina-kang24open-universe,
  author     = {Rio Aguina-Kang and Maxim Gumin and Do Heon Han and Stewart
               Morris and Seung Jean Yoo and Aditya Ganeshan and R. Kenny
               Jones and Qiuhong Anna Wei and Kailiang Fu and Daniel
               Ritchie},
  journal    = {arXiv},
  title      = {Open-Universe Indoor Scene Generation using {LLM} Program
               Synthesis and Uncurated Object Databases},
  volume     = {2403.09675},
  year       = {2024}
}

@Article{huang25roompainter:,
  author     = {Zhipeng Huang and Wangbo Yu and Xinhua Cheng and ChengShu
               Zhao and Yunyang Ge and Mingyi Guo and Li Yuan and Yonghong
               Tian},
  journal    = arxiv,
  title      = {{RoomPainter:} View-Integrated Diffusion for Consistent
               Indoor Scene Texturing},
  volume     = {2412.16778},
  year       = {2025}
}

@InProceedings{feng25text-guided,
  author     = {Feng, Yuchuan and Jiang, Jihang and Ren, Jie and Li, Wenrui
               and Li, Ruotong and Fan, Xiaopeng},
  booktitle  = icassp,
  title      = {Text-Guided Editable 3D City Scene Generation},
  year       = {2025}
}

@InProceedings{raistrick24infinigen,
  author     = {Alexander Raistrick and Lingjie Mei and Karhan Kayan and
               David Yan and Yiming Zuo and Beining Han and Hongyu Wen and
               Meenal Parakh and Stamatis Alexandropoulos and Lahav Lipson
               and Zeyu Ma and Jia Deng},
  booktitle  = cvpr,
  title      = {Infinigen Indoors: Photorealistic Indoor Scenes using
               Procedural Generation},
  year       = {2024}
}

@Article{li24scenedreamer360:,
  author     = {Wenrui Li and Fucheng Cai and Yapeng Mi and Zhe Yang and
               Wangmeng Zuo and Xingtao Wang and Xiaopeng Fan},
  journal    = arxiv,
  title      = {{SceneDreamer360:} Text-Driven {3D}-Consistent Scene
               Generation with Panoramic {Gaussian} Splatting},
  volume     = {2408.13711},
  year       = {2024}
}

@InProceedings{zhou24dreamscene360:,
  author     = {Shijie Zhou and Zhiwen Fan and Dejia Xu and Haoran Chang and
               Pradyumna Chari and Tejas K Bharadwaj and Suya You and
               Zhangyang Wang and Achuta Kadambi},
  booktitle  = eccv,
  title      = {{DreamScene360:} Unconstrained {Text-to-3D} Scene Generation
               with Panoramic {Gaussian} Splatting},
  year       = {2024}
}

@Article{yu25wonderworld:,
  author     = {Hong-Xing Yu and Haoyi Duan and Charles Herrmann and William
               T. Freeman and Jiajun Wu},
  journal    = arxiv,
  title      = {{WonderWorld:} Interactive {3D} Scene Generation from a
               Single Image},
  volume     = {2406.09394},
  year       = {2025}
}

@InProceedings{shi223d-aware,
  author     = {Zifan Shi and Yujun Shen and Jiapeng Zhu and Dit-Yan Yeung
               and Qifeng Chen},
  booktitle  = eccv,
  title      = {{3D}-Aware Indoor Scene Synthesis with Depth Priors},
  year       = {2022}
}

@InProceedings{zhang24berfscene:,
  author     = {Qihang Zhang and Yinghao Xu and Yujun Shen and Bo Dai and
               Bolei Zhou and Ceyuan Yang},
  booktitle  = cvpr,
  title      = {{BerfScene:} Bev-conditioned Equivariant Radiance Fields for
               Infinite {3D} Scene Generation},
  year       = {2024}
}

@InProceedings{zhang243d-scenedreamer:,
  author     = {Songchun Zhang and Yibo Zhang and Quan Zheng and Rui Ma and
               Wei Hua and Hujun Bao and Weiwei Xu and Changqing Zou},
  booktitle  = cvpr,
  title      = {{3D-SceneDreamer:} Text-Driven {3D}-Consistent Scene
               Generation},
  year       = {2024}
}

@InProceedings{lin24instructscene:,
  author     = {Chenguo Lin and Yadong MU},
  booktitle  = {International Conference on Learning Representations (ICLR)},
  title      = {{InstructScene:} Instruction-Driven 3D Indoor Scene Synthesis
               with Semantic Graph Prior},
  year       = {2024}
}

@InProceedings{tang24diffuscene:,
  author     = {Jiapeng Tang and Yinyu Nie and Lev Markhasin and Angela Dai
               and Justus Thies and Matthias Nie{\ss}ner},
  booktitle  = cvpr,
  title      = {{DiffuScene:} Denoising Diffusion Models for Generative
               Indoor Scene Synthesis},
  year       = {2024}
}

@InProceedings{yang24physcene:,
  author     = {Yandan Yang and Baoxiong Jia and Peiyuan Zhi and Siyuan
               Huang},
  booktitle  = cvpr,
  title      = {{PhyScene:} Physically Interactable 3D Scene Synthesis for
               Embodied {AI}},
  year       = {2024}
}

@InProceedings{zhao23roomdesigner:,
  author     = {Yiqun Zhao and Zibo Zhao and Jing Li and Sixun Dong and
               Shenghua Gao},
  booktitle  = arxiv,
  title      = {{RoomDesigner:} Encoding Anchor-latents for Style-consistent
               and Shape-compatible Indoor Scene Generation},
  year       = {2023}
}

@InProceedings{wang18deep-conv,
  author     = {Kai Wang and M. Savva and Angel X. Chang and Daniel Ritchie},
  booktitle  = siggraph,
  title      = {Deep convolutional priors for indoor scene synthesis},
  year       = {2018}
}

@Article{xie25generative,
  author     = {Haozhe Xie and Zhaoxi Chen and Fangzhou Hong and Ziwei Liu},
  journal    = arxiv,
  title      = {Generative {Gaussian} Splatting for Unbounded {3D} City
               Generation},
  volume     = {2406.06526},
  year       = {2025}
}

@InProceedings{zhang21fast,
  author     = {Song-Hai Zhang and Shaokui Zhang and Wei-Yu Xie and
               Cheng-Yang Luo and Yong-Liang Yang and Hongbo Fu},
  booktitle  = {{IEEE} Trans. on on Visualization and Computer Graphics},
  title      = {Fast {3D} Indoor Scene Synthesis by Learning Spatial Relation
               Priors of Objects},
  year       = {2021}
}

@InProceedings{ocal24sceneteller:,
  author     = {Basak Melis Ocal and Maxim Tatarchenko and Sezer Karaoglu and
               Theo Gevers},
  booktitle  = eccv,
  title      = {{SceneTeller:} Language-to-{3D} Scene Generation},
  year       = {2024}
}

@InProceedings{zhang24scenewiz3d:,
  author     = {Qihang Zhang and Chaoyang Wang and Aliaksandr Siarohin and
               Peiye Zhuang and Yinghao Xu and Ceyuan Yang and Dahua Lin and
               Bolei Zhou and S. Tulyakov and Hsin-Ying Lee},
  booktitle  = cvpr,
  title      = {{SceneWiz3D:} Towards Text-guided {3D} Scene Composition},
  year       = {2024}
}

@Article{dong25hiscene:,
  author     = {Wenqi Dong and Bangbang Yang and Zesong Yang and Yuan Li and
               Tao Hu and Hujun Bao and Yuewen Ma and Zhaopeng Cui},
  journal    = arxiv,
  title      = {{HiScene:} Creating Hierarchical {3D} Scenes with Isometric
               View Generation},
  volume     = {2504.13072},
  year       = {2025}
}

@Article{liu25controllable,
  author     = {Yuheng Liu and Xinke Li and Yuning Zhang and Lu Qi and Xin Li
               and Wenping Wang and Chongshou Li and Xueting Li and
               Ming-Hsuan Yang},
  journal    = arxiv,
  title      = {Controllable {3D} Outdoor Scene Generation via Scene Graphs},
  volume     = {2503.07152},
  year       = {2025}
}

@InProceedings{zhai23commonscenes:,
  author     = {Guangyao Zhai and Evin Pinar {\"O}rnek and Shun-Cheng Wu and
               Yan Di and Federico Tombari and Nassir Navab and Benjamin
               Busam},
  booktitle  = nips,
  title      = {{CommonScenes:} Generating Commonsense {3D} Indoor Scenes
               with Scene Graphs},
  year       = {2023}
}

@Article{yang25mmgdreamer:,
  author     = {Zhifei Yang and Keyang Lu and Chao Zhang and Jiaxing Qi and
               Hanqi Jiang and Ruifei Ma and Shenglin Yin and Yifan Xu and
               Mingzhe Xing and Zhen Xiao and Jieyi Long and Guangyao Zhai},
  journal    = arxiv,
  title      = {{MMGDreamer:} Mixed-Modality Graph for Geometry-Controllable
               3D Indoor Scene Generation},
  volume     = {2502.05874},
  year       = {2025}
}

@InProceedings{zhai24echoscene:,
  author     = {Guangyao Zhai and Evin Pinar {\"O}rnek and Dave Zhenyu Chen
               and Ruotong Liao and Yan Di and Nassir Navab and Federico
               Tombari and Benjamin Busam},
  booktitle  = eccv,
  title      = {{EchoScene:} Indoor Scene Generation via Information Echo
               over Scene Graph Diffusion},
  year       = {2024}
}

@InProceedings{gao24graphdreamer:,
  author     = {Gege Gao and Weiyang Liu and Anpei Chen and Andreas Geiger
               and Bernhard Sch{\"o}lkopf},
  booktitle  = cvpr,
  title      = {{GraphDreamer:} Compositional {3D} Scene Synthesis from Scene
               Graphs},
  year       = {2024}
}

@Article{tang25recent,
  author     = {Xiang Tang and Ruotong Li and Xiaopeng Fan},
  journal    = arxiv,
  title      = {Recent Advance in {3D} Object and Scene Generation: A
               Survey},
  volume     = {2504.11734},
  year       = {2025}
}

@Article{chang14interactive,
  author     = {Angel X. Chang and M. Savva and Christopher D. Manning},
  journal    = arxiv,
  title      = {Interactive Learning of Spatial Knowledge for Text to {3D}
               Scene Generation},
  year       = {2014}
}

@InProceedings{qiao22learning,
  author     = {Xiaotian Qiao and Gerhard P. Hancke and Rynson W.H. Lau},
  booktitle  = cvpr,
  title      = {Learning Object Context for Novel-View Scene Layout
               Generation},
  year       = {2022}
}

@Article{tam24scenemotifcoder:,
  author     = {Hou In Ivan Tam and Hou In Derek Pun and Austin T. Wang and
               Angel X. Chang and Manolis Savva},
  journal    = arxiv,
  title      = {{SceneMotifCoder:} Example-driven Visual Program Learning for
               Generating {3D} Object Arrangements},
  volume     = {2408.02211},
  year       = {2024}
}

@Article{xu24sketch2scene:,
  author     = {Yongzhi Xu and Yonhon Ng and Yifu Wang and Inkyu Sa and
               Yunfei Duan and Yang Li and Pan Ji and Hongdong Li},
  journal    = arixv,
  title      = {Sketch2Scene: Automatic Generation of Interactive 3D Game
               Scenes from Users Casual Sketches},
  volume     = {2408.04567},
  year       = {2024}
}

@Article{schwarz25generative,
  author     = {Katja Schwarz and Norman Mueller and Peter Kontschieder},
  journal    = arxiv,
  title      = {{Generative Gaussian Splatting}: Generating {3D} Scenes with
               Video Diffusion Priors},
  volume     = {2503.13272},
  year       = {2025}
}

@Article{ren25gen3c:,
  author     = {Xuanchi Ren and Tianchang Shen and Jiahui Huang and Huan Ling
               and Yifan Lu and Merlin Nimier-David and Thomas M{\"u}ller and
               Alexander Keller and Sanja Fidler and Jun Gao},
  journal    = {arXiv},
  title      = {{GEN3C:} {3D}-Informed World-Consistent Video Generation with
               Precise Camera Control},
  volume     = {2503.03751},
  year       = {2025}
}

@InProceedings{vermandere24semantic,
  author     = {Jelle Vermandere and Maarten Bassier and Maarten Vergauwen},
  booktitle  = {Proc. {Computer Graphics \& Visual Computing}},
  title      = {Semantic {UV} mapping to improve texture inpainting for
               indoor scenes},
  volume     = {2407.09248},
  year       = {2024}
}

@Article{bokhovkin24scenefactor:,
  author     = {Alexey Bokhovkin and Quan Meng and Shubham Tulsiani and
               Angela Dai},
  journal    = arxiv,
  title      = {{SceneFactor:} Factored Latent {3D} Diffusion for
               Controllable 3D Scene Generation},
  volume     = {2412.01801},
  year       = {2024}
}

@InProceedings{liu21infinite,
  author     = {A Liu and R Tucker and V Jampani and A Makadia and N
               Snavely{\ldots}},
  booktitle  = iccv,
  title      = {Infinite Nature: Perpetual View Generation of Natural Scenes
               from a Single Image},
  year       = {2021}
}

@InProceedings{yin21learning,
  author     = {Yin, Wei and Zhang, Jianming and Wang, Oliver and Niklaus,
               Simon and Mai, Long and Chen, Simon and Shen, Chunhua},
  booktitle  = cvpr,
  title      = {Learning to recover {3D} scene shape from a single image},
  year       = {2021}
}

@Misc{parker-holder24genie,
  author     = {Jack Parker-Holder and Philip Ball and Jake Bruce and
               Vibhavari Dasagi and Kristian Holsheimer and Christos Kaplanis
               and Alexandre Moufarek and Guy Scully and Jeremy Shar and
               Jimmy Shi and Stephen Spencer and Jessica Yung and Michael
               Dennis and Sultan Kenjeyev and Shangbang Long and Vlad Mnih
               and Harris Chan and Maxime Gazeau and Bonnie Li and Fabio
               Pardo and Luyu Wang and Lei Zhang and Frederic Besse and Tim
               Harley and Anna Mitenkova and Jane Wang and Jeff Clune and
               Demis Hassabis and Raia Hadsell and Adrian Bolton and Satinder
               Singh and Tim Rockt{\"a}schel},
  title      = {Genie 2: A Large-Scale Foundation World Model},
  url        = {https://deepmind.google/discover/blog/genie-2-a-large-scale-foundation-world-model/},
  year       = {2024}
}

@Article{wu24diorama:,
  author     = {Qirui Wu and Denys Iliash and Daniel Ritchie and Manolis
               Savva and Angel X. Chang},
  journal    = arxiv,
  title      = {Diorama: Unleashing Zero-shot Single-view {3D} Scene
               Modeling},
  volume     = {abs/2411.19492},
  year       = {2024}
}

@Article{yao25cast:,
  author     = {Kaixin Yao and Longwen Zhang and Xinhao Yan and Yan Zeng and
               Qixuan Zhang and Lan Xu and Wei Yang and Jiayuan Gu and Jingyi
               Yu},
  journal    = arxiv,
  title      = {{CAST:} Component-Aligned {3D} Scene Reconstruction from an
               {RGB} Image},
  volume     = {2502.12894},
  year       = {2025}
}

@InProceedings{zhang233dshape2vecset:,
  author     = {Biao Zhang and Jiapeng Tang and Matthias Niessner and Peter
               Wonka},
  booktitle  = {ACM Transactions on Graphics},
  title      = {{3DShape2VecSet}: A {3D} Shape Representation for Neural
               Fields and Generative Diffusion Models},
  year       = {2023}
}

@InProceedings{shriram25realmdreamer:,
  author     = {Jaidev Shriram and Alex Trevithick and Lingjie Liu and Ravi
               Ramamoorthi},
  booktitle  = threedv,
  title      = {{RealmDreamer:} Text-Driven 3D Scene Generation with
               Inpainting and Depth Diffusion},
  year       = {2025}
}

@Article{gao24diffcad:,
  author     = {Daoyi Gao and D{\'a}vid Rozenberszki and Stefan Leutenegger
               and Angela Dai},
  journal    = arxiv,
  title      = {{DiffCAD:} Weakly-Supervised Probabilistic {CAD} Model
               Retrieval and Alignment from an {RGB} Image},
  volume     = {2311.18610},
  year       = {2024}
}

@Article{zhou24scenex:,
  author     = {Mengqi Zhou and Yuxi Wang and Jun Hou and Shougao Zhang and
               Yiwei Li and Chuanchen Luo and Junran Peng and Zhaoxiang
               Zhang},
  journal    = arxiv,
  title      = {{SceneX:} Procedural Controllable Large-scale Scene
               Generation},
  volume     = {2403.15698},
  year       = {2024}
}

@Article{liu24reconx:,
  author     = {Fangfu Liu and Wenqiang Sun and Hanyang Wang and Yikai Wang
               and Haowen Sun and Junliang Ye and Jun Zhang and Yueqi Duan},
  journal    = arxiv,
  title      = {{ReconX:} Reconstruct Any Scene from Sparse Views with Video
               Diffusion Model},
  volume     = {2408.16767},
  year       = {2024}
}

@Article{dogaru24generalizable,
  author     = {Andreea Dogaru and Mert {\"O}zer and Bernhard Egger},
  journal    = arxiv,
  title      = {Generalizable {3D} Scene Reconstruction via Divide and
               Conquer from a Single View},
  volume     = {2404.03421},
  year       = {2024}
}

@Article{yang25prometheus:,
  author     = {Yuanbo Yang and Jiahao Shao and Xinyang Li and Yujun Shen and
               Andreas Geiger and Yiyi Liao},
  journal    = arxiv,
  title      = {Prometheus: {3D}-Aware Latent Diffusion Models for
               Feed-Forward Text-to-3D Scene Generation},
  volume     = {2412.21117},
  year       = {2025}
}

@Article{liang24wonderland:,
  author     = {Hanwen Liang and Junli Cao and Vidit Goel and Guocheng Qian
               and Sergei Korolev and Demetri Terzopoulos and Konstantinos N.
               Plataniotis and Sergey Tulyakov and Jian Ren},
  journal    = arxiv,
  title      = {Wonderland: Navigating {3D} Scenes from a Single Image},
  volume     = {2412.12091},
  year       = {2024}
}

@Article{xiang24structured,
  author     = {Jianfeng Xiang and Zelong Lv and Sicheng Xu and Yu Deng and
               Ruicheng Wang and Bowen Zhang and Dong Chen and Xin Tong and
               Jiaolong Yang},
  journal    = arxiv,
  title      = {Structured {3D} Latents for Scalable and Versatile 3D
               Generation},
  volume     = {2412.01506},
  year       = {2024}
}

@Article{sun24dimensionx:,
  author     = {Wenqiang Sun and Shuo Chen and Fangfu Liu and Zilong Chen and
               Yueqi Duan and Jun Zhang and Yikai Wang},
  journal    = {arXiv},
  title      = {{DimensionX:} Create Any {3D} and {4D} Scenes from a Single
               Image with Controllable Video Diffusion},
  volume     = {2411.04928},
  year       = {2024}
}

@Article{duisterhof24mast3r-sfm:,
  author     = {Bardienus Duisterhof and Lojze Zust and Philippe Weinzaepfel
               and Vincent Leroy and Yohann Cabon and Jerome Revaud},
  journal    = arxiv,
  title      = {{MASt3R-SfM:} a Fully-Integrated Solution for Unconstrained
               Structure-from-Motion},
  volume     = {2409.19152},
  year       = {2024}
}

@Article{ye24no-pose,
  author     = {Botao Ye and Sifei Liu and Haofei Xu and Xueting Li and Marc
               Pollefeys and Ming-Hsuan Yang and Songyou Peng},
  journal    = arxiv,
  title      = {No Pose, No Problem: Surprisingly Simple {3D Gaussian} Splats
               from Sparse Unposed Images},
  volume     = {2410.24207},
  year       = {2024}
}

@Article{zhang24monst3r:,
  author     = {Junyi Zhang and Charles Herrmann and Junhwa Hur and Varun
               Jampani and Trevor Darrell and Forrester Cole and Deqing Sun
               and Ming-Hsuan Yang},
  journal    = arxiv,
  title      = {{MonST3R:} A Simple Approach for Estimating Geometry in the
               Presence of Motion},
  volume     = {2410.03825},
  year       = {2024}
}

@InProceedings{hu24scenecraft:,
  author     = {Ziniu Hu and Ahmet Iscen and Aashi Jain and Thomas Kipf and
               Yisong Yue and David A. Ross and Cordelia Schmid and Alireza
               Fathi},
  booktitle  = icml,
  title      = {{SceneCraft}: An {LLM} Agent for Synthesizing 3D Scenes as
               {Blender} Code},
  year       = {2024}
}

@Article{meng24lt3sd:,
  author     = {Quan Meng and Lei Li and Matthias Nie{\ss}ner and Angela Dai},
  journal    = arxiv,
  title      = {{LT3SD:} Latent Trees for {3D} Scene Diffusion},
  volume     = {2409.08215},
  year       = {2024}
}

@Article{li24director3d:,
  author     = {Xinyang Li and Zhangyu Lai and Linning Xu and Yansong Qu and
               Liujuan Cao and Shengchuan Zhang and Bo Dai and Rongrong Ji},
  journal    = arxiv,
  title      = {{Director3D}: Real-world Camera Trajectory and {3D} Scene
               Generation from Text},
  volume     = {2406.17601},
  year       = {2024}
}

@InProceedings{zhang24clay:,
  author     = {Longwen Zhang and Ziyu Wang and Qixuan Zhang and Qiwei Qiu
               and Anqi Pang and Haoran Jiang and Wei Yang and Lan Xu and
               Jingyi Yu},
  booktitle  = siggraph,
  title      = {{CLAY}: A Controllable Large-scale Generative Model for
               Creating High-quality {3D} Assets},
  year       = {2024}
}

@Misc{tripo3d24text-to-3d,
  author     = {TripoAI},
  title      = {{Tripo3D} Text-to-{3D}},
  url        = {https://www.tripo3d.ai},
  year       = {2024}
}

@Misc{deemos24rodin,
  author     = {Deemos},
  title      = {Rodin Text-to-{3D} Gen-1 (0525) V0.5},
  url        = {https://hyperhuman.deemos.com/rodin},
  year       = {2024}
}

@Article{gao24magicdrive3d:,
  author     = {Ruiyuan Gao and Kai Chen and Zhihao Li and Lanqing Hong and
               Zhenguo Li and Qiang Xu},
  journal    = {arXiv},
  title      = {{MagicDrive3D:} Controllable {3D} Generation for Any-View
               Rendering in Street Scenes},
  volume     = {2405.14475},
  year       = {2024}
}

@Article{gao24cat3d:,
  author     = {Ruiqi Gao and Aleksander Holynski and Philipp Henzler and
               Arthur Brussee and Ricardo Martin-Brualla and Pratul
               Srinivasan and Jonathan T. Barron and Ben Poole},
  journal    = arxiv,
  title      = {{CAT3D:} Create Anything in 3D with Multi-View Diffusion
               Models},
  volume     = {2405.10314},
  year       = {2024}
}

@Article{chen24mvsplat:,
  author     = {Yuedong Chen and Haofei Xu and Chuanxia Zheng and Bohan
               Zhuang and Marc Pollefeys and Andreas Geiger and Tat-Jen Cham
               and Jianfei Cai},
  journal    = arxiv,
  title      = {{MVSplat:} Efficient {3D} Gaussian Splatting from Sparse
               Multi-View Images},
  volume     = {2403.14627},
  year       = {2024}
}

@Article{zhou23scene-conditional,
  author     = {Jinghao Zhou and Tomas Jakab and Philip Torr and Christian
               Rupprecht},
  journal    = arxiv,
  title      = {Scene-Conditional {3D} Object Stylization and Composition},
  volume     = {2312.12419},
  year       = {2023}
}

@Article{chen24v3d:,
  author     = {Zilong Chen and Yikai Wang and Feng Wang and Zhengyi Wang and
               Huaping Liu},
  journal    = arxiv,
  title      = {{V3D}: Video Diffusion Models are Effective {3D} Generators},
  volume     = {2403.06738},
  year       = {2024}
}

@Article{epstein24disentangled,
  author     = {Dave Epstein and Ben Poole and Ben Mildenhall and Alexei A.
               Efros and Aleksander Holynski},
  journal    = arxiv,
  title      = {Disentangled {3D} Scene Generation with Layout Learning},
  volume     = {2402.16936},
  year       = {2024}
}

@Article{stan23ldm3d:,
  author     = {Gabriela Ben Melech Stan and Diana Wofk and Scottie Fox and
               Alex Redden and Will Saxton and Jean Yu and Estelle Aflalo and
               Shao-Yen Tseng and Fabio Nonato and Matthias Muller and
               Vasudev Lal},
  journal    = corr,
  title      = {{LDM3D}: Latent Diffusion Model for {3D}},
  volume     = {2305.10853},
  year       = {2023}
}

@Article{shen24neural,
  author     = {Shihao Shen and Louis Kerofsky and Varun Ravi Kumar and
               Senthil Yogamani},
  journal    = arxiv,
  number     = {2402.06826},
  title      = {Neural Rendering based Urban Scene Reconstruction for
               Autonomous Driving},
  year       = {2024}
}

@Article{zhou24gala3d:,
  author     = {Xiaoyu Zhou and Xingjian Ran and Yajiao Xiong and Jinlin He
               and Zhiwei Lin and Yongtao Wang and Deqing Sun and
               Ming{-}Hsuan Yang},
  journal    = corr,
  title      = {{GALA3D}: Towards Text-to-{3D} Complex Scene Generation via
               Layout-guided Generative Gaussian Splatting},
  volume     = {abs/2402.07207},
  year       = {2024}
}

@InProceedings{wiles20synsin:,
  author     = {Olivia Wiles and Georgia Gkioxari and Richard Szeliski and
               Justin Johnson},
  booktitle  = cvpr,
  title      = {SynSin: End-to-End View Synthesis From a Single Image},
  year       = {2020}
}

@InProceedings{tulsiani18layer-structured,
  author     = {Shubham Tulsiani and Richard Tucker and Noah Snavely},
  booktitle  = eccv,
  title      = {Layer-Structured 3D Scene Inference via View Synthesis},
  year       = {2018}
}

@Article{ouyang23text2immersion:,
  author     = {Hao Ouyang and Kathryn Heal and Stephen Lombardi and
               Tiancheng Sun},
  journal    = corr,
  title      = {{Text2Immersion}: Generative Immersive Scene with {3D}
               Gaussians},
  volume     = {abs/2312.09242},
  year       = {2023}
}

@Article{yu23wonderjourney:,
  author     = {Hong{-}Xing Yu and Haoyi Duan and Junhwa Hur and Kyle Sargent
               and Michael Rubinstein and William T. Freeman and Forrester
               Cole and Deqing Sun and Noah Snavely and Jiajun Wu and Charles
               Herrmann},
  journal    = corr,
  title      = {WonderJourney: Going from Anywhere to Everywhere},
  volume     = {abs/2312.03884},
  year       = {2023}
}

@Article{zhang23text2nerf:,
  author     = {Jingbo Zhang and Xiaoyu Li and Ziyu Wan and Can Wang and Jing
               Liao},
  journal    = corr,
  title      = {{Text2NeRF}: Text-Driven {3D} Scene Generation with Neural
               Radiance Fields},
  volume     = {abs/2305.11588},
  year       = {2023}
}

@InProceedings{hollein23text2room:,
  author     = {Lukas H{\"{o}}llein and Ang Cao and Andrew Owens and Justin
               Johnson and Matthias Nie{\ss}ner},
  booktitle  = iccv,
  title      = {{Text2Room}: Extracting Textured {3D} Meshes from {2D}
               Text-to-Image Models},
  year       = {2023}
}

@InProceedings{cai23diffdreamer:,
  author     = {Shengqu Cai and Eric Ryan Chan and Songyou Peng and Mohamad
               Shahbazi and Anton Obukhov and Luc Van Gool and Gordon
               Wetzstein},
  booktitle  = iccv,
  title      = {{DiffDreamer}: Towards Consistent Unsupervised Single-view
               Scene Extrapolation with Conditional Diffusion Models},
  year       = {2023}
}

@InProceedings{rockwell21pixelsynth:,
  author     = {Chris Rockwell and David F. Fouhey and Justin Johnson},
  booktitle  = iccv,
  title      = {{PixelSynth}: Generating a {3D}-Consistent Experience from a
               Single Image},
  year       = {2021}
}

@Article{wu24blockfusion:,
  author     = {Zhennan Wu and Yang Li and Han Yan and Taizhang Shang and
               Weixuan Sun and Senbo Wang and Ruikai Cui and Weizhe Liu and
               Hiroyuki Sato and Hongdong Li and Pan Ji},
  journal    = corr,
  title      = {{BlockFusion}: Expandable {3D} Scene Generation using Latent
               Tri-plane Extrapolation},
  year       = {2024}
}

@InProceedings{lei23rgbd2:,
  author     = {Jiabao Lei and Jiapeng Tang and Kui Jia},
  booktitle  = cvpr,
  title      = {{RGBD2:} Generative Scene Synthesis via Incremental View
               Inpainting Using {RGBD} Diffusion Models},
  year       = {2023}
}

@InProceedings{cohen-bar23set-the-scene:,
  author     = {Dana Cohen{-}Bar and Elad Richardson and Gal Metzer and Raja
               Giryes and Daniel Cohen{-}Or},
  booktitle  = iccvw,
  title      = {Set-the-Scene: Global-Local Training for Generating
               Controllable {NeRF} Scenes},
  year       = {2023}
}

@Article{fridman23scenescape:,
  author     = {Rafail Fridman and Amit Abecasis and Yoni Kasten and Tali
               Dekel},
  journal    = {CoRR},
  title      = {SceneScape: Text-Driven Consistent Scene Generation},
  volume     = {abs/2302.01133},
  year       = {2023}
}

@InProceedings{chung23luciddreamer:,
  author     = {Jaeyoung Chung and Suyoung Lee and Hyeongjin Nam and Jaerin
               Lee and Kyoung Mu Lee},
  booktitle  = arxiv,
  title      = {{LucidDreamer}: Domain-free Generation of 3D Gaussian
               Splatting Scenes},
  year       = {2023}
}

@InProceedings{charatan24pixelsplat:,
  author     = {David Charatan and Sizhe Li and Andrea Tagliasacchi and
               Vincent Sitzmann},
  booktitle  = cvpr,
  title      = {{pixelSplat}: {3D Gaussian} Splats from Image Pairs for
               Scalable Generalizable {3D} Reconstruction},
  year       = {2024}
}

@InProceedings{wang24dust3r:,
  author     = {Shuzhe Wang and Vincent Leroy and Yohann Cabon and Boris
               Chidlovskii and Jerome Revaud},
  booktitle  = cvpr,
  title      = {{DUSt3R}: Geometric {3D} Vision Made Easy},
  year       = {2024}
}

@InProceedings{chen24scenetex:,
  author     = {Dave Zhenyu Chen and Haoxuan Li and Hsin-Ying Lee and Sergey
               Tulyakov and Matthias Nie{\ss}ner},
  booktitle  = cvpr,
  title      = {SceneTex: High-Quality Texture Synthesis for Indoor Scenes
               via Diffusion Priors},
  year       = {2024}
}

@Article{tang23mvdiffusion:,
  author     = {Shitao Tang and Fuyang Zhang and Jiacheng Chen and Peng Wang
               and Yasutaka Furukawa},
  journal    = corr,
  title      = {{MVDiffusion}: Enabling Holistic Multi-view Image Generation
               with Correspondence-Aware Diffusion},
  volume     = {abs/2307.01097},
  year       = {2023}
}

@Article{lin23componerf:,
  author     = {Yiqi Lin and Haotian Bai and Sijia Li and Haonan Lu and
               Xiaodong Lin and Hui Xiong and Lin Wang},
  journal    = corr,
  title      = {{CompoNeRF}: Text-guided Multi-object Compositional NeRF with
               Editable 3D Scene Layout},
  volume     = {abs/2303.13843},
  year       = {2023}
}

@InProceedings{po24compositional,
  author     = {Ryan Po and Gordon Wetzstein},
  booktitle  = threedv,
  title      = {Compositional {3D} Scene Generation using Locally Conditioned
               Diffusion},
  year       = {2024}
}

@InProceedings{raistrick23infinite,
  author     = {Alexander Raistrick and Lahav Lipson and Zeyu Ma and Lingjie
               Mei and Mingzhe Wang and Yiming Zuo and Karhan Kayan and
               Hongyu Wen and Beining Han and Yihan Wang and Alejandro Newell
               and Hei Law and Ankit Goyal and Kaiyu Yang and Jia Deng},
  booktitle  = cvpr,
  title      = {Infinite Photorealistic Worlds using Procedural Generation},
  year       = {2023}
}

@Article{kerbl233d-gaussian,
  author     = {Kerbl, Bernhard and Kopanas, Georgios and Leimk{\"u}hler,
               Thomas and Drettakis, George},
  journal    = siggraph,
  number     = {4},
  title      = {{3D} {Gaussian Splatting} for Real-Time Radiance Field
               Rendering},
  volume     = {42},
  year       = {2023}
}

@PhDThesis{mildenhall20neural,
  author     = {Ben Mildenhall},
  school     = {University of California, Berkeley, {USA}},
  title      = {Neural Scene Representations for View Synthesis},
  year       = {2020}
}

@Article{zhang23nerflets:,
  author     = {Xiaoshuai Zhang and Abhijit Kundu and Thomas A. Funkhouser
               and Leonidas J. Guibas and Hao Su and Kyle Genova},
  journal    = corr,
  title      = {Nerflets: Local Radiance Fields for Efficient Structure-Aware
               3D Scene Representation from 2D Supervision},
  volume     = {abs/2303.03361},
  year       = {2023}
}

@InProceedings{kim23neuralfield-ldm:,
  author     = {Seung Wook Kim and Bradley Brown and Kangxue Yin and Karsten
               Kreis and Katja Schwarz and Daiqing Li and Robin Rombach and
               Antonio Torralba and Sanja Fidler},
  booktitle  = cvpr,
  title      = {{NeuralField-LDM}: Scene Generation with Hierarchical Latent
               Diffusion Models},
  year       = {2023}
}

@Article{xiang233d-aware,
  author     = {Jianfeng Xiang and Jiaolong Yang and Binbin Huang and Xin
               Tong},
  journal    = corr,
  title      = {{3D}-aware Image Generation using {2D} Diffusion Models},
  volume     = {abs/2303.17905},
  year       = {2023}
}

@InProceedings{yang21learning,
  author     = {Bangbang Yang and Yinda Zhang and Yinghao Xu and Yijin Li and
               Han Zhou and Hujun Bao and Guofeng Zhang and Zhaopeng Cui},
  booktitle  = iccv,
  title      = {Learning Object-Compositional Neural Radiance Field for
               Editable Scene Rendering},
  year       = {2021}
}

@InProceedings{xu23discoscene:,
  author     = {Yinghao Xu and Menglei Chai and Zifan Shi and Sida Peng and
               Ivan Skorokhodov and Aliaksandr Siarohin and Ceyuan Yang and
               Yujun Shen and Hsin{-}Ying Lee and Bolei Zhou and Sergey
               Tulyakov},
  booktitle  = cvpr,
  title      = {{DisCoScene}: Spatially Disentangled Generative Radiance
               Fields for Controllable {3D}-aware Scene Synthesis},
  year       = {2023}
}

@InProceedings{poole23dreamfusion:,
  author     = {Ben Poole and Ajay Jain and Jonathan T. Barron and Ben
               Mildenhall},
  booktitle  = iclr,
  title      = {{DreamFusion}: Text-to-{3D} using {2D} Diffusion},
  year       = {2023}
}

@InProceedings{kulkarni21whats,
  author     = {Kulkarni, Nilesh and Johnson, Justin and Fouhey, David F},
  booktitle  = eccv,
  title      = {What's behind the couch? Directed ray distance functions
               ({DRDF}) for {3D} scene reconstruction},
  year       = 2021
}

@InProceedings{xu22sinnerf:,
  author     = {Xu, Dejia and Jiang, Yifan and Wang, Peihao and Fan, Zhiwen
               and Shi, Humphrey and Wang, Zhangyang},
  booktitle  = eccv,
  title      = {{SinNeRF}: Training neural radiance fields on complex scenes
               from a single image},
  year       = 2022
}

@InProceedings{li22compnvs:,
  author     = {Li, Zuoyue and Fan, Tianxing and Li, Zhenqiang and Cui,
               Zhaopeng and Sato, Yoichi and Pollefeys, Marc and Oswald,
               Martin R},
  booktitle  = eccv,
  title      = {{CompNVS}: Novel view synthesis with scene completion},
  year       = 2022
}

@Article{bautista22gaudi:,
  author     = {Miguel {\'{A}}ngel Bautista and Pengsheng Guo and Samira
               Abnar and Walter Talbott and Alexander Toshev and Zhuoyuan
               Chen and Laurent Dinh and Shuangfei Zhai and Hanlin Goh and
               Daniel Ulbricht and Afshin Dehghan and Josh M. Susskind},
  journal    = corr,
  title      = {{GAUDI:} {A} Neural Architect for Immersive {3D} Scene
               Generation},
  volume     = {abs/2207.13751},
  year       = {2022}
}

@InProceedings{chan22efficient,
  author     = {Eric R. Chan and Connor Z. Lin and Matthew A. Chan and Koki
               Nagano and Boxiao Pan and Shalini De Mello and Orazio Gallo
               and Leonidas J. Guibas and Jonathan Tremblay and Sameh Khamis
               and Tero Karras and Gordon Wetzstein},
  booktitle  = cvpr,
  journal    = {CoRR},
  title      = {Efficient Geometry-aware {3D} Generative Adversarial
               Networks},
  year       = {2022}
}

@InProceedings{mildenhall20nerf:,
  author     = {Ben Mildenhall and Pratul P. Srinivasan and Matthew Tancik
               and Jonathan T. Barron and Ravi Ramamoorthi and Ren Ng},
  booktitle  = eccv,
  title      = {{NeRF}: Representing Scenes as Neural Radiance Fields for
               View Synthesis},
  year       = {2020}
}

@InProceedings{dosovitskiy21an-image,
  author     = {Alexey Dosovitskiy and Lucas Beyer and Alexander Kolesnikov
               and Dirk Weissenborn and Xiaohua Zhai and Thomas Unterthiner
               and Mostafa Dehghani and Matthias Minderer and Georg Heigold
               and Sylvain Gelly and Jakob Uszkoreit and Neil Houlsby},
  booktitle  = iclr,
  title      = {An Image Is Worth 16$\times$16 Words: Transformers For Image
               Recognition At Scale},
  year       = {2021}
}

@Article{lorensen87marching,
  author     = {Lorensen, W. and Cline, H.},
  journal    = {ACM Computer Graphocs},
  number     = {24},
  title      = {Marching Cubes: {A} High Resolution {3D} Surface Construction
               Algorithm},
  volume     = {21},
  year       = {1987}
}

@Article{nguyen-phuoc19hologan:,
  author     = {Thu Nguyen{-}Phuoc and Chuan Li and Lucas Theis and Christian
               Richardt and Yong{-}Liang Yang},
  journal    = corr,
  title      = {{HoloGAN}: Unsupervised learning of {3D} representations from
               natural images},
  volume     = {abs/1904.01326},
  year       = {2019}
}

@InProceedings{goodfellow14generative,
  author     = {Ian J. Goodfellow and Jean Pouget{-}Abadie and Mehdi Mirza
               and Bing Xu and David Warde{-}Farley and Sherjil Ozair and
               Aaron C. Courville and Yoshua Bengio},
  booktitle  = nips,
  title      = {Generative Adversarial Nets},
  year       = {2014}
}

@InProceedings{chen25autopartgen,
  author     = {Minghao Chen and Jianyuan Wang and Roman Shapovalov and Tom
               Monnier and Hyunyoung Jung and Dilin Wang and Rakesh Ranjan
               and Iro Laina and Andrea Vedaldi},
  booktitle  = nips,
  title      = {{AutoPartGen}: Autogressive {3D} Part Generation and
               Discovery},
  year       = {2025}
}

@InProceedings{engstler25syncity,
  author     = {Paul Engstler and Aleksandar Shtedritski and Iro Laina and
               Christian Rupprecht and Andrea Vedaldi},
  booktitle  = iccv,
  title      = {{SynCity}: Training-Free Generation of {3D} Cities},
  year       = {2025}
}

@InProceedings{wang25vggt,
  author     = {Jianyuan Wang and Minghao Chen and Nikita Karaev and Andrea
               Vedaldi and Christian Rupprecht and David Novotny},
  booktitle  = cvpr,
  title      = {{VGGT}: Visual Geometry Grounded Transformer},
  year       = {2025}
}

@InProceedings{engstler25invisible,
  author     = {Paul Engstler and Andrea Vedaldi and Iro Laina and Christian
               Rupprecht},
  booktitle  = threedv,
  title      = {Invisible Stitch: Generating Smooth {3D} Scenes with Depth
               Inpainting},
  year       = {2025}
}

@InProceedings{chen24mvsplat360,
  author     = {Yuedong Chen and Chuanxia Zheng and Haofei Xu and Bohan
               Zhuang and Andrea Vedaldi and Tat-Jen Cham and Jianfei Cai},
  booktitle  = nips,
  title      = {{MVSplat360}: Benchmarking 360 Generalizable {3D} Novel View
               Synthesis from Sparse Views},
  year       = {2024}
}

@InProceedings{szymanowicz25flash3d,
  author     = {Stanislaw Szymanowicz and Eldar Insafutdinov and Chuanxia
               Zheng and Dylan Campbell and Jo{\~a}o F. Henriques and
               Christian Rupprecht and Andrea Vedaldi},
  booktitle  = threedv,
  title      = {{Flash3D}: Feed-Forward Generalisable {3D} Scene
               Reconstruction from a Single Image},
  year       = {2025}
}

@Article{bensadoun24metatexture,
  author     = {Raphael Bensadoun and Yanir Kleiman and Idan Azuri and Omri
               Harosh and Andrea Vedaldi and Natalia Neverova and Oran
               Gafni},
  journal    = arxiv,
  title      = {{Meta 3D Texture Gen}: Fast and Consistent Texture Generation
               for {3D} Objects},
  year       = {2024}
}

@InProceedings{szymanowicz24splatter,
  author     = {Stanislaw Szymanowicz and Christian Rupprecht and Andrea
               Vedaldi},
  booktitle  = cvpr,
  title      = {{Splatter Image}: Ultra-Fast Single-View {3D}
               Reconstruction},
  year       = {2024}
}

@InProceedings{melas-kyriazi24im-3d,
  author     = {Luke Melas-Kyriazi and Iro Laina and Christian Rupprecht and
               Natalia Neverova and Andrea Vedaldi and Oran Gafni and
               Filippos Kokkinos},
  booktitle  = icml,
  title      = {{IM-3D}: Iterative Multiview Diffusion and Reconstruction for
               High-Quality {3D} Generation},
  year       = {2024}
}

@InProceedings{wimbauer23behind,
  author     = {Felix Wimbauer and Nan Yang and Christian Rupprecht and
               Daniel Cremers},
  booktitle  = cvpr,
  title      = {Behind the Scenes: Density Fields for Single View
               Reconstruction},
  year       = {2023}
}

\clearpage
\newpage
\beginappendix
\appendix
\section{Scenes generated by \method}%
\label{sec:appendix}

In the following several pages, we include a gallery of additional scenes generated by our \method system.

\foreach \name/\captiontext in {
    scene_1014_018/Space port,
    scene_1015_005/Fruit-themed village,
    scene_1015_008/Sci-fi colony,
    scene_1015_009/Old industrial dockyard,
    scene_1015_034/Steampunk miniature city,
    scene_1017_043/Ancient temple courtyard,
    scene_1017_050/Military outpost,
    scene_1017_054/Japanese style medieval town,
    scene_1029_002/Fantasy mushroom village,
    scene_1029_007/Ancient East Asian temple complex,
    scene_1029_013/Cargo yard,
    scene_1029_021/Desert town,
    scene_1029_iso15/Futuristic industrial complex,
    scene_1029_mj/Charming city block,
    scene_1029_mj4/Suburban neighborhood,
    scene_1029_mj8/Forest outpost,
    scene_1029_mj14/Seaside terminal,
    scene_1029_nano16/Halloween-themed village
}{
    \clearpage
    \begin{figure}[htbp]
        \centering
        \includegraphics[width=0.99\linewidth]{assets/scene_render_picked/\name.jpg}
        \caption{\captiontext}%
        \label{fig:\name}
    \end{figure}
    \clearpage
}

\end{document}